\documentclass[journal]{IEEEtran}

\usepackage[utf8]{inputenc} 
\usepackage[T1]{fontenc}    
\usepackage{hyperref}       
\usepackage{url}            
\usepackage{booktabs}       
\usepackage{amsfonts}       
\usepackage{amsmath, amssymb}
\usepackage{bbold}
\usepackage{xcolor}
\usepackage{enumitem}
\usepackage{tcolorbox}

\usepackage{nicefrac}       
\usepackage{microtype}      
\usepackage{lipsum}
\usepackage{tikz}

\usepackage[numbers]{natbib}

\newcommand\blfootnote[1]{%
  \begingroup
  \renewcommand\thefootnote{}\footnote{#1}%
  \addtocounter{footnote}{-1}%
  \endgroup
}

\DeclareRobustCommand{\myhammer}{%
  \begingroup\normalfont
  \includegraphics[height=1.1\fontcharht\font`\B]{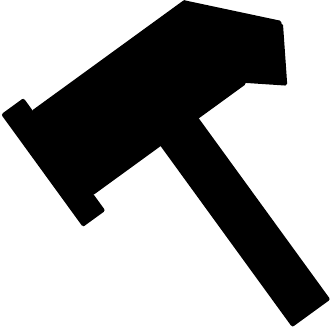}%
  \endgroup
}

\usepackage{epsf}

\usepackage{psfrag}
\usepackage{graphicx}
\usepackage{amssymb}
\usepackage{amsbsy}
\usepackage{epsfig}
\usepackage{ifthen}
\usepackage{eucal}
\usepackage{theorem}

\newcommand{\commentout}[1]{}

\newcommand{\RR}{\mathbb{R}}    
\newcommand{\R}{\mathbb{R}}    

\newcommand{\x}{\mathbf{x}}    
\newcommand{\argmin}{\mathop{\mathrm{argmin}\,}}

\newcommand{\eq}[1]{(\protect\ref{#1})}

\newcommand{\independent}{\perp\mkern-11mu\perp}

\newcommand{\E}{\mathbb{E}}

\newcommand{\B}[1]{\mathbf{#1}}

\newcommand{\PA}{\B{PA}}

  \mathchardef\ordinarycolon\mathcode`\:
  \mathcode`\:=\string"8000
  \begingroup \catcode`\:=\active
    \gdef:{\mathrel{\mathop\ordinarycolon}}
  \endgroup

\usepackage{amsmath,amsfonts,bm}

\def\1{\bm{1}}

\DeclareMathAlphabet{\mathsfit}{\encodingdefault}{\sfdefault}{m}{sl}
\SetMathAlphabet{\mathsfit}{bold}{\encodingdefault}{\sfdefault}{bx}{n}

\title{Towards Causal Representation Learning}

\author{Bernhard Sch\"olkopf $^\dagger$, Francesco Locatello $^\dagger$, Stefan Bauer $^\star$, Nan Rosemary Ke $^\star$, Nal Kalchbrenner \\ Anirudh Goyal, Yoshua Bengio
\thanks{$\dagger$: equal contribution.}
\thanks{$\star$: equal contribution.}
\thanks{B. Sch\"olkopf is at the Max-Planck Institute for Intelligent Systems, Max-Planck-Ring 4, 72076 T\"ubingen, Germany, \texttt{bs@tuebingen.mpg.de}.}
\thanks{F. Locatello is at ETH Zurich, Computer Science Department and the Max Planck Institute for Intelligent Systems. Work partially done while interning at Google Research Amsterdam. \texttt{francesco.locatello@gmail.com}}
\thanks{S. Bauer is at the Max-Planck Institute for Intelligent Systems, \texttt{stefan.bauer@tuebingen.mpg.de}.}
\thanks{N. R. Ke is at Mila and the University of Montreal, \textit{rosemary.nan.ke@gmail.com}.}
\thanks{N. Kalchbrenner is at Google Research Amsterdam, \texttt{nalk@google.com}.}
\thanks{A. Goyal is at Mila and the University of Montreal, 
 \texttt{anirudhgoyal9119@gmail.com}}
\thanks{Y. Bengio is at Mila, the University of Montreal, CIFAR Senior Fellow \textit{yoshua.bengio@mila.quebec}}
}

\begin{document}
\maketitle

\begin{abstract}

The two fields of machine learning and graphical causality arose and developed separately. However, there is now cross-pollination and increasing interest in both fields to benefit from the advances of the other. In the present paper, we review fundamental concepts of causal inference and relate them to crucial open problems of machine learning, including transfer and generalization, thereby assaying how causality can contribute to modern machine learning research. This also applies in the opposite direction: we note that most work in causality starts from the premise that the causal variables are given. A central problem for AI and causality is, thus, causal representation learning, the discovery of high-level causal variables from low-level observations. Finally, we delineate some implications of causality for machine learning and propose key research areas at the intersection of both communities.
\end{abstract}

\section{Introduction}
\looseness=-1 If we compare what machine learning can do to what animals accomplish, we observe that the former is rather limited at some crucial feats where natural intelligence excels. These include transfer to new problems and any form of generalization that is not from one data point to the next (sampled from the same distribution), but rather from one problem to the next --- both have been termed {\em generalization}, but the latter is a much harder form thereof, sometimes referred to as {\em horizontal}, {\em strong}, or {\em out-of-distribution} generalization.  This shortcoming is not too surprising, given that machine learning often disregards information that animals use heavily: interventions in the world, domain shifts, temporal structure --- by and large, we consider these factors a nuisance and try to engineer them away. In accordance with this, the majority of current successes of machine learning boil down to large scale pattern recognition on suitably collected {\em independent and identically distributed (i.i.d.)}\ data.

To illustrate the implications of this choice and its relation to causal models, we start by highlighting key research challenges. 

\paragraph{Issue 1 -- Robustness} With the widespread adoption of deep learning approaches in computer vision~\citep{he2016deep,krizhevsky2012imagenet}, natural language processing~\citep{devlin2018bert}, and speech recognition~\citep{graves2013speech}, a substantial body of literature explored the robustness of the prediction of state-of-the-art deep neural network architectures. The underlying motivation originates from the fact that in the real world there is often little control over the distribution from which the data comes from. In computer vision ~\citep{geirhos2018imagenet,shetty2019not}, changes in the test distribution may, for instance, come from aberrations like camera blur, noise or compression quality~\citep{hendrycks2019benchmarking,karahan2016image,michaelis2019benchmarking,roy2018effects}, or from shifts, rotations, or viewpoints~\citep{azulay2019deep,barbu2019objectnet,engstrom2017exploring,zhang2019making}. Motivated by this, new benchmarks were proposed to specifically test generalization of classification and detection methods with respect to simple algorithmically generated interventions like spatial shifts, blur, changes in brightness or contrast~\citep{hendrycks2019benchmarking,michaelis2019benchmarking},  time consistency~\citep{gu2019using,shankarimage},  control over background and rotation~\citep{barbu2019objectnet}, as well as images collected in multiple environments~\citep{beery2018recognition}. Studying the failure modes of deep neural networks from simple interventions has the potential to lead to insights into the inductive biases of state-of-the-art architectures. 
So far, there has been no definitive consensus on how to solve these problems, although progress has been made using data augmentation, pre-training, self-supervision, and architectures with suitable inductive biases w.r.t.\ a perturbation of interest~\citep{tangent_prop,djolonga2020robustness,engstrom2017exploring,kolesnikov2019big,michaelis2019benchmarking,roy2018effects}.
It has been argued~\citep{PetJanSch17} that such fixes may not be sufficient, and generalizing well outside the i.i.d.~setting requires learning not mere statistical associations between variables, but an underlying \textit{causal model}. The latter contains the mechanisms giving rise to the observed statistical dependences, and allows to model distribution shifts through the notion of interventions \citep{Pearl2009,Spirtes2000,Schoelkopf2012,Bottou2013,PetJanSch17,ParKilRojSch18}.

 \paragraph{Issue 2 -- Learning Reusable Mechanisms} \looseness=-1Infants’ understanding of physics relies upon objects that can be tracked over time and behave consistently~\citep{dehaene2020we,spelke1990principles}. Such a representation allows children to quickly learn new tasks as their knowledge and intuitive understanding of physics can be re-used~\citep{battaglia2013simulation,dehaene2020we,lake2017building,teglas2011pure}. Similarly, intelligent agents that robustly solve real-world tasks need to re-use and re-purpose their knowledge and skills in novel scenarios. Machine learning models that incorporate or learn structural knowledge of an environment have been shown to be more efficient and generalize better~\citep{battaglia2016interaction,bapst2019structured, battaglia2018relational,RIMs, rahaman2021spatially, santoro2017simple, bahdanau2018systematic, zambaldi2018deep, berner2019dota,gondal2019transfer,goyal2020object,kulkarni2019unsupervised,locatello2019fairness,mrowca2018flexible,sanchez2020learning,sun2019stochastic,vinyals2019grandmaster,yi2019clevrer,dittadi2021on,parascandolo2021learning}. In a modular representation of the world where the modules correspond to physical causal mechanisms, many modules can be expected to behave similarly across different tasks and environments. An agent facing a new environment or task may thus only need to adapt a few modules in its internal representation of the world~\citep{SchJanLop16,RIMs}. When learning a causal model, one should thus require fewer examples to adapt as most knowledge, i.e., modules, can be re-used without further training.

\paragraph{A Causality Perspective} \looseness=-1Causation is a subtle concept that cannot be fully described using the language of Boolean logic~\citep{lewis1974causation} or that of probabilistic inference; it requires the additional notion of {\em intervention}~\citep{Spirtes2000,Pearl2009}. 
The manipulative definition of causation~\citep{Spirtes2000,Pearl2009,imbens2015causal} focuses on the fact that conditional probabilities (“seeing people with open umbrellas suggests that it is raining”)
cannot reliably predict the outcome of an active intervention (“closing umbrellas does not stop the rain”).
Causal relations can also be viewed as the components of reasoning chains~\citep{lewis1974causation} that provide predictions for situations that are very far from the observed distribution and may even remain purely hypothetical~\citep{Lorenz73,Pearl2009} or require conscious deliberation \citep{kahneman2011thinking}. In that sense, discovering causal relations means acquiring robust knowledge that holds beyond the support of an observed data distribution and a set of training tasks, and it extends to situations involving forms of reasoning.

\emph{Our Contributions: \,}\looseness=-1 In the present paper, we argue that causality, with its focus on representing structural knowledge about the data generating process that allows interventions and changes, can contribute towards understanding and resolving some limitations of current machine learning methods. This would take the field a step closer to a form of artificial intelligence that involves {\em thinking} in the sense of Konrad Lorenz, i.e., acting in an imagined space \cite{Lorenz73}. Despite its  success, statistical learning provides a rather superficial description of reality that only holds when the experimental conditions are fixed. Instead, the field of {\em causal learning} seeks to model the effect of interventions and distribution changes with a combination of data-driven learning and assumptions not already included in the statistical description of a system. The present work reviews and synthesizes key contributions that have been made to this end\blfootnote{The present paper expands \cite{1911.10500}, leading to partial text overlap.}: 
\begin{itemize}[leftmargin=10pt]
    \item We describe different levels of modeling in physical systems in Section \ref{sec:levelsofcausalmodeling} and present the differences between causal and statistical models in Section \ref{sec:causal_models}. We do so not only in terms of modeling abilities but also discuss the assumptions and challenges involved.
    \item We expand on the Independent Causal Mechanisms (ICM) principle as a key component that enables the estimation of causal relations from data in Section \ref{sec:icm}. In particular, we state the Sparse Mechanism Shift hypothesis as a consequence of the \hyperlink{pri:im}{ICM} principle and discuss its implications for learning causal models.
    \item We review existing approaches to learn causal relations from appropriate descriptors (or features) in Section \ref{sec:causal_discovery}. We cover both classical approaches and modern re-interpretations based on deep neural networks, with a focus on the underlying principles that enable causal discovery.
    \item We discuss how useful models of reality may be learned from data in the form of causal representations, and discuss several current problems of machine learning from a causal point of view in Section \ref{sec:learning_variables}.
    \item \looseness=-1We assay the implications of causality for practical machine learning in Section \ref{sec:benefits}. Using causal language, we revisit robustness and generalization, as well as existing common practices such as semi-supervised learning, self-supervised learning, data augmentation, and pre-training. We discuss examples at the intersection between causality and machine learning in scientific applications and speculate on the advantages of combining the strengths of both fields to build a more versatile AI.
\end{itemize}

\section{Levels of Causal Modeling}
\label{sec:levelsofcausalmodeling}
\begin{table*}[t]
\caption{\label{t:taxonomy}A simple taxonomy of models. The most detailed model (top) is a mechanistic or physical one, usually in terms of differential equations. At the other end of the spectrum (bottom), we have a purely statistical model; this can be learned from data, but it often provides little insight beyond modeling associations between epiphenomena.
Causal models can be seen as descriptions that lie in between, abstracting away from physical realism while retaining the power to answer certain interventional or counterfactual questions. }
\begin{center}
{\small
\begin{tabular}{ c || c | c | c | c | c }
Model& Predict in i.i.d.\ & Predict under distr.\ & Answer counter- & Obtain & Learn from  \\
 & setting & shift/intervention & factual questions & physical insight & data \\\hline\hline
Mechanistic/physical  & yes & yes & yes & yes & ? \\\hline
Structural causal  & yes & yes & yes & ? & ?\\\hline
Causal graphical & yes & yes & no & ? & ?\\\hline
Statistical & yes & no & no & no & yes
\end{tabular}}
\end{center}
\end{table*}

The gold standard for modeling natural phenomena is a set of coupled differential equations modeling physical mechanisms responsible for the time evolution. This allows us to predict the future behavior of a physical system, reason about the effect of interventions, and predict {\em statistical} dependencies between variables that are generated by coupled time evolution. It also offers physical insights, explaining the functioning of the system, and lets us read off its causal structure.
To this end, consider the coupled set of differential equations
\begin{equation}\label{eq:ode}
\frac{d\x}{dt} = f(\x), \; \x \in \RR^d,
\end{equation}
with initial value $\x(t_0)=\x_0$. The Picard–Lindel{\"o}f theorem states that at least locally, if $f$ is Lipschitz, there exists a unique solution $\x(t)$. This implies in particular that the immediate future of $\x$ is implied by its past values.

If we formally write this in terms of infinitesimal differentials $dt$ and $d\x = \x(t+dt)-\x(t)$, we get:
\begin{equation}\label{eq:odecausal}
\x(t+dt) = \x(t) + dt\cdot f(\x(t)).
\end{equation}
From this, we can ascertain which entries of the vector $\x(t)$ mathematically determine the future of others $\x(t+dt)$. 
This tells us that if we have a physical system whose physical mechanisms are correctly described using such an ordinary differential equation \eq{eq:ode}, solved for $\frac{d\x}{dt}$ (i.e., the derivative only appears on the left-hand side), then its causal structure can be directly read off.\footnote{Note that this requires that the differential equation system describes the causal physical mechanisms. If, in contrast, we considered a set of differential equations that phenomenologically correctly describe the time evolution of a system without capturing the underlying mechanisms (e.g., due to unobserved confounding, or a form of course-graining that does not preserve the causal structure \cite{Rubensteinetal17}), then \eq{eq:odecausal} may not be causally meaningful \citep{1911.10500,peters2020causal}.}

\looseness=-1While a differential equation is a rather comprehensive description of a system, a statistical model can be viewed as a much more superficial one. It often does not refer to dynamic processes; instead, it tells us how some of the variables allow prediction of others as long as experimental conditions do not change. E.g., if we drive a differential equation system with certain types of noise, or we average over time, then it may be the case that statistical dependencies between components of $\x$ emerge, and those can then be exploited by machine learning. Such a model does not allow us to predict the effect of interventions; however, its strength is that it can often be learned from observational data, while a differential equation usually requires an intelligent human to come up with it.
Causal modeling lies in between these two extremes. Like models in physics, it aims to provide understanding and predict the effect of interventions. However, causal discovery and learning try to arrive at such models in a data-driven way, replacing expert knowledge with weak and generic assumptions.
The overall situation is summarized in Table~\ref{t:taxonomy}, adapted from \citep{PetJanSch17}. Below, we address some of the tasks listed in Table~\ref{t:taxonomy} in more detail.

\subsection{Predicting in the i.i.d.\ setting}
Statistical models are a superficial description of reality as they are only required to model associations. For a given set of input examples $X$ and target labels $Y$, we may be interested in approximating $P(Y | X)$ to answer questions like: ``what is the probability that this particular image contains a dog?'' or ``what is the probability of heart failure given certain diagnostic measurements (e.g., blood pressure) carried out on a patient?''. Subject to suitable assumptions, these questions can be provably answered by observing a sufficiently large amount of i.i.d.\ data from $P(X,Y)$~\citep{Vapnik98}.
Despite the impressive advances of machine learning, causality offers an under-explored complement: accurate predictions may not be sufficient to inform decision making. For example, the frequency of storks is a reasonable predictor for human birth rates in Europe~\citep{matthews2000storks}. However, as there is no direct causal link between those two variables, a change to the stork population would not affect the birth rates, even though a statistical model may predict so. The predictions of a statistical model are only accurate within identical experimental conditions. Performing an intervention changes the data distribution, which may lead to (arbitrarily) inaccurate predictions~\citep{Pearl2009,Spirtes2000,Schoelkopf2012,PetJanSch17}.

\subsection{Predicting Under Distribution Shifts} Interventional questions are more challenging than predictions as they involve actions that take us out of the usual i.i.d.\ setting of statistical learning. Interventions may affect both the value of a subset of causal variables and their relations. For example, ``is increasing the number of storks in a country going to boost its human birth rate?'' and ``would fewer people smoke if cigarettes were more socially stigmatized?''.
As interventions change the joint distribution of the variables of interest, classical statistical learning guarantees~\citep{Vapnik98} no longer apply. On the other hand, learning about interventions may allow to train predictive models that are robust against the changes in distribution that naturally happen in the real world. Here, interventions do not need to be deliberate actions to achieve a goal. Statistical relations may change dynamically over time (e.g.,  people's preferences and tastes) or there may simply be a mismatch between a carefully controlled training distribution and the test distribution of a model deployed in production. The robustness of deep neural networks has recently been scrutinized and become an active research topic related to causal inference. We argue that predicting under distribution shift should not be reduced to just the accuracy on a test set. If we wish to incorporate learning algorithms into human decision making, we need to trust that the predictions of the algorithm will remain valid if the experimental conditions are changed. 

\subsection{Answering Counterfactual Questions}
\looseness=-1Counterfactual problems involve reasoning about why things happened, imagining the consequences of different actions in hindsight, and determining which actions would have achieved a desired outcome.
Answering counterfactual questions can be more difficult than answering interventional questions. 
However, this may be a key challenge for AI, as an intelligent agent may benefit from imagining the consequences of its actions as well as understanding in retrospect what led to certain outcomes, at least to some degree of approximation.\footnote{Note that the two types of questions occupy a continuum: to this end, consider a probability which is both conditional and interventional $P(A|B,do(C))$. If $B$ is the empty set, we have a classical intervention; if $B$ contained all (unobserved) noise terms, we have a counterfactual. If $B$ is not identical to the noise terms, but nevertheless informative about them, we get something in between. For instance, reinforcement learning practitioners may call $Q$ functions as providing counterfactuals, even though they model $P$(return from $t|\,$agent state at time $t$, $do\,$(action at time $t$)), and therefore closer to an intervention (which is why they can be estimated from data).} 
We have above mentioned the example of statistical predictions of heart failure. An interventional question would be ``how does the probability of heart failure change if we convince a patient to exercise regularly?'' A counterfactual one would be ``would a given patient have suffered heart failure if they had started exercising a year earlier?''. As we shall discuss below, counterfactuals, or approximations thereof, are especially critical in reinforcement learning. They can enable agents to reflect on their decisions and formulate hypotheses that can be empirically verified in a process akin to the scientific method.

\subsection{Nature of Data: Observational, Interventional, (Un)structured}
\looseness=-1The data format plays a substantial role in which type of relation can be inferred. We can distinguish two axes of data modalities: observational versus interventional, and hand-engineered versus raw (unstructured) perceptual input. 

\textit{Observational and Interventional Data:} an extreme form of data which is often assumed but seldom strictly available is observational i.i.d.\ data, where each data point is independently sampled from the same distribution. Another extreme is interventional data with known interventions, where we observe data sets sampled from multiple distributions each of which is the result of a known intervention. In between, we have data with (domain) shifts or unknown interventions. This is observational in the sense that the data is only observed passively, but it is interventional in the sense that there are interventions/shifts, but unknown to us. 

\textit{Hand Engineered Data vs.\ Raw Data:} especially in classical AI, data is often assumed to be structured into high-level and semantically meaningful variables which may partially (modulo some variables being unobserved) correspond to the causal variables of the underlying graph. \textit{Raw Data}, in contrast, is unstructured and does not expose any direct information about causality.

While statistical models are weaker than causal models, they can be efficiently learned from observational data alone on both hand-engineered features and raw perceptual input such as images, videos, speech etc. On the other hand, although methods for learning causal structure from observations exist \cite{Spirtes2000,PetJanSch17,Shimizu2006,Hoyer2008,Mooij2009,PetMooJanSch14,Kpotufe14,BauSchPet16,Sun2006,Zhang2009,Mooij11,Janzing2009uai,Peters2011b,Mooijetal16,Vreeken,1711.08936,LopMuaSchTol15},
learning causal relations frequently requires collecting data from multiple environments, or the ability to perform interventions \citep{Tian2001}. In some cases, it is assumed that all common causes of measured variables are also observed (causal sufficiency).\footnote{There are also algorithms that do not require causal sufficiency \cite{Spirtes2000}.} Overall, a significant amount of prior knowledge is encoded in which variables are measured. Moving forward, one would hope to develop methods that replace expert data collection with suitable inductive biases and learning paradigms such as meta-learning and self-supervision. 
If we wish to learn a causal model that is useful for a particular set of tasks and environments, the appropriate granularity of the high-level variables depends on the tasks of interest and on the type of data we have at our disposal, for example which interventions can be performed and what is known about the domain.

\section{Causal Models and Inference}\label{sec:causal_models}
As discussed, reality can be modeled at different levels, from the physical one to statistical associations between epiphenomena. In this section, we expand on the difference between statistical and causal modeling and review a formal language to talk about interventions and distribution changes.
\subsection{Methods driven by i.i.d.\ data}

\looseness=-1The machine learning community has produced impressive successes with machine learning applications to big data problems \citep{LeCBenHin15,mnih2015human,schrittwieser2019mastering,silver2016mastering,deng2009imagenet}. In these successes, there are several trends at work \citep{schoelkopf15}:
(1) we have massive amounts of data, often from simulations or large scale human labeling, (2) we use high capacity machine learning systems (i.e., complex function classes with many adjustable parameters), (3) we employ high-performance computing systems, and finally (often ignored, but crucial when it comes to causality) (4) the problems are i.i.d.\ 
The latter can be guaranteed by the construction of a task including training and test set (e.g., image recognition using benchmark datasets). Alternatively, problems can be made approximately i.i.d., e.g.. by carefully collecting the right training set for a given application problem, or by methods such as ``experience replay'' \citep{mnih2015human} where a reinforcement learning agent stores observations in order to later permute them for the purpose of re-training. 

\looseness=-1For i.i.d.\ data, strong universal consistency results from statistical learning theory apply, guaranteeing convergence of a learning algorithm to the lowest achievable risk. Such algorithms do exist, for instance, nearest neighbor classifiers, support vector machines, and neural networks \citep{Vapnik98,SchSmo02,SteChr08,farago2006strong}. 
Seen in this light, it is not surprising that we can indeed match or surpass human performance if given enough data. However, current machine learning methods  often perform poorly when faced with problems that violate the i.i.d.\ assumption, yet seem trivial to humans. Vision systems can be grossly misled if an object that is normally recognized with high accuracy is placed in a context that {\em in the training set} may be negatively correlated with the presence of the object.  
Distribution shifts may also arise from simple corruptions that are common in real-world data collection pipelines~\citep{Baird90,hendrycks2019benchmarking,karahan2016image,michaelis2019benchmarking,roy2018effects}. An example of this is the impact of socio-economic factors in clinics in Thailand on the accuracy of a detection system for Diabetic Retinopathy~\citep{beede2020human}.
More dramatically, the phenomenon of ``adversarial vulnerability'' \citep{1312.6199} highlights how even tiny but targeted violations of the i.i.d.\ assumption, generated by adding suitably chosen perturbations to images, imperceptible to humans, can lead to dangerous errors such as confusion of traffic signs. Overall, it is fair to say that much of the current practice (of solving i.i.d.\ benchmark problems) and most theoretical results (about generalization in i.i.d.\ settings) fail to tackle the hard open challenge of generalization across problems.

\looseness=-1To further understand how the i.i.d.\ assumption is problematic, let us consider a shopping example. Suppose Alice is looking for a laptop rucksack on the internet (i.e., a rucksack with a padded compartment for a laptop). The web shop's recommendation system suggests that she should buy a laptop to go along with the rucksack. This seems odd because she probably already has a laptop, otherwise she would not be looking for the rucksack in the first place. In a way, the laptop is the cause, and the rucksack is an effect. 
Now suppose we are told whether a customer has bought a laptop. This reduces our uncertainty about whether she also bought a laptop rucksack, and vice versa –-- and it does so by the same amount (the {\em mutual information}), so the directionality of cause and effect is lost.
However, the directionality is present in the physical mechanisms generating statistical dependence, for instance the mechanism that makes a customer want to buy a rucksack once she owns a laptop.\footnote{Note that the physical mechanisms take place in time, and if available, time order may provide additional information about causality.}
Recommending an item to buy constitutes an intervention in a system, taking us outside the i.i.d.\ setting. We no longer work with the observational distribution, but a distribution where certain variables or mechanisms have changed. 

\subsection{The Reichenbach Principle: From Statistics to Causality}
\looseness=-1\citet{Reichenbach1956} clearly articulated the connection between causality and statistical dependence. He postulated:
\begin{tcolorbox}[colback=black!0!white]
\textit{Common Cause Principle}: if two observables $X$ and $Y$ are statistically dependent, then there exists a variable $Z$ that causally influences both and explains all the dependence in the sense of making them independent when conditioned on $Z$. 
\end{tcolorbox}
As a special case, this variable can coincide with $X$ or $Y$. 
Suppose that $X$ is the frequency of storks and $Y$ the human birth rate. If storks bring the babies, then the correct causal graph is $X\rightarrow Y$. If babies attract storks, it is $X \leftarrow Y$. If there is some other variable that causes both (such as economic development), we have $X \leftarrow Z \rightarrow Y$. 

\looseness=-1 Without additional assumptions, we cannot distinguish these three cases using observational data. The class of observational distributions over $X$ and $Y$ that can be realized by these models is the same in all three cases. A causal model thus contains genuinely more information than a statistical one.

\looseness=-1While causal structure discovery is hard if we have only two observables \citep{PetMooJanSch14}, the case of more observables is surprisingly easier, the reason being that in that case, there are nontrivial conditional independence properties \citep{Spohn78,Dawid79,GeiPea90} implied by causal structure. These generalize the Reichenbach Principle and can be described by using the language of causal graphs or structural causal models, merging probabilistic graphical models and the notion of interventions \citep{Spirtes2000,Pearl2009}. They are best described using directed functional parent-child relationships rather than conditionals. While conceptually simple in hindsight, this constituted a major step in the understanding of causality.

\subsection{Structural causal models (SCMs)}
The SCM viewpoint considers a set of {\em observables} (or {\em variables}) $X_1,\dots,X_n$ associated with the vertices of a directed acyclic graph (DAG). We assume that each observable is the result of an assignment
\begin{equation}\label{eq:SA}
X_i := f_i (\PA_i, U_i),   ~~~~ (i=1,\dots,n),
\end{equation}
using a deterministic function $f_i$ depending on $X_i$'s parents in the graph (denoted by $\PA_i$) and on an {\em unexplained} random variable $U_i$. Mathematically, the observables are thus random variables, too. Directed edges in the graph represent direct causation, since the parents are connected to $X_i$ by directed edges and through \eq{eq:SA} directly affect the assignment of $X_i$. The noise $U_i$ ensures that the overall object \eq{eq:SA} can represent a general conditional distribution $P(X_i|\PA_i)$, and the set of noises $U_1,\dots,U_n$ are assumed to be {\em jointly independent}. If they were not, then by the Common Cause Principle there should be another variable that causes their dependence, and thus our model would not be {\em causally sufficient}.

If we specify the distributions of $U_1,\dots,U_n$, recursive application of \eq{eq:SA} allows us to compute the {\em entailed observational joint distribution} $P(X_1,\dots,X_n)$. This distribution has structural properties inherited from the graph \citep{Lauritzen1996,Pearl2009}: it satisfies the {\em causal Markov condition} stating that conditioned on its parents, each $X_j$ is independent of its non-descendants. 

Intuitively, we can think of the independent noises as ``information probes'' that spread through the graph (much like independent elements of gossip can spread through a social network). Their information gets entangled, manifesting itself in a footprint of conditional dependencies making it possible to infer aspects of the graph structure from observational data using independence testing. Like in the gossip analogy, the footprint may not be sufficiently characteristic to pin down a unique causal structure. In particular, it certainly is not if there are only two observables, since any nontrivial conditional independence statement requires at least three variables.
The two-variable problem can be addressed by making additional assumptions, as not only the graph topology leaves a footprint in the observational distribution, but the functions $f_i$ do, too. This point is interesting for machine learning, where much attention is devoted to properties of function classes (e.g., priors or capacity measures), and we shall return to it below.

\paragraph{Causal Graphical Models}
The graph structure along with the joint independence of the noises implies a canonical factorization of the joint distribution entailed by \eq{eq:SA} into causal conditionals that we refer to as the {\em causal (or disentangled) factorization},
\begin{equation}\label{eq:cf}
P(X_1,\dots,X_n) = \prod_{i=1}^n  P(X_i \mid \PA_i).
\end{equation}
While many other {\em entangled factorizations} are possible, e.g., 
\begin{equation}\label{eq:non-cf}
P(X_1,\dots,X_n) = \prod_{i=1}^n P(X_i \mid X_{i+1},\dots,X_n),
\end{equation} 
the factorization \eq{eq:cf} yields practical computational advantages during inference, which is in general hard, even when it comes to non-trivial approximations~\citep{russell2002artificial}.
But more interestingly, it is the only one that decomposes the joint distribution into conditionals corresponding to the structural assignments \eq{eq:SA}. We think of these as the {\em causal mechanisms} that are responsible for all statistical dependencies among the observables. Accordingly, in contrast to \eq{eq:non-cf}, the disentangled factorization represents the joint distribution as a product of causal mechanisms.

\paragraph{Latent variables and Confounders}
Variables in a causal graph may be unobserved, which can make causal inference particularly challenging. Unobserved variables may \textit{confound} two observed variables so that they either appear statistically related while not being causally related (i.e., neither of the variables is an ancestor of the other), or their statistical relation is altered by the presence of the confounder (e.g., one variable is a causal ancestor for the other, but the confounder is a causal ancestor of both). 
Confounders may or may not be known or observed.

\paragraph{Interventions}
The SCM language makes it straightforward to formalize {\em interventions} as operations that modify a subset of assignments \eq{eq:SA}, e.g., changing $U_i$, setting $f_i$ (and thus $X_i$) to a constant, or changing the functional form of $f_i$ (and thus the dependency of $X_i$ on its parents) \citep{Spirtes2000,Pearl2009}. 

Several types of interventions may be possible \citep{eaton2007exact} which can be categorized as:
\textit{No intervention:} only observational data is obtained from the causal model. \textit{Hard/perfect:} the function in the structural assignment \eq{eq:SA} of a variable (or, analogously, of multiple variables) is set to a constant (implying that the value of the variable is fixed), and then the entailed distribution for the modified SCM is computed. \textit{Soft/imperfect:} the structural assignment \eq{eq:SA} for a variable is modified by changing the function or the noise term (this corresponds to changing the conditional distribution given its parents). \textit{Uncertain:} the learner is not sure which mechanism/variable is affected by the intervention. 

One could argue that stating the structural assignments as in \eq{eq:SA} is not yet sufficient to formulate a causal model. In addition, one should specify the set of possible interventions on the structural causal model. This may be done implicitly via the functional form of structural equations by allowing any intervention over the domain of the mechanisms. 
This becomes relevant when learning a causal model from data, as the SCM depends on the interventions. Pragmatically, we should aim at learning causal models that are useful for specific sets of tasks of interest~\citep{Rubensteinetal17,weichwald2019pragmatism} on appropriate descriptors (in terms of which causal statements they support) that must either be provided or learned. We will return to the assumptions that allow learning causal models and features in Section~\ref{sec:icm}.

\begin{figure*}[htb]
\begin{center}
\includegraphics[width=0.8\textwidth]{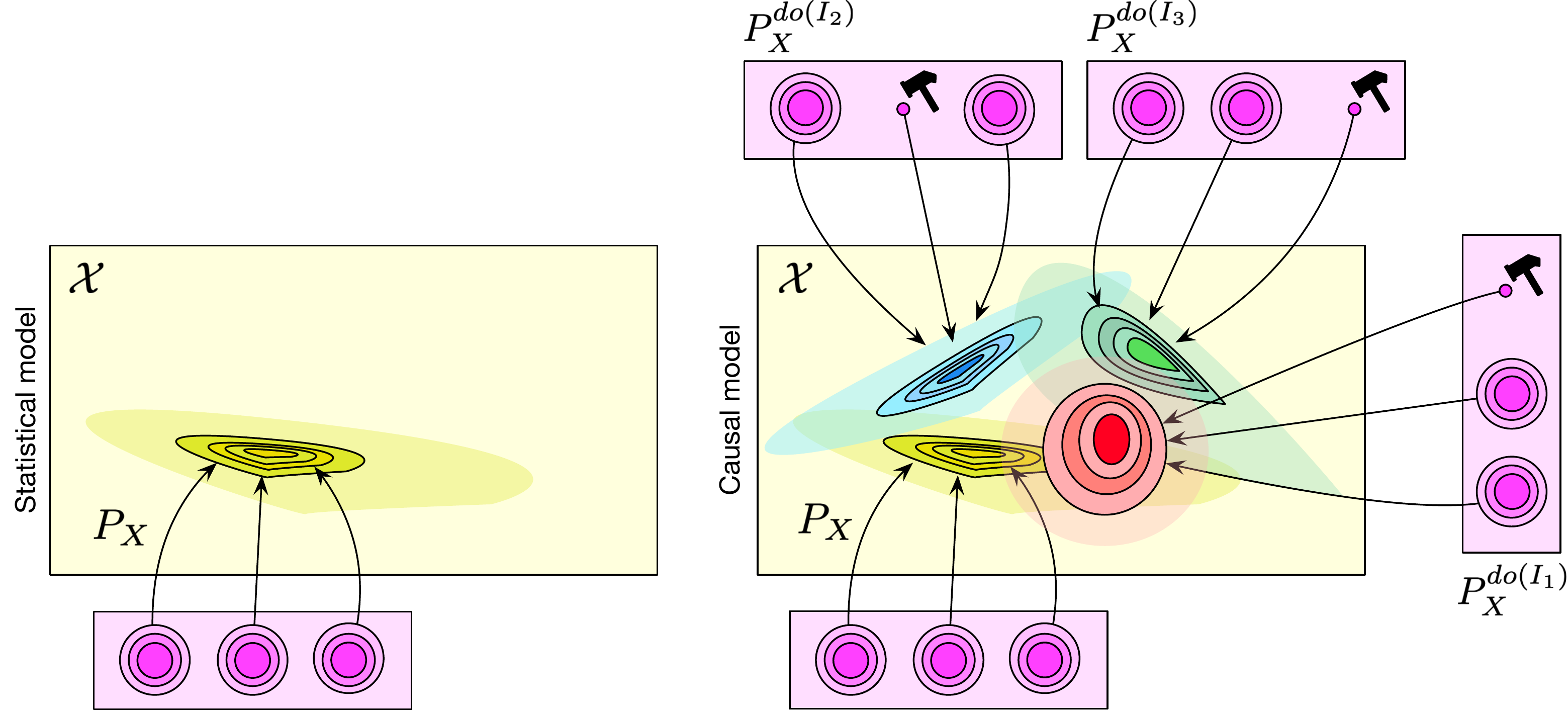}
\end{center}
  \caption{Difference between statistical (left) and causal models (right) on a given set of three variables. While a statistical model specifies a single probability distribution, a causal model represents a set of distributions, one for each possible intervention (indicated with a \myhammer{} 
 in the figure). 
 }\label{fig_stats_vs_causal}
\end{figure*}

\subsection{Difference Between Statistical Models, Causal Graphical Models, and SCMs} 
An example of the difference between a statistical and a causal model is depicted in Figure~\ref{fig_stats_vs_causal}. A statistical model may be defined for instance through a graphical model, i.e., a probability distribution along with a graph such that the former is Markovian with respect to the latter (in which case it can be factorized as \eq{eq:cf}).
However, the edges in a (generic) graphical model do not need to be causal \citep{GuyonJS2010}. For instance, the two graphs $X_1\rightarrow X_2 \rightarrow X_3$ and $X_1\leftarrow X_2 \leftarrow X_3$ imply the same conditional independence(s)
($X_1$ and $X_3$ are independent given $X_2$).
They are thus in the same Markov equivalence class, i.e., if a distribution is Markovian w.r.t.\ one of the graphs, then it also is w.r.t.\ the other graph.
Note that the above serves as an example that the Markov condition is not sufficient for causal discovery. Further assumptions are needed, cf.\ below and \cite{Spirtes2000,Pearl2009,PetJanSch17}.

\looseness=-1A graphical model becomes causal if the edges of its graph are causal (in which case the graph is referred to as a ``causal graph''), cf.\ \eq{eq:SA}. This allows to compute interventional distributions as depicted in Figure~\ref{fig_stats_vs_causal}. When a variable is intervened upon, we disconnect it from its parents, fix its value, and perform ancestral sampling on its children. 

\looseness=-1A structural causal model is composed of (i) a set of causal variables and (ii) a set of structural equations with a distribution over the noise variables $U_i$ (or a set of causal conditionals). While both causal graphical models and SCMs allow to compute interventional distributions, only the SCMs allow to compute counterfactuals. To compute counterfactuals, we need to fix the value of the noise variables. Moreover, there are many ways to represent a conditional as a structural assignment (by picking different combinations of functions and noise variables).

\paragraph{Causal Learning and Reasoning} 
The conceptual basis of statistical learning is a joint distribution $P(X_1,\dots,X_n)$ (where often one of the $X_i$ is a response variable denoted as $Y$), and we make assumptions about function classes used to approximate, say, a regression $\E [Y|X]$. {\em Causal learning} considers a richer class of assumptions, and seeks to exploit the fact that the joint distribution possesses a causal factorization \eq{eq:cf}. It involves the causal conditionals $P(X_i \mid \PA_i)$ (e.g., represented by the functions $f_i$ and the distribution of $U_i$ in \eq{eq:SA}), how these conditionals relate to each other, and interventions or changes that they admit. Once a causal model is available, either by external human knowledge or a learning process, {\em causal reasoning} allows to draw conclusions on the effect of interventions, counterfactuals and potential outcomes. In contrast, statistical models only allow to reason about the outcome of i.i.d.\ experiments.

\section{Independent Causal Mechanisms\label{sec:icm}}

We now return to the disentangled factorization \eq{eq:cf} of the joint distribution $P(X_1,\dots,X_n)$. This factorization according to the causal graph is always possible when the $U_i$ are independent, but we will now consider an additional notion of independence relating the factors in \eq{eq:cf} to one another.

Whenever we perceive an object, our brain assumes that the object and the mechanism by which the information contained in its light reaches our brain are {\em independent}. We can violate this by looking at the object from an accidental viewpoint, which can give rise to optical illusions \cite{PetJanSch17}. 
The above independence assumption is useful because in practice, it holds most of the time, and our brain thus relies on objects being independent of our vantage point and the illumination. Likewise, there should not be accidental coincidences, such as 3D structures lining up in 2D, or shadow boundaries coinciding with texture boundaries. In vision research, this is called the generic viewpoint assumption. 

If we move around the object, our vantage point changes, but we assume that the other variables of the overall generative process (e.g., lighting, object position and structure) are unaffected by that. This is an {\em invariance} implied by the above independence, allowing us to infer 3D information even without stereo vision (``structure from motion''). 

\looseness=-1For another example, consider a dataset that consists of altitude $A$ and average annual temperature $T$ of weather stations \citep{PetJanSch17}. $A$ and $T$ are correlated, which we believe is due to the fact that the altitude has a causal effect on temperature. Suppose we had two such datasets, one for Austria and one for Switzerland. The two joint distributions $P(A,T)$ may be rather different since the marginal distributions $P(A)$ over altitudes will differ. The conditionals $P(T|A)$, however, may be (close to) invariant, since they characterize the physical mechanisms that generate temperature from altitude. This similarity is lost upon us if we only look at the overall joint distribution, without information about the causal structure $A\to T$. The causal factorization $P(A)P(T|A)$ will contain a component $P(T|A)$ that generalizes across countries, while the entangled factorization $P(T)P(A|T)$ will exhibit no such robustness.
Cum grano salis, the same applies when we consider interventions in a system. For a model to correctly predict the effect of interventions, it needs to be robust to generalizing from an observational distribution to certain {\em interventional} distributions.

\looseness=-1One can express the above insights as follows \citep{Schoelkopf2012,PetJanSch17}:
\begin{tcolorbox}[colback=black!0!white]
\hypertarget{pri:im}{{\bf Independent Causal Mechanisms (ICM) Principle.}} \hspace{0.1mm}
{\em The causal generative process of a system's variables is composed of autonomous modules that do not inform or influence each other.
In the probabilistic case, this means that the conditional distribution of each variable given its causes (i.e., its mechanism) does not inform or influence the other mechanisms. }
\end{tcolorbox}

\looseness=-1This principle entails several notions important to causality, including separate intervenability of causal variables, modularity and autonomy of subsystems, and invariance \citep{Pearl2009,PetJanSch17}.
If we have only two variables, it reduces to an independence between the cause distribution and the mechanism producing the effect distribution. 

Applied to the causal factorization \eq{eq:cf}, the principle tells us that the factors should be independent in the sense that
\begin{itemize}
\item[(a)] changing (or performing an intervention upon) one mechanism $P(X_i|\PA_i)$ does not change any of the other mechanisms $P(X_j|\PA_j)$ ($i\ne j$) \citep{Schoelkopf2012}, and 
\item[(b)] knowing some other mechanisms $P(X_i|\PA_i)$ ($i\ne j$) does not give us information about a mechanism $P(X_j|\PA_j)$ \citep{JanSch10}. 
\end{itemize}
This notion of independence thus subsumes two aspects: the former pertaining to influence, and the latter to information.

The notion of invariant, autonomous, and independent mechanisms has appeared in various guises throughout the history of causality research~\citep{Haavelmo1944,Frisch1948,Hoover06,Pearl2009,JanSch10,Steudel2010a,PetJanSch17}. Early work on this was done by \citet{Haavelmo1944}, stating the assumption that changing one of the structural assignments leaves the other ones invariant. 
\citet{Hoover06} attributes to Herb Simon the {\em invariance criterion}: the true causal order is the one that is invariant under the right sort of intervention.
\citet{Aldrich89} discusses the historical development of these ideas in economics. He argues that the ``most basic question one can ask about a relation should be: How autonomous is it?'' \citep[][preface]{Frisch1948}. \citet{Pearl2009} discusses autonomy in detail, arguing that a causal mechanism remains invariant when other mechanisms are subjected to external influences. He points out that causal discovery methods may best work ``in longitudinal studies conducted under slightly varying conditions, where accidental independencies are destroyed and only structural independencies are preserved.'' 
Overviews are provided by \citet{Aldrich89,Hoover06,Pearl2009}, and \citet[Sec.~2.2]{PetJanSch17}. These seemingly different notions can be unified~\citep{JanSch10,Steudel2010a}.

\looseness=-1 We view any real-world distribution as a product of causal mechanisms. A change in such a distribution (e.g., when moving from one setting/domain to a related one) will always be due to changes in at least one of those mechanisms. Consistent with the implication (a) of the ICM Principle, we state the following hypothesis:

\begin{tcolorbox}[colback=black!0!white]
\hypertarget{pri:scsh}{{\bf Sparse Mechanism Shift (SMS).}} \hspace{0.1mm}
{\em Small distribution changes tend to manifest themselves in a sparse or local way in the causal/disentangled factorization \eq{eq:cf}, i.e., they should usually not affect all factors simultaneously.}
\end{tcolorbox}

In contrast, if we consider a non-causal factorization, e.g., \eq{eq:non-cf}, then many, if not all, terms will be affected simultaneously as we change one of the physical mechanisms responsible for a system's statistical dependencies. Such a factorization may thus be called {\em entangled}, a term that has gained popularity in machine learning \citep{1206.5538,higgins2016beta,1811.12359,Suter.1811.00007}.

The SMS hypothesis was stated in \citep{ParKilRojSch18,bengio2019meta, 1911.10500,JMLR:v21:19-232}, and in earlier form in \citep{Schoelkopf2012,zhang_domain_2013,SchJanLop16}. An intellectual ancestor is Simon's invariance criterion, i.e., that the causal structure remains invariant across changing background conditions~\citep{Simon53}. 
The hypothesis is also related to ideas of looking for features that vary slowly \citep{foldiak1991learning,Wiskott2002}.
It has recently been used for learning causal models \citep{ke2019learning}, modular architectures \citep{RIMs,BesSunJanSch21} and disentangled representations \citep{locatello2020weakly}.

We have informally talked about the dependence of two mechanisms $P(X_i|\PA_i)$ and $P(X_j|\PA_j)$ when discussing the ICM Principle and the disentangled factorization \eq{eq:cf}.
Note that the dependence of two such mechanisms does {\em not} coincide with the statistical dependence of the random variables $X_i$ and $X_j$. Indeed, in a causal graph, many of the random variables will be dependent even if all mechanisms are independent. Also, the independence of the noise terms $U_i$ does not translate into the independence of the $X_i$.
Intuitively speaking, the independent noise terms $U_i$ provide and parameterize the uncertainty contained in the fact that a mechanism $P(X_i|\PA_i)$ is non-deterministic,\footnote{In the sense that the mapping from $\PA_i$ to $X_i$ is described by a non-trivial conditional distribution, rather than by a function.} and thus ensure that each mechanism adds an independent element of uncertainty. In this sense, the \hyperlink{pri:im}{ICM Principle} contains the independence of the unexplained noise terms in an SCM \eq{eq:SA} as a special case.

In the \hyperlink{pri:im}{ICM Principle}, we have stated that independence of two mechanisms (formalized as conditional distributions) should mean that the two conditional distributions do not {\em inform} or {\em influence} each other. The latter can be thought of as requiring that independent interventions are possible. To better understand the former, we next discuss a formalization in terms of {\em algorithmic independence}. In a nutshell, we encode each mechanism as a bit string, and require that joint compression of these strings does not save space relative to independent compressions.

To this end, first recall that we have so far discussed links between causal and statistical structures. Of the two, the more fundamental one is the causal structure, since it captures the physical mechanisms that generate statistical dependencies in the first place. The statistical structure is an epiphenomenon that follows if we make the unexplained variables random.
It is awkward to talk about statistical information contained in a mechanism since deterministic functions in the generic case neither generate nor destroy information. This serves as a motivation to devise an alternative model of causal structures in terms of Kolmogorov complexity \citep{JanSch10}. The Kolmogorov complexity (or algorithmic information) of a bit string is essentially the length of its shortest compression on a Turing machine, and thus a measure of its information content. Independence of mechanisms can be defined as vanishing mutual algorithmic information; i.e., two conditionals are considered independent if knowing (the shortest compression of) one does not help us achieve a shorter compression of the other.

\looseness=-1Algorithmic information theory provides a natural framework for non-statistical graphical models \citep{JanSch10,JanChaSch16}. Just like the latter are obtained from structural causal models by making the unexplained variables $U_i$ random, we obtain algorithmic graphical models by making the $U_i$ bit strings, jointly independent across nodes, and viewing $X_i$ as the output of a fixed Turing machine running the program $U_i$ on the input $\PA_i$. Similar to the statistical case, one can define a local causal Markov condition, a global one in terms of d-separation, and an additive decomposition of the joint Kolmogorov complexity in analogy to \eq{eq:cf}, and prove that they are implied by the structural causal model \citep{JanSch10}. Interestingly, in this case, independence of noises and independence of mechanisms coincide, since the independent programs play the role of the unexplained noise terms. This approach shows that causality is not intrinsically bound to statistics.

\section{Causal Discovery and Machine Learning}\label{sec:causal_discovery}

Let us turn to the problem of causal discovery from data. Subject to suitable assumptions such as \emph{faithfulness} \citep{Spirtes2000}, one can sometimes recover aspects of the underlying graph\footnote{One can recover the causal structure up to a \textit{Markov equivalence class}, where DAGs have the same undirected skeleton and ``immoralities'' ($X_i \rightarrow X_j \leftarrow X_k$).} from observational data by performing conditional independence tests. However, there are several problems with this approach.
One is that our datasets are always finite  in practice, and conditional independence testing is a notoriously difficult problem, especially if conditioning sets are continuous and multi-dimensional. So while, in principle, the conditional independencies implied by the causal Markov condition hold irrespective of the complexity of the functions appearing in an SCM, for finite datasets, conditional independence testing is hard without additional assumptions \citep{1804.07203}. Recent progress in (conditional) independence testing heavily relies on kernel function classes to represent probability distributions in reproducing kernel Hilbert spaces \citep{Gretton2005,Gretton2005JMLR,Fukumizu2008,Zhang2011uai,DorMuaZhaSch14,PfiBuhSchPet18,1804.02747}.
The other problem is that in the case of only two variables, the ternary concept of conditional independence collapses and the Markov condition thus has no nontrivial implications.

It turns out that both problems can be addressed by making assumptions on function classes. This is typical for machine learning, where it is well-known that finite-sample generalization without assumptions on function classes is impossible. Specifically, although there are universally consistent learning algorithms, i.e., approaching minimal expected error in the infinite sample limit, there are always cases where this convergence is arbitrarily slow. So for a given sample size, it will depend on the problem being learned whether we achieve low expected error, and statistical learning theory provides probabilistic guarantees in terms of measures of complexity of function classes \citep{DevGyoLug96,Vapnik98}.

Returning to causality, we provide an intuition why assumptions on the functions in an SCM should be necessary to learn about them from data. Consider a toy SCM with only two observables $X\to Y$. In this case, \eq{eq:SA} turns into
\begin{align}
X & = U \\
Y & = f(X, V) \label{eq:SA2}
\end{align}
with $U\independent V$.
Now think of $V$ acting as a random selector variable choosing from among a set of functions ${\cal F} = \{ f_v (x) \equiv f(x,v) \; | \; v \in \mbox{supp}(V)\}$. If $f(x,v)$ depends on $v$ in a non-smooth way, it should be hard to glean information about the SCM from a finite dataset, given that $V$ is not observed and its value randomly selects among arbitrarily different $f_v$.

\looseness=-1This motivates restricting the complexity with which $f$ depends on $V$. A natural restriction is to assume an additive noise model
\begin{align}
X & = U \\
Y & = f(X) + V.
\end{align}
If $f$ in \eq{eq:SA2} depends smoothly on $V$, and if $V$ is relatively well concentrated, this can be motivated by a local Taylor expansion argument. It drastically reduces the effective size of the function class --- without such assumptions, the latter could depend exponentially on the cardinality of the support of $V$. 
Restrictions of function classes not only make it easier to learn functions from data, but it turns out that they can break the symmetry between cause and effect in the two-variable case: one can show that given a distribution over $X,Y$ generated by an additive noise model, one cannot fit an additive noise model in the opposite direction (i.e., with the roles of $X$ and $Y$ interchanged) \citep{Hoyer2008,Mooij2009,PetMooJanSch14,Kpotufe14,BauSchPet16}, cf.\ also \citep{Sun2006}. This is subject to certain genericity assumptions, and notable exceptions include the case where $U,V$ are Gaussian and $f$ is linear. It generalizes results of \citet{Shimizu2006} for linear functions, and it can be generalized to include non-linear rescalings \citep{Zhang2009}, loops \citep{Mooij11}, confounders \citep{Janzing2009uai}, and multi-variable settings \citep{Peters2011b}. Empirically, there is a number of methods that can detect causal direction better than chance \citep{Mooijetal16}, some of them building on the above Kolmogorov complexity model \citep{Vreeken}, some on generative models \citep{1711.08936}, and some directly learning to classify bivariate distributions into causal vs.\ anticausal \citep{LopMuaSchTol15}.

\looseness=-1While restrictions of function classes are one possibility to allow to identify the causal structure, other assumptions or scenarios are possible. So far, we have discussed that causal models are expected to generalize under certain distribution shifts since they explicitly model interventions. By the \hyperlink{pri:scsh}{SMS hypothesis}, much of the causal structure is assumed to remain invariant. Hence distribution shifts such as observing a system in different ``environments / contexts'' can significantly help to identify causal structure \citep{Tian2001,PetJanSch17}. These contexts can come from interventions \citep{Schoelkopf2012,peters2016causal, pfister2019learning}, non-stationary time series \citep{hyvarinen2017nonlinear, halva2020hidden,pfister2019invariant} or multiple views \citep{gresele2019incomplete,JMLR:v21:19-232}. The contexts can likewise be interpreted as different tasks, which provide a  connection to meta-learning \citep{bengio1990learning, finn2017model,schmidhuber1987evolutionary}. 

\looseness=-1The work of \citet{bengio2019meta} ties the generalization in  meta-learning to invariance properties of causal models, using the idea that a causal model should adapt faster to interventions than purely predictive models. 
This was extended to multiple variables and unknown interventions in \citep{ke2019learning}, proposing a framework for causal discovery using neural networks by turning the discrete graph search into a continuous optimization problem.
While \citep{bengio2019meta,ke2019learning} focus on learning a causal model using neural networks with an unsupervised loss, the work of \citet{dasgupta2019causal} explores learning a causal model using a reinforcement learning agent. These approaches have in common that semantically meaningful abstract representations are given and do not need to be learned from high-dimensional and low-level (e.g., pixel) data.

\begin{figure*}[tbh]
\begin{center}
\includegraphics[width=0.8\textwidth]{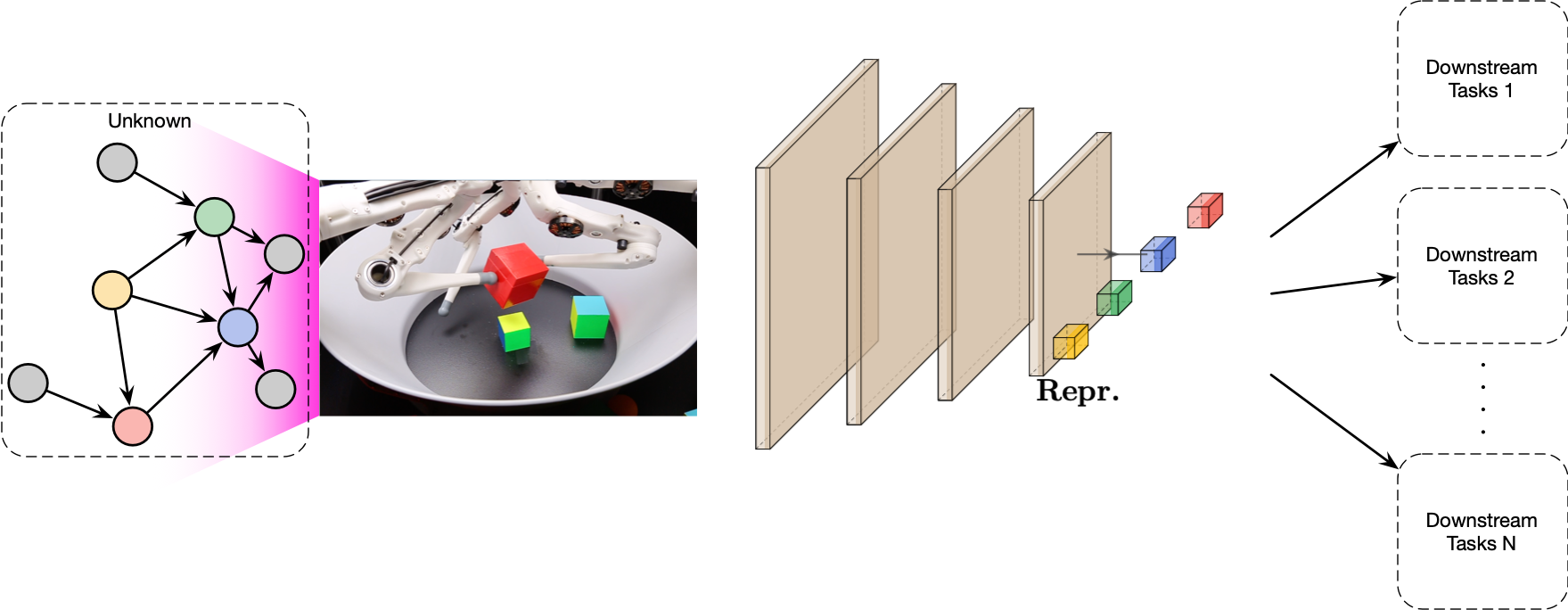}
\end{center}
  \caption{\label{fig_causal_rep} Illustration of the causal representation learning problem setting. Perceptual data, such as images or other high-dimensional sensor measurements, can be thought of as entangled views of the state of an unknown causal system as described in \eq{eq:causal_rep_learning}. With the exception of possible task labels, none of the variables describing the causal variables generating the system may be known. The goal of causal representation learning is to learn a representation (partially) exposing this unknown causal structure (e.g., which variables describe the system, and their relations). As full recovery may often be unreasonable, neural networks may map the low-level features to some high-level variables supporting causal statements relevant to a set of downstream tasks of interest. For example, if the task is to detect the manipulable objects in a scene, the representation may separate intrinsic object properties from their pose and appearance to achieve robustness to distribution shifts on the latter variables. 
  Usually, we do not get labels for the high-level variables, but the properties of causal models can serve as useful inductive biases for learning (e.g., the \protect\hyperlink{pri:scsh}{SMS hypothesis}).}
\end{figure*}
\begin{figure}[tbh]
\begin{center}
\includegraphics[width=0.4\textwidth]{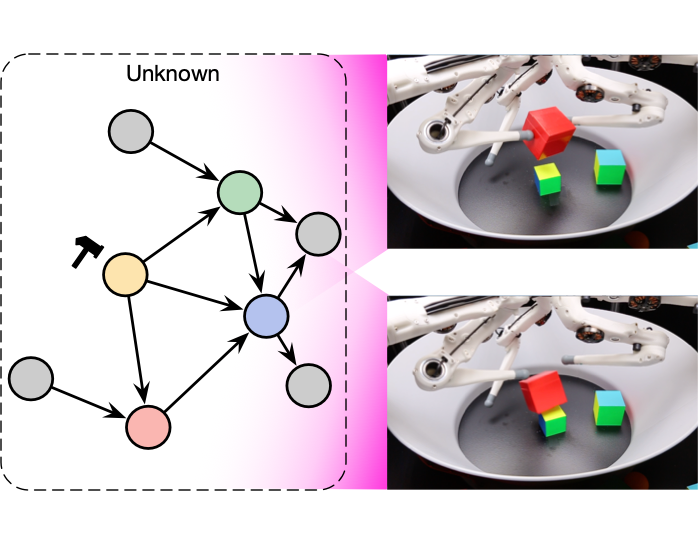}
\end{center}
\vspace{-5mm}
  \caption{\label{fig_causal_shifts} Example of the \protect\hyperlink{pri:scsh}{SMS hypothesis} where an intervention (which may or may not be intentional/observed) changes the position of one finger (\myhammer{}), and as a consequence, the object falls. The change in pixel space is entangled (or distributed), in contrast to 
  the change in the causal model.
  }
\end{figure}

\section{Learning Causal Variables}\label{sec:learning_variables}
\looseness=-1Traditional causal discovery and reasoning assume that the units are random variables connected by a causal graph. However, real-world observations are usually not structured into those units to begin with, e.g., objects in images \citep{LopNisChiSchBot17}. Hence, the emerging field of causal representation learning strives to learn these variables from data, much like machine learning went beyond symbolic AI in not requiring that the symbols that algorithms manipulate be given a priori (cf.\ \citet{geffner}). To this end, we could try to connect causal variables $S_1,\ldots, S_n$ to observations
\begin{equation}\label{eq:causal_rep_learning}
    X = G(S_1,\ldots,S_n),
\end{equation}
where G is a non-linear function. An example can be seen in Figure~\ref{fig_causal_rep}, where high-dimensional observations are the result of a view on the state of a causal system that is then processed by a neural network to extract high-level variables that are useful on a variety of tasks. Although causal models in economics, medicine, or psychology often use variables that are abstractions of underlying quantities, it is challenging to state general conditions under which coarse-grained variables admit causal models with well-defined interventions \citep{1512.07942,Rubensteinetal17}.
Defining objects or variables that can be causally related amounts to coarse-graining of more detailed models of the world, including microscopic structural equation models~\citep{Rubensteinetal17}, ordinary differential equations~\citep{MooijJ2013,RubBonMooSch18}, and temporally aggregated time series~\citep{Gongetal17}.  
The task of identifying suitable units that admit causal models is challenging for both human and machine intelligence. Still, it aligns with the general goal of modern machine learning to learn meaningful representations of data, where meaningful can include \emph{robust}, \emph{explainable}, or \emph{fair} \citep{NIPS2017_6995,Kilbertusetal17,ZhaBar18,karimi2020algorithmic,vonkugelgen2020fairness}.

To combine structural causal modeling \eq{eq:SA} and representation learning, we should strive to embed an SCM into larger machine learning models whose inputs and outputs may be high-dimensional and unstructured, but whose inner workings are at least partly governed by an SCM (that can be parameterized with a neural network). The result may be a modular architecture, where the different modules can be individually fine-tuned and re-purposed for new tasks \citep{ParKilRojSch18,RIMs} and the SMS hypothesis can be used to enforce the appropriate structure. 
We visualize an example in Figure~\ref{fig_causal_shifts} where changes are sparse for the appropriate causal variables (the position of the finger and the cube changed as a result of moving the finger), but dense in other representations, for example in the pixel space (as finger and cube move, many pixels change their value). At the extreme, all pixels may change as a result of a sparse intervention, for example, if the camera view or the lighting changes.

We now discuss three problems of modern machine learning in the light of causal representation learning.
\paragraph{Problem 1 -- Learning Disentangled Representations}
We have earlier discussed the \hyperlink{pri:im}{ICM Principle} implying both the independence of the SCM noise terms in \eq{eq:SA} and thus the feasibility of the disentangled representation
\begin{equation}\label{eq:cf2}
P(S_1,\dots,S_n) = \prod_{i=1}^n  P(S_i \mid \PA_i)
\end{equation}
as well as the property that the conditionals $ P(S_i \mid \PA_i)$ be independently manipulable and largely invariant across related problems. Suppose we seek to reconstruct such a {\em disentangled representation using independent mechanisms} \eq{eq:cf2} from data, but the causal variables $S_i$ are not provided to us a priori. Rather, we are given (possibly high-dimensional) $X=(X_1,\dots,X_d)$ (below, we think of $X$ as an image with pixels $X_1,\dots,X_d$) as in \eq{eq:causal_rep_learning}, from which we should construct causal variables $S_1,\dots,S_n$ ($n\ll d$) as well as mechanisms, cf.\ \eq{eq:SA},
\begin{equation}\label{eq_disent}
S_i := f_i (\PA_i, U_i),   ~~~~ (i=1,\dots,n),
\end{equation}
modeling the causal relationships among the $S_i$ .
To this end, as a first step, we can use an {\em encoder} $q:\R^d \to \R^n$ taking $X$ to a latent ``bottleneck'' representation comprising the unexplained noise variables  $U=(U_1,\dots,U_n)$.
The next step is the mapping $f(U)$ determined by the structural assignments $f_1,\dots,f_n$.
Finally, we apply a {\em decoder} $p:\R^n \to \R^d$. For suitable $n$, the system can be trained using reconstruction error to satisfy $p \circ f \circ q\approx id$ on the observed images. If the causal graph is known, the topology of a neural network implementing $f$ can be fixed accordingly; if not, the neural network decoder learns the composition $\tilde{p} = p\circ f$. In practice, one may not know $f$, and thus only learn an autoencoder $\tilde{p}\circ q$, where the causal graph effectively becomes an unspecified part of the decoder $\tilde{p}$, possibly aided by a suitable choice of architecture \citep{Leeb-SAE}.

Much of the existing work on disentanglement \citep{higgins2016beta,1811.12359,locatello2020weakly,van2019disentangled,locatello2019fairness,kim2018disentangling,ridgeway2018learning,eastwood2018framework} focuses on independent factors of variation. This can be viewed as the special case where the causal graph is trivial, i.e., $\forall i: \PA_i = \emptyset$ in \eq{eq_disent}. In this case, the factors are functions of the independent exogenous noise variables, and thus independent themselves.\footnote{For an example to see why this is often not desirable, note that the presence of fork and knife may be statistically dependent, yet we might want a disentangled representation to represent them as separate entities.} However, the \hyperlink{pri:im}{ICM Principle} is more general and contains statistical independence as a special case. 

Note that the problem of {\em object-centric} representation learning ~\citep{bapst2019structured,burgess2019monet,goyal2020object,greff2019multi,greff2020binding,kosiorek2018sequential,lin2019space,locatello2020object,Julius_ECON,van2018relational} can also be considered a special case of disentangled factorization as discussed here. Objects are constituents of scenes that in principle permit separate interventions. A disentangled representation of a scene containing objects should probably use objects as some of the building blocks of an overall causal factorization\footnote{Objects can be represented at different levels of granularity~\citep{Rubensteinetal17}, i.e. as a single entity or as a composition of other causal variables encoding parts, properties, and other factors of variation.
}, complemented by mechanisms such as orientation, viewing direction, and lighting.

The problem of recovering the exogenous noise variables 
is ill-defined in the i.i.d.\ case as there are infinitely many equivalent solutions yielding the same observational distribution~\citep{1811.12359,hyvarinen1999nonlinear,PetJanSch17}. 
Additional assumptions or biases can help favoring certain solutions over others ~\citep{1811.12359,rolinek2019variational}. \citet{Leeb-SAE} propose a structured decoder that embeds an SCM and automatically learns a hierarchy of disentangled factors.

To make \eq{eq_disent} causal, we can use the \hyperlink{pri:im}{ICM Principle}, i.e., we should make the $U_i$ statistically independent, and we should make the mechanisms independent.
This could be done by ensuring that they are invariant across problems, exhibit sparse changes to actions, or that they can be independently intervened upon~\citep{1911.10500,1703.07718,1812.03253}.
~\citet{locatello2020weakly} showed that the sparse mechanism shift hypothesis stated above is theoretically sufficient when given suitable training data. Further, the SMS hypothesis can be used as supervision signal in practice even if $\PA_i \neq \emptyset$~\citep{trauble2020independence}. However, which factors of variation can be disentangled depend on which interventions can be observed~\citep{shu2019weakly,locatello2020weakly}. As discussed by \citet{SchJanLop16,shu2019weakly}, different supervision signals 
may be used to identify subsets of factors. Similarly, when learning causal variables from data, which variables can be extracted and their granularity depends on which distribution shifts, explicit interventions, and other supervision signals are available.

\paragraph{Problem 2 -- Learning Transferable  Mechanisms}
\looseness=-1An artificial or natural agent in a complex world is faced with limited resources. This concerns training data, i.e., we only have limited data for each task/domain, and thus need to find ways of pooling/re-using data, in stark contrast to the current industry practice of large-scale labeling work done by humans. It also concerns computational resources: animals have constraints on the size of their brains, and evolutionary neuroscience knows many examples where brain regions get re-purposed. Similar constraints on size and energy apply as ML methods get embedded in (small) physical devices that may be battery-powered.
Future AI models that robustly solve a range of problems in the real world will thus likely need to re-use components, which requires them to be robust across tasks and environments \citep{SchJanLop16}. 
An elegant way to do this is to employ a modular structure that mirrors a corresponding modularity in the world. In other words, if the world is indeed modular, in the sense that components/mechanisms of the world play roles across a range of environments, tasks, and settings, then it would be prudent for a model to employ corresponding modules \citep{RIMs}. For instance, if variations of natural lighting (the position of the sun, clouds, etc.) imply that the visual environment can appear in brightness conditions spanning several orders of magnitude, then visual processing algorithms in our nervous system should employ methods that can factor out these variations, rather than building separate sets of face recognizers, say, for every lighting condition. If, for example, our nervous system were to compensate for the lighting changes by a gain control mechanism, then this mechanism in itself need not have anything to do with the physical mechanisms bringing about brightness differences. However, it would play a role in a modular structure that corresponds to the role that the physical mechanisms play in the world's modular structure. This could produce a bias towards models that exhibit certain forms of structural homomorphism to a world that we cannot directly recognize, which would be rather intriguing, given that ultimately our brains do nothing but turn neuronal signals into other neuronal signals.
A sensible inductive bias to learn such models is to look for independent causal mechanisms \citep{ParRojKilSch17} and competitive training can play a role in this. For pattern recognition tasks, \citep{ParKilRojSch18,RIMs} suggest that learning causal models that contain independent mechanisms may help in transferring modules across substantially different domains.

\paragraph{Problem 3 -- Learning Interventional World Models and Reasoning}
Deep learning excels at learning representations of data that preserve relevant statistical properties \citep{1206.5538,LeCBenHin15}. However, it does so without taking into account the causal properties of the variables, i.e., it does not care about the interventional properties of the variables it analyzes or reconstructs. Causal representation learning should move beyond the representation of statistical dependence structures towards models that support intervention, planning, and reasoning, realizing Konrad Lorenz' notion of \emph{thinking} as \emph{acting in an imagined space} \cite{Lorenz73}.
This ultimately requires the ability to reflect back on one's actions and envision alternative scenarios, possibly necessitating (the illusion of) free will \citep{Pearl2009forbes}. The biological function of self-consciousness may be related to the need for a variable representing oneself in one's Lorenzian {\em imagined space}, and free will may then be a means to communicate about actions taken by that variable, crucial for social and cultural learning, a topic which has not yet entered the stage of machine learning research although it is at the core of human intelligence \citep{Henrich}.

\section{Implications for Machine Learning}\label{sec:benefits}
\looseness=-1All of this discussion calls for a learning paradigm that does not rest on the usual i.i.d.\ assumption. Instead, we wish to make a weaker assumption: that the data on which the model will be applied comes from a possibly different distribution, but involving (mostly) the same causal mechanisms \citep{PetJanSch17}. 
This raises serious challenges: (a) in many cases, we need to infer abstract causal variables from the available low-level input features; (b) there is no consensus on which aspects of the data reveal causal relations; (c) the usual experimental protocol of training and test set may not be sufficient for inferring and evaluating causal relations on existing data sets, and we may need to create new benchmarks, for example with access to environment information and interventions; (d) even in the limited cases we understand, we often lack scalable and numerically sound algorithms. Despite these challenges, we argue this endeavor has concrete implications for machine learning and may shed light on desiderata and current practices alike.

\subsection{Semi-Supervised Learning (SSL)}
Suppose our underlying causal graph is $X\to Y$, and at the same time we are trying to learn a mapping $X\to Y$. The causal factorization \eq{eq:cf} for this case is 
\begin{equation}
P(X,Y) = P(X) P(Y|X).
\end{equation}
The \hyperlink{pri:im}{ICM Principle} posits that the modules in a joint distribution's causal decomposition do not inform or influence each other.
This means that in particular, $P(X)$ should contain no information about $P(Y|X)$, which implies that SSL should be futile, in as far as it is using additional information about $P(X)$ (from unlabelled data) to improve our estimate of $P(Y|X=x)$.

In the opposite ({\em anticausal}) direction (i.e., the direction of prediction is opposite to the causal generative process), however, SSL may be possible. To see this, we refer to \citet{Daniusisetal10} who define a measure of dependence between input $P(X)$ and  conditional $P(Y|X)$.\footnote{Other dependence measures have been proposed for high-dimensional linear settings and time series \citep{JanHoySch10,Shajarisales15,BesShaSchJan18,JanSch18b,Janzing_NIPS2019,Janzingetal12}.}
Assuming that this measure is zero in the causal direction (applying the ICM assumption described in Section~\ref{sec:icm} to the two-variable case), they show that it is strictly positive in the anticausal direction. 
Applied to SSL in the anticausal direction, this implies that the distribution of the input (now: effect) variable should contain information about the conditional of output (cause) given input, i.e., the quantity that machine learning is usually concerned with.

The study \citep{Schoelkopf2012} empirically corroborated these predictions, thus establishing an intriguing bridge between the {\em structure} of learning problems and certain {\em physical} properties (cause-effect direction) of real-world data generating processes. It also led to a range of follow-up work \citep{zhang_domain_2013,WeiSchBalGro14,zhang_multi-source_2015,GonZhaLiuTaoSch16,HuaZhaZhaSanGlySch17,Zhangetal17,1610.03263,1809.09337,1903.06256,1812.04597,1810.11953,1807.08479,1802.03916,Li_2018_ECCV,1707.06422,RojSchTurPet18,JMLR:v21:19-232}, complementing the studies of Bareinboim and Pearl \citep{Bareinboim2014,1503.01603}, and it inspired a thread of work in the statistics community exploiting invariance for causal discovery and other tasks \citep{peters2016causal,pfister2019learning,1706.08576,1710.11469,JMLR:v21:19-232}.

On the SSL side, subsequent developments include further theoretical analyses \citep[Section~5.1.2]{JanSch15,PetJanSch17} and a form of conditional SSL \citep{KugMeyLooSch19}. 
The view of SSL as exploiting dependencies between a marginal $P(X)$ and a non-causal conditional $P(Y|X)$ is consistent with the common assumptions employed to justify SSL \citep{ChaSchZie06}. The \emph{cluster assumption} asserts that the labeling function (which is a property of $P(Y|X)$) should not change within clusters of $P(X)$. The \emph{low-density separation assumption} posits that the area where $P(Y|X)$ takes the value of $0.5$ should have small $P(X)$; and the \emph{semi-supervised smoothness assumption}, applicable also to continuous outputs, states that if two points in a high-density region are close, then so should be the corresponding output values. Note, moreover, that some of the theoretical results in the field use assumptions well-known from causal graphs (even if they do not mention causality): the {\em co-training theorem} \citep{BluMit98} makes a statement about learnability from unlabelled data, and relies on an assumption of predictors being conditionally independent given the label, which we would normally expect if the predictors are (only) caused by the label, i.e., an anticausal setting. This is nicely consistent with the above findings.

\subsection{Adversarial Vulnerability}
\looseness=-1One can hypothesize that the causal direction should also have an influence on whether classifiers are vulnerable to \emph{adversarial attacks}. These attacks have recently become popular, and consist of minute changes to inputs, invisible to a human observer yet changing a classifier's output \citep{1312.6199}. 
This is related to causality in several ways. First, these attacks clearly constitute violations of the i.i.d.\ assumption that underlies statistical machine learning. If all we want to do is a prediction in an i.i.d.\ setting, then statistical learning is fine.
In the adversarial setting, however, the modified test examples are not drawn from the same distribution as the training examples. The adversarial phenomenon also shows that the kind of robustness current classifiers exhibit is rather different from the one a human exhibits. If we knew both robustness measures, we could try to maximize one while minimizing the other. Current methods can be viewed as crude approximations to this, effectively modeling the human's robustness as a mathematically simple set, say, an $l_p$ ball of radius $\epsilon>0$: they often try to find examples which lead to maximal changes in the classifier's output, subject to the constraint that they lie in an $l_p$ ball in the pixel metric.  As we think of a classifier as the approximation of a function, the large gradients exploited by these attacks are either a property of this function or a defect of the approximation.

\looseness=-1There are different ways of relating this to causal models. As described in \cite[Section 1.4]{PetJanSch17}, different causal  models can generate the same statistical pattern recognition model. In one of those, we might provide a writer with a sequence of class labels $y$, with the instruction to produce a set of corresponding images $x$. Clearly, intervening on $y$ will impact $x$, but intervening on $x$ will not impact $y$, so this is an anticausal learning problem. In another setting, we might ask the writer to decide for herself which digits to write, and to record the labels alongside the digit (in this case, the classifier would try to predict one effect from another one, a situation which we might call a confounded one). In a last one, we might provide images to a person, and ask the person to generate labels by classifying them.

Let us now assume that we are in the {\em causal} setting where the causal generative model factorizes into independent components, one of which is (essentially) the classification function.
As discussed in Section~\ref{sec:causal_models}, when specifying a causal model, one needs to determine which interventions are allowed, and a structural assignment will then, by definition, be valid under every possible (allowed) intervention. 
One may thus expect that if the predictor approximates the causal mechanism
that is inherently transferable and robust, adversarial examples should be harder to find \citep{Schoelkopf2017icml,KilParSch19arxiv}.\footnote{Adversarial attacks may still exploit the quality of the (parameterized) approximation of a structural equation.} Recent work supports this view: it was shown that a possible defense against adversarial attacks is to solve the anticausal classification problem by modeling the causal generative direction, a method which in vision is referred to as {\em analysis by synthesis} \citep{schott2018towards}. A related defense method proceeds by reconstructing the input using an autoencoder before feeding it to a classifier \citep{DBR}.

\subsection{Robustness and Strong Generalization}\label{sec:robustness}
We can speculate that structures composed of autonomous modules, such as given by a causal factorization \eq{eq:cf}, should be relatively robust to swapping out or modifying individual components.
Robustness should also play a role when studying {\em strategic behavior}, i.e., decisions or actions that take into account the actions of other agents (including AI agents). Consider a system that tries to predict the probability of successfully paying back a credit, based on a set of features. The set could include, for instance, the current debt of a person, as well as their address. To get a higher credit score, people could thus change their current debt (by paying it off), or they could change their address by moving to a more affluent neighborhood. The former probably has a positive causal impact on the probability of paying back; for the latter, this is less likely. Thus, we could build a scoring system that is more robust with respect to such strategic behavior by only using causal features as inputs \citep{1905.09239}.

To formalize this general intuition, one can consider a form of out-of-distribution generalization, which can be optimized by minimizing the empirical risk over a class of distributions induced by a causal model of the data~\citep{arjovsky2019invariant,RojSchTurPet18,meinshausen2018causality,peters2016causal,Schoelkopf2012}.
To describe this notion, we start by recalling the usual empirical risk minimization setup. We have access to data from a distribution $P(X,  Y)$ and train a predictor $g$ in a hypothesis space $\mathcal{H}$ (e.g., a neural network with a certain architecture predicting $Y$ from $X$) to minimize the empirical risk $\hat{R}$
\begin{equation}
g^\star = \argmin_{g\in\mathcal{H}}\hat{R}_{P(X, Y)}(g)
\end{equation}
where
\begin{equation}
\hat{R}_{P(X, Y)}(g) = \hat{\mathbb{E}}_{P(X, Y)} \left[ \text{loss}(Y, g(X))\right].
\end{equation}
Here, we denote by $\hat{\mathbb{E}}_{P(X, Y)}$ the empirical mean computed from a sample drawn from $P(X, Y)$.
When we refer to ``out-of-distribution generalization'' we mean having a small expected risk for a different distribution $P^{\dagger}(X, Y)$:
\begin{equation}
R^{OOD}_{P^{\dagger}(X, Y)}(g) = \mathbb{E}_{P^{\dagger}(X, Y)}\left[\text{loss}(Y, g(X))\right].
\end{equation}
Clearly, the gap between $\hat{R}_{P(X, Y)}(g)$ and $R^{OOD}_{P^{\dagger}(X, Y)}(g)$ will depend on how different the test distribution $P^{\dagger}$ is from the training distribution $P$. To quantify this difference, we call \textit{environments} the collection of different circumstances that give rise to the distribution shifts such as locations, times, experimental conditions, etc. Environments can be modeled in a causal factorization~\eqref{eq:cf} as they can be seen as interventions on one or several causal variables or mechanisms. As a motivating example, one environment may correspond to \emph{where} a measurement is taken (for example a certain room), and from each environment, we obtain a collection of measurements (images of objects in the same room). It is nontrivial (and in some cases provably hard \citep{ben2010impossibility}) to learn statistical models that are stable across training environments and generalize to novel testing environments~\citep{peters2016causal,RojSchTurPet18,1707.06422,arjovsky2019invariant,akkaya2019solving} drawn from the same environment distribution.

Using causal language, one could restrict $P^{\dagger}(X, Y)$ to be the result of a certain set of interventions, i.e., $P^{\dagger}(X, Y)\in \mathbb{P}_\mathcal{G}$ where $\mathbb{P}_\mathcal{G}$ is a set of interventional distributions over a causal graph $\mathcal{G}$. The worst case out-of-distribution risk then becomes
\begin{equation}\label{eq_robust_gen}
R^{OOD}_{\mathbb{P}_\mathcal{G}}(g) = \max_{P^{\dagger}\in\mathbb{P}_\mathcal{G}}\mathbb{E}_{P^{\dagger}(X, Y)} \left[\text{loss}(Y, g(X))\right].
\end{equation}
To learn a robust predictor, we should have available a subset of environment distributions $\mathcal{E}\subset \mathbb{P}_\mathcal{G}$
and solve
\begin{equation}\label{eq_robust_pred}
g^\star = \argmin_{g\in\mathcal{H}}\max_{P^{\dagger}\in \mathcal{E}}\hat{\mathbb{E}}_{P^{\dagger}(X, Y)} \left[\text{loss}(Y, g(X))\right].
\end{equation}
In practice, solving \eq{eq_robust_pred} requires specifying a causal model with an associated set of interventions. If the set of observed environments $\mathcal{E}$ does not coincide with the set of possible environments $\mathbb{P}_\mathcal{G}$, we have an additional estimation error that may be arbitrarily large in the worst case \citep{arjovsky2019invariant,ben2010impossibility}.

\subsection{Pre-training, Data Augmentation, and Self-Supervision}
Learning predictive models solving the min-max optimization problem of \eq{eq_robust_pred} is challenging. We now interpret several common techniques in Machine Learning as means of approximating \eq{eq_robust_pred}.

The first approach is enriching the distribution of the training set. This does not mean obtaining more examples from $P(X, Y)$, but  training on a richer dataset~\citep{sun2017revisiting,deng2009imagenet}, for example, through pre-training on a huge and diverse corpus~\citep{radford2018improving, devlin2018bert, howard2018universal, kolesnikov2019big,djolonga2020robustness, brown2020language, chen2020generative,tschannen2020self}. Since this strategy is based on standard empirical risk minimization, it can achieve stronger generalization in practice only if the new training distribution is sufficiently diverse to contain information about other distributions in $\mathbb{P}_\mathcal{G}$.

The second approach, often coupled with the previous one, is to rely on data augmentation to increase the diversity of the data by ``augmenting'' it through a certain type of artificially generated interventions \citep{Baird90,simard2003best, krizhevsky2012imagenet}. For the visual domain, common augmentations include performing transformations such as rotating the image, translating the image by a few pixels, or flipping the image horizontally, etc. The high-level idea behind data augmentation is to encourage a system to learn underlying invariances or symmetries present in the augmented data distribution. For example, in a classification task, translating the image by a few pixels does not change the class label. One may view it as specifying a set of interventions $\mathcal{E}$ the model should be robust to (e.g., random crops/interpolations/translation/rotations, etc). Instead of computing the maximum over all distributions in $\mathcal{E}$, one can relax the problem by sampling from the interventional distributions and optimize an expectation over the different augmented images on a suitably chosen subset \citep{BurSch97}, using a search algorithm like reinforcement learning \citep{cubuk2019autoaugment} or an algorithm based on density matching \citep{lim2019fast}.

\looseness=-1The third approach is to rely on self-supervision to learn about $P(X)$. Certain pre-training methods ~\citep{radford2018improving, devlin2018bert, howard2018universal, brown2020language, chen2020generative,tschannen2020self} have shown that it is possible to achieve good results using only very few class labels by first pre-training on a large unlabeled dataset and then fine-tuning on few labeled examples. Similarly, pre-training on large unlabeled image datasets can improve performance by learning representations that can efficiently transfer to a downstream task, as demonstrated by \citep{oord2018representation, bachman2019learning, he2020momentum, chen2020simple, grill2020bootstrap}. These methods fall under the umbrella of self-supervised learning, a family of techniques for converting an unsupervised learning problem into a supervised one by using so-called pretext tasks with artificially generated labels without human annotations. The basic idea behind using pretext tasks is to force the learner to  learn representations that contain information about $P(X)$ that may be useful for (an unknown) downstream task. Much of the work on methods that use self-supervision relies on carefully constructing pretext tasks. A central challenge here is to extract features that are indeed informative about the data generating distribution. Ideas from the \hyperlink{pri:im}{ICM Principle} could help develop methods that can automate the process of constructing pretext tasks. Finally, one can explicitly optimize \eq{eq_robust_pred}, for example, through adversarial training~\citep{goodfellow2014explaining}. In that case, $\mathbb{P}_\mathcal{G}$ would contain a set of attacks an adversary might perform, while presently, we consider a set of natural interventions.

An interesting research direction is the combination of all these techniques, large scale training, data augmentation, self-supervision, and robust fine-tuning on the available data from multiple, potentially simulated environments.

\subsection{Reinforcement Learning}
Reinforcement Learning (RL) is closer to causality research than the machine learning mainstream in that it sometimes effectively directly estimates do-probabilities. E.g., on-policy learning estimates do-probabilities for the interventions specified by the policy (note that these may not be hard interventions if the policy depends on other variables). However, as soon as off-policy learning is considered, in particular in the batch (or observational) setting \citep{Lange2012}, issues of causality become subtle \citep{1812.10576,1805.12298}. An emerging line of work devoted to the intersection of RL and causality includes \citep{Bareinboim2015,1703.07718,1812.10576,1811.06272,dasgupta2019causal,Bareinboim_NIPS2019,ahmed2021causalworld}. Causal learning applied to reinforcement learning can be divided into two aspects, causal induction and causal inference. \emph{Causal induction (discovery)} involves learning causal relations from data, for example, an RL agent learning a causal model of the environment. \emph{Causal inference} learns to plan and act based on a causal model. Causal induction in an RL setting poses different challenges than the classic causal learning settings where the causal variables are often given. However, there is accumulating evidence supporting the usefulness of an appropriate structured representation of the environment~\citep{akkaya2019solving,berner2019dota,vinyals2019grandmaster}. 
\vspace{2mm}
\paragraph{World Models} \looseness=-1 Model-based RL~\citep{sutton1998introduction,finn2017model} is related to causality as it aims at modeling the effect of actions (interventions) on the current state of the world. 
Particularly relevant for causal leaning are generative world models that capture some of the causal relations underlying the environment and serve as Lorenzian imagined spaces (see {\sc Introduction} above) to train RL agents~\citep{kaelbling1996reinforcement,sutton1998introduction,ha2018world,chiappa2017recurrent,xie2016model,oh2015action,silver2017predictron,schmidhuber1991curious,wiering2012reinforcement}.
Structured generative approaches further aim at decomposing an environment into multiple entities with causally correct relations among them, modulo the completeness of the variables, and confounding~\citep{diuk2008object,watters2019cobra,chang2016compositional,watters2017visual,battaglia2016interaction,kipf2018neural}. However, many of the current approaches (regardless of structure), only build partial models of the environment \citep{gregor2019shaping}. Since they do not observe the environment at every time step, the environment may become an unobserved confounder affecting both the agent's actions and the reward. To address this issue, a model can use the backdoor criterion conditioning on its policy~\citep{rezende2020causally}.

\paragraph{Generalization, Robustness, and Fast Transfer} \looseness=-1
While RL has already achieved impressive results, the sample complexity required to achieve consistently good performance is often prohibitively high. Further, RL agents are often brittle (if data is limited) in the face of even tiny changes to the environment (either visual or mechanistic changes) unseen in the training phase. The question of generalization in RL is essential to the field’s future both in theory and practice. One proposed solution towards the goal of designing machines that can extrapolate experience across environments and tasks is to learn invariances in a causal graph structure. A key requirement to learn invariances from data may be the possibility to perform and learn from interventions. Work in developmental psychology argues that there is a need to experiment in order to discover causal relationships \citep{gopnik2004theory}. This can be modelled as an RL environment, where the agent can discover causal factors through interventions and observing their effects.
Further, causal models may allow to model the environment as a set of underlying independent causal mechanisms such that, if there is a change in distribution, not all the mechanisms need to be re-learned. However, there are still open questions about the right way to think about generalization in RL, the right way to formalize the problem, and the most relevant tasks. 

\paragraph{Counterfactuals}\looseness=-1 Counterfactual reasoning has been found to improve the data efficiency of RL algorithms \citep{1811.06272,2012.09092}, improve performance \citep{dasgupta2019causal}, and it has been applied to communicate about past experiences in the multi-agent setting \citep{foerster2018counterfactual,su2020counterfactual}. These findings are consistent with work in cognitive psychology \citep{epstude2008functional}, arguing that counterfactuals allow to reason about the usefulness of past actions and transfer these insights to corresponding behavioral intentions in future scenarios \citep{roese1994functional,reichert1999reflective,landman1995missed}. 

We argue that future work in RL should consider counterfactual reasoning as a critical component to enable acting in imagined spaces and formulating hypotheses that can be subsequently tested with suitably chosen interventions. 

\paragraph{Offline RL}\looseness=-1 The success of deep learning methods in the case of supervised learning can be largely attributed to the availability of large datasets and methods that can scale to large amounts of data. In the case of reinforcement learning, collecting large amounts of high-fidelity diverse data from scratch can be expensive and hence becomes a bottleneck. Offline RL \citep{fujimoto2019off, levine2020offline} tries to address this concern by learning a policy from a \textit{fixed} dataset of trajectories, without requiring any experimental or interventional data (i.e., without any interaction with the environment). The effective use of observational data (or logged data) may make  real-world RL more practical by incorporating diverse prior experiences. To succeed at it, an agent should be able to infer the consequence of different sets of actions compared to those seen during training (i.e., the actions in the logged data), which essentially makes it a counterfactual inference problem. The distribution mismatch between the current policy and the policy that was used to collect offline data makes  offline RL  challenging as this requires us to move well beyond the assumption of independently and identically distributed data. Incorporating invariances, by factorizing knowledge in terms of independent causal mechanisms can help make progress towards the offline RL setting.   

\subsection{Scientific Applications}
A fundamental question in the application of machine learning in natural sciences is to which extent we can complement our understanding of a physical system with machine learning. One interesting aspect is physics simulation with neural networks \citep{grzeszczuk1998neuroanimator}, which can substantially increase the efficiency of hand-engineered simulators \citep{he2019learning,ladicky2015data,wiewel2019latent,sanchez2020learning,watters2017visual}. Significant out-of-distribution generalization of learned physical simulators may not be necessary if experimental conditions are carefully controlled, although the simulator has to be completely re-trained if the conditions change. 

\looseness=-1On the other hand, the lack of systematic experimental conditions may become problematic in other applications such as healthcare.
One example is personalized medicine, where we may wish to build a model of a patient health state through a multitude of data sources, like electronic health records and genetic information \citep{esteva2019guide,henry2015targeted}. However, if we train a clinical system on doctors’ actions in controlled settings, the system will likely provide little additional insight compared to the doctors’ knowledge and may fail in surprising ways when deployed \citep{beede2020human}. While it may be useful to automate certain decisions, an understanding of causality may be necessary to recommend treatment options that are personalized and reliable \citep{Richens2020vs,subbaswamy2018counterfactual,schulam2017reliable,yoon2018ganite,atan2018deep,alaa2018limits,bica2019time,2012.09092}. 

Causality also has significant potential in helping understand medical phenomena, e.g., in the current Covid-19 pandemic, where causal mediation analysis helps disentangle different effects contributing towards case fatality rates when a textbook example of Simpson's paradox was observed \citep{vonkugelgen2020simpsons}.

Another example of a scientific application is in astronomy, where causal models were used to identify exoplanets under the confounding of the instrument. Exoplanets are often detected as they partially occlude their host star when they transit in front of it, causing a slight decrease in brightness. Shared patterns in measurement noise across stars light-years apart can be removed in order to reduce the instrument's influence on the measurement \citep{Scholkopfetal16}, which is critical especially in the context of partial technical failures as experienced in the Kepler exoplanet search mission. 
The application of \citep{Scholkopfetal16} lead to the discovery of 36 planet candidates \citep{Foreman-Mackeyetal15}, of which 21 were
subsequently validated as bona fide exoplanets \citep{Montet_2015}.
Four years later, astronomers found traces of water in the atmosphere of the exoplanet K2-18b --- the first such discovery for an exoplanet in the habitable zone, i.e., allowing for liquid water \citep{1909.04642,Tsiaras}. This planet turned out to be one that had first been detected in \citep[exoplanet candidate EPIC 201912552]{Foreman-Mackeyetal15}.

\subsection{Multi-Task Learning and Continual Learning}

State-of-the-art AI is relatively {\em narrow}, i.e.,  trained to perform specific tasks, as opposed to the {\em broad}, versatile intelligence allowing humans to adapt to a wide range of environments and  develop a rich set of skills. The human  ability  to  discover robust, invariant high-level concepts and abstractions, and to identify causal relationships from observations appears to be one of the key factors allowing for a  successful generalization from prior experiences to new, often quite different, ``out-of-distribution'' settings.

\looseness=-1Multi-task learning refers to building a system that can solve multiple tasks across different environments \citep{caruana1997multitask,ruder2017overview}. These tasks usually share some common traits. By learning similarities across tasks, a system could utilize knowledge acquired from previous tasks more efficiently when encountering a new task. One possibility of learning such similarities across tasks is to learn a shared underlying data-generating process as a causal generative model whose components satisfy the SMS hypothesis \citep{SchJanLop16}.
In certain cases, causal models adapt faster to sparse interventions in distribution  \citep{ke2019learning, priol2020analysis}.

At the same time, we have clearly come a long way already without explicitly treating the multi-task problem as a causal one.
Fuelled by abundant data and compute, AI has made remarkable advances in a wide range of applications, from image processing and natural language processing \cite{brown2020language} to beating human world champions in games such as chess, poker and Go \citep{schrittwieser2019mastering}, improving medical diagnoses \citep{lundervold2019overview}, and generating 
music \citep{dhariwal2020jukebox}. 
A critical question thus arises: \textit{``Why can’t we just train a huge model that learns environments’ dynamics (e.g. in a RL setting) including all possible interventions? After all, distributed representations can generalize to unseen examples and if we train over a large number of interventions we may expect that a big neural network will generalize across them''}. To address this, we make several points. To begin with, if data was not sufficiently diverse (which is an untestable assumption a priori), the worst-case error to unseen shifts may still be arbitrarily high (see Section \ref{sec:robustness}). While in the short term, we can often beat “out-of-distribution” benchmarks by training bigger models on bigger datasets, causality offers an important complement. The generalization capabilities of a model are tied to its assumptions (e.g., how the model is structured and how it was trained). The causal approach makes these assumptions more explicit and aligned with our understanding of physics and human cognition, for instance by relying on the \hyperlink{pri:im}{Independent Causal Mechanisms principle}. When these assumptions are valid, a learner that does not use them should fare worse than one that does.
Further, if we had a model that was successful in all interventions over a certain environment, we may want to use it in different environments that share similar albeit not necessarily identical dynamics.  The causal approach, and in particular the ICM principle, point to the need to decompose knowledge about the world into independent and recomposable pieces (recomposable depending on the interventions or changes in environment), which suggests more work on modular ML architectures and other ways to enforce the ICM principle in future ML approaches. 

At its core, i.i.d.\ pattern recognition is but a mathematical abstraction, and causality may be essential to most forms of animate learning. Until now, machine learning has neglected a full integration of causality, and this paper argues that it would indeed benefit from integrating causal concepts.
We argue that combining the strengths of both fields, i.e., current deep learning methods as well as tools and ideas from causality, may be a necessary step on the path towards versatile AI systems.

\section{Conclusion}\label{sec:conclusions}
In this work, we discussed different levels of models, including causal and statistical ones. We argued that this spectrum builds upon a range of assumptions both in terms of modeling and data collection. In an effort to bring together causality and machine learning research programs, we first presented a discussion on the fundamentals of causal inference. Second, we discussed how the independent mechanism assumptions and related notions such as invariance offer a powerful bias for causal learning. Third, we discussed how causal relations might be learned from observational and interventional data when causal variables are observed. Fourth, we discussed the open problem of causal representation learning, including its relation to recent interest in the concept of disentangled representations in deep learning. Finally, we discussed how some open research questions in the machine learning community may be better understood and tackled within the causal framework, including semi-supervised learning, domain generalization, and adversarial robustness. 

Based on this discussion, we list some critical areas for future research:
\paragraph{Learning Non-Linear Causal Relations at Scale} \looseness=-1Not all real-world data is unstructured and the effect of interventions can often be observed, for example, by stratifying the data collection across multiple environments. The approximation abilities of modern machine learning methods may prove useful to model non-linear causal relations among large numbers of variables. For practical applications, classical tools are not only limited in the linearity assumptions often made but also in their scalability. The paradigms of meta- and multi-task learning are close to the assumptions and desiderata of causal modeling, and future work should consider (1) understanding under which conditions non-linear causal relations can be learned, (2) which training frameworks allow to best exploit the scalability of machine learning approaches, and (3) providing compelling evidence on the advantages over (non-causal) statistical representations in terms of generalization, re-purposing, and transfer of causal modules on real-world tasks.

\paragraph{Learning Causal Variables} \looseness=-1“Disentangled” representations learned by state-of-the-art neural network methods are still distributed in the sense that they are represented in a vector format with an arbitrary ordering in the dimensions. This fixed-format implies that the representation size cannot be dynamically changed; for example, we cannot change the number of objects in a scene. Further, structured and modular representation should also arise when a network is trained for (sets of) specific tasks, not only auteoncoding. Different high-level variables may be extracted depending on the task and affordances at hand. Understanding under which conditions causal variables can be recovered could provide insights into which interventions we are robust to in predictive tasks. 

\paragraph{Understanding the Biases of Existing Deep Learning Approaches}
Scaling to massive data sets, relying on data augmentation and self-supervision have all been successfully explored to improve the robustness of the predictions of deep learning models. It is nontrivial to disentangle the benefits of the individual components and it is often unclear which ``trick'' should be used when dealing with a new task, even if we have an intuition about useful invariances. The notion of strong generalization over a specific set of interventions may be used to probe existing methods, training schemes, and datasets in order to build a taxonomy of inductive biases. In particular, it is desirable to understand how design choices in pre-training (e.g., which datasets/tasks) positively impact both transfer and robustness downstream in a causal sense.

\paragraph{Learning Causally Correct Models of the World and the Agent} 
In many real-world reinforcement learning (RL) settings, abstract state representations are not available. Hence, the ability to derive abstract causal variables from  high-dimensional, low-level pixel representations and then recover causal graphs is important for causal induction in real-world reinforcement learning settings. 
Moreover, building a causal description for both a model of the agent and the environment (world models) should be essential for robust and versatile model-based reinforcement learning.

\section{Acknowledgments}
Many thanks to the past and present members of the T\"ubingen causality team, without whose work and insights this article would not exist, in particular to Dominik Janzing, Chaochao Lu and Julius von K\"ugelgen who gave helpful comments on \cite{1911.10500}. The text has also benefitted from discussions with Elias Bareinboim, Christoph Bohle, Leon Bottou, Isabelle Guyon, Judea Pearl, and Vladimir Vapnik. Thanks to Wouter van Amsterdam for pointing out typos in the first version. We also thank Thomas Kipf, Klaus Greff, and Alexander d’Amour for the useful discussions.
Finally, we thank the thorough anonymous reviewers for highly valuable feedback and suggestions.

\bibliographystyle{plainnat}  

{\small
\bibliography{references}

\begin{thebibliography}{282}
\providecommand{\natexlab}[1]{#1}
\providecommand{\url}[1]{\texttt{#1}}
\expandafter\ifx\csname urlstyle\endcsname\relax
  \providecommand{\doi}[1]{doi: #1}\else
  \providecommand{\doi}{doi: \begingroup \urlstyle{rm}\Url}\fi

\bibitem[Ahmed et~al.(2021)Ahmed, Tr{\"a}uble, Goyal, Neitz, Wuthrich, Bengio,
  Sch{\"o}lkopf, and Bauer]{ahmed2021causalworld}
Ossama Ahmed, Frederik Tr{\"a}uble, Anirudh Goyal, Alexander Neitz, Manuel
  Wuthrich, Yoshua Bengio, Bernhard Sch{\"o}lkopf, and Stefan Bauer.
\newblock Causalworld: A robotic manipulation benchmark for causal structure
  and transfer learning.
\newblock In \emph{International Conference on Learning Representations}, 2021.

\bibitem[Akkaya et~al.(2019)Akkaya, Andrychowicz, Chociej, Litwin, McGrew,
  Petron, Paino, Plappert, Powell, Ribas, et~al.]{akkaya2019solving}
Ilge Akkaya, Marcin Andrychowicz, Maciek Chociej, Mateusz Litwin, Bob McGrew,
  Arthur Petron, Alex Paino, Matthias Plappert, Glenn Powell, Raphael Ribas,
  et~al.
\newblock Solving rubik's cube with a robot hand.
\newblock \emph{arXiv preprint 1910.07113}, 2019.

\bibitem[Alaa and Schaar(2018)]{alaa2018limits}
Ahmed Alaa and Mihaela Schaar.
\newblock Limits of estimating heterogeneous treatment effects: Guidelines for
  practical algorithm design.
\newblock In \emph{International Conference on Machine Learning}, pages
  129--138, 2018.

\bibitem[Aldrich(1989)]{Aldrich89}
J.~Aldrich.
\newblock Autonomy.
\newblock \emph{Oxford Economic Papers}, 41:\penalty0 15--34, 1989.

\bibitem[Arjovsky et~al.(2019)Arjovsky, Bottou, Gulrajani, and
  Lopez-Paz]{arjovsky2019invariant}
Martin Arjovsky, L{\'e}on Bottou, Ishaan Gulrajani, and David Lopez-Paz.
\newblock Invariant risk minimization.
\newblock \emph{arXiv preprint 1907.02893}, 2019.

\bibitem[Atan et~al.(2018)Atan, Jordon, and van~der Schaar]{atan2018deep}
Onur Atan, James Jordon, and Mihaela van~der Schaar.
\newblock Deep-treat: Learning optimal personalized treatments from
  observational data using neural networks.
\newblock In \emph{Thirty-Second AAAI Conference on Artificial Intelligence},
  2018.

\bibitem[Azulay and Weiss(2019)]{azulay2019deep}
Aharon Azulay and Yair Weiss.
\newblock Why do deep convolutional networks generalize so poorly to small
  image transformations?
\newblock \emph{Journal of Machine Learning Research}, 20\penalty0
  (184):\penalty0 1--25, 2019.

\bibitem[Bahdanau et~al.(2018)Bahdanau, Murty, Noukhovitch, Nguyen, de~Vries,
  and Courville]{bahdanau2018systematic}
Dzmitry Bahdanau, Shikhar Murty, Michael Noukhovitch, Thien~Huu Nguyen, Harm
  de~Vries, and Aaron Courville.
\newblock Systematic generalization: what is required and can it be learned?
\newblock \emph{arXiv preprint 1811.12889}, 2018.

\bibitem[Baird(1990)]{Baird90}
H.~Baird.
\newblock Document image defect models.
\newblock In \emph{Proc., IAPR Workshop on Syntactic and Structural Pattern
  Recognition}, pages 38--46, Murray Hill, NJ, 1990.

\bibitem[Bapst et~al.(2019)Bapst, Sanchez-Gonzalez, Doersch, Stachenfeld,
  Kohli, Battaglia, and Hamrick]{bapst2019structured}
Victor Bapst, Alvaro Sanchez-Gonzalez, Carl Doersch, Kimberly Stachenfeld,
  Pushmeet Kohli, Peter Battaglia, and Jessica Hamrick.
\newblock Structured agents for physical construction.
\newblock In \emph{International Conference on Machine Learning}, pages
  464--474, 2019.

\bibitem[Barbu et~al.(2019)Barbu, Mayo, Alverio, Luo, Wang, Gutfreund,
  Tenenbaum, and Katz]{barbu2019objectnet}
Andrei Barbu, David Mayo, Julian Alverio, William Luo, Christopher Wang, Dan
  Gutfreund, Josh Tenenbaum, and Boris Katz.
\newblock Objectnet: A large-scale bias-controlled dataset for pushing the
  limits of object recognition models.
\newblock In \emph{Advances in Neural Information Processing Systems}, pages
  9448--9458, 2019.

\bibitem[Bareinboim and Pearl(2014)]{Bareinboim2014}
E.~Bareinboim and J.~Pearl.
\newblock Transportability from multiple environments with limited experiments:
  Completeness results.
\newblock In \emph{{A}dvances in {N}eural {I}nformation {P}rocessing {S}ystems
  27}, pages 280--288, 2014.

\bibitem[Bareinboim et~al.(2015)Bareinboim, Forney, and Pearl]{Bareinboim2015}
E.~Bareinboim, A.~Forney, and J.~Pearl.
\newblock Bandits with unobserved confounders: A causal approach.
\newblock In \emph{{A}dvances in {N}eural {I}nformation {P}rocessing {S}ystems
  28}, pages 1342--1350, 2015.

\bibitem[Battaglia et~al.(2016)Battaglia, Pascanu, Lai, Rezende,
  et~al.]{battaglia2016interaction}
Peter Battaglia, Razvan Pascanu, Matthew Lai, Danilo~Jimenez Rezende, et~al.
\newblock Interaction networks for learning about objects, relations and
  physics.
\newblock In \emph{Advances in neural information processing systems}, pages
  4502--4510, 2016.

\bibitem[Battaglia et~al.(2013)Battaglia, Hamrick, and
  Tenenbaum]{battaglia2013simulation}
Peter~W Battaglia, Jessica~B Hamrick, and Joshua~B Tenenbaum.
\newblock Simulation as an engine of physical scene understanding.
\newblock \emph{Proceedings of the National Academy of Sciences}, 110\penalty0
  (45):\penalty0 18327--18332, 2013.

\bibitem[Battaglia et~al.(2018)Battaglia, Hamrick, Bapst, Sanchez-Gonzalez,
  Zambaldi, Malinowski, Tacchetti, Raposo, Santoro, Faulkner,
  et~al.]{battaglia2018relational}
Peter~W Battaglia, Jessica~B Hamrick, Victor Bapst, Alvaro Sanchez-Gonzalez,
  Vinicius Zambaldi, Mateusz Malinowski, Andrea Tacchetti, David Raposo, Adam
  Santoro, Ryan Faulkner, et~al.
\newblock Relational inductive biases, deep learning, and graph networks.
\newblock \emph{arXiv preprint 1806.01261}, 2018.

\bibitem[Bauer et~al.(2016)Bauer, Sch{\"o}lkopf, and Peters]{BauSchPet16}
S.~Bauer, B.~Sch{\"o}lkopf, and J.~Peters.
\newblock The arrow of time in multivariate time series.
\newblock In \emph{Proceedings of the 33nd International Conference on Machine
  Learning}, volume~48 of \emph{{JMLR} Workshop and Conference Proceedings},
  pages 2043--2051, 2016.

\bibitem[Beede et~al.(2020)Beede, Baylor, Hersch, Iurchenko, Wilcox,
  Ruamviboonsuk, and Vardoulakis]{beede2020human}
Emma Beede, Elizabeth Baylor, Fred Hersch, Anna Iurchenko, Lauren Wilcox,
  Paisan Ruamviboonsuk, and Laura~M Vardoulakis.
\newblock A human-centered evaluation of a deep learning system deployed in
  clinics for the detection of diabetic retinopathy.
\newblock In \emph{Proceedings of the 2020 CHI Conference on Human Factors in
  Computing Systems}, pages 1--12, 2020.

\bibitem[Beery et~al.(2018)Beery, Van~Horn, and Perona]{beery2018recognition}
Sara Beery, Grant Van~Horn, and Pietro Perona.
\newblock Recognition in terra incognita.
\newblock In \emph{Proceedings of the European Conference on Computer Vision
  (ECCV)}, pages 456--473, 2018.

\bibitem[Ben-David et~al.(2010)Ben-David, Lu, Luu, and
  P{\'a}l]{ben2010impossibility}
S.~Ben-David, T.~Lu, T.~Luu, and D.~P{\'a}l.
\newblock Impossibility theorems for domain adaptation.
\newblock In \emph{Proceedings of the International Conference on Artificial
  Intelligence and Statistics 13 ({AISTATS})}, pages 129--136, 2010.

\bibitem[Bengio et~al.(2017)Bengio, Thomas, Pineau, Precup, and
  Bengio]{1703.07718}
Emmanuel Bengio, Valentin Thomas, Joelle Pineau, Doina Precup, and Yoshua
  Bengio.
\newblock Independently controllable features.
\newblock \emph{arXiv preprint 1703.07718}, 2017.

\bibitem[Bengio et~al.(1990)Bengio, Bengio, and Cloutier]{bengio1990learning}
Yoshua Bengio, Samy Bengio, and Jocelyn Cloutier.
\newblock \emph{Learning a synaptic learning rule}.
\newblock IJCNN-91-Seattle International Joint Conference on Neural Networks
  (Vol. 2, pp. 969-vol). IEEE., 1990.

\bibitem[Bengio et~al.(2012)Bengio, Courville, and Vincent]{1206.5538}
Yoshua Bengio, Aaron Courville, and Pascal Vincent.
\newblock Representation learning: A review and new perspectives.
\newblock \emph{arXiv preprint 1206.5538}, 2012.

\bibitem[Bengio et~al.(2019)Bengio, Deleu, Rahaman, Ke, Lachapelle, Bilaniuk,
  Goyal, and Pal]{bengio2019meta}
Yoshua Bengio, Tristan Deleu, Nasim Rahaman, Rosemary Ke, S{\'e}bastien
  Lachapelle, Olexa Bilaniuk, Anirudh Goyal, and Christopher Pal.
\newblock A meta-transfer objective for learning to disentangle causal
  mechanisms.
\newblock \emph{arXiv preprint 1901.10912}, 2019.

\bibitem[Benneke et~al.(2019)Benneke, Wong, Piaulet, Knutson, Crossfield,
  Lothringer, Morley, Gao, Greene, Dressing, Dragomir, Howard, McCullough,
  Fortney, and Fraine]{1909.04642}
Björn Benneke, Ian Wong, Caroline Piaulet, Heather~A. Knutson, Ian J.~M.
  Crossfield, Joshua Lothringer, Caroline~V. Morley, Peter Gao, Thomas~P.
  Greene, Courtney Dressing, Diana Dragomir, Andrew~W. Howard, Peter~R.
  McCullough, Eliza M. R. Kempton Jonathan~J. Fortney, and Jonathan Fraine.
\newblock Water vapor on the habitable-zone exoplanet {K2}-18b.
\newblock \emph{arXiv preprint 1909.04642}, 2019.

\bibitem[Berner et~al.(2019)Berner, Brockman, Chan, Cheung, D{\k{e}}biak,
  Dennison, Farhi, Fischer, Hashme, Hesse, et~al.]{berner2019dota}
Christopher Berner, Greg Brockman, Brooke Chan, Vicki Cheung, Przemys{\l}aw
  D{\k{e}}biak, Christy Dennison, David Farhi, Quirin Fischer, Shariq Hashme,
  Chris Hesse, et~al.
\newblock Dota 2 with large scale deep reinforcement learning.
\newblock \emph{arXiv preprint 1912.06680}, 2019.

\bibitem[Besserve et~al.(2018{\natexlab{a}})Besserve, Shajarisales,
  Sch{\"o}lkopf, and Janzing]{BesShaSchJan18}
M.~Besserve, N.~Shajarisales, B.~Sch{\"o}lkopf, and D.~Janzing.
\newblock Group invariance principles for causal generative models.
\newblock In \emph{Proceedings of the 21st International Conference on
  Artificial Intelligence and Statistics (AISTATS)}, pages 557--565,
  2018{\natexlab{a}}.

\bibitem[Besserve et~al.(2021)Besserve, Sun, Janzing, and
  Sch{\"o}lkopf]{BesSunJanSch21}
M.~Besserve, R.~Sun, D.~Janzing, and B.~Sch{\"o}lkopf.
\newblock A theory of independent mechanisms for extrapolation in generative
  models.
\newblock In \emph{35th AAAI Conference on Artificial Intelligence: A Virtual
  Conference}, February 2021.

\bibitem[Besserve et~al.(2018{\natexlab{b}})Besserve, Sun, and
  Schölkopf]{1812.03253}
Michel Besserve, Rémy Sun, and Bernhard Schölkopf.
\newblock Counterfactuals uncover the modular structure of deep generative
  models.
\newblock \emph{arXiv preprint 1812.03253, published at ICLR 2020},
  2018{\natexlab{b}}.

\bibitem[Bica et~al.(2019)Bica, Alaa, and van~der Schaar]{bica2019time}
Ioana Bica, Ahmed~M Alaa, and Mihaela van~der Schaar.
\newblock Time series deconfounder: Estimating treatment effects over time in
  the presence of hidden confounders.
\newblock \emph{arXiv preprint 1902.00450}, 2019.

\bibitem[Blum and Mitchell(1998)]{BluMit98}
Avrim Blum and Tom Mitchell.
\newblock Combining labeled and unlabeled data with co-training.
\newblock In \emph{Proceedings of the Eleventh Annual Conference on
  Computational Learning Theory}, pages 92--100, New York, NY, USA, 1998. ACM.

\bibitem[Blöbaum et~al.(2016)Blöbaum, Washio, and Shimizu]{1610.03263}
Patrick Blöbaum, Takashi Washio, and Shohei Shimizu.
\newblock Error asymmetry in causal and anticausal regression.
\newblock \emph{arXiv preprint 1610.03263}, 2016.

\bibitem[Bonet and Geffner(2019)]{geffner}
Blai Bonet and Hector Geffner.
\newblock Learning first-order symbolic representations for planning from the
  structure of the state space.
\newblock \emph{arXiv preprint 1909.05546}, 2019.

\bibitem[Bottou et~al.(2013)Bottou, Peters, Qui{\~n}onero-Candela, Charles,
  Chickering, Portugualy, Ray, Simard, and Snelson]{Bottou2013}
L.~Bottou, J.~Peters, J.~Qui{\~n}onero-Candela, D.~X. Charles, D.~M.
  Chickering, E.~Portugualy, D.~Ray, P.~Simard, and E.~Snelson.
\newblock Counterfactual reasoning and learning systems: The example of
  computational advertising.
\newblock \emph{Journal of Machine Learning Research}, 14:\penalty0 3207--3260,
  2013.

\bibitem[Brown et~al.(2020)Brown, Mann, Ryder, Subbiah, Kaplan, Dhariwal,
  Neelakantan, Shyam, Sastry, Askell, et~al.]{brown2020language}
Tom~B Brown, Benjamin Mann, Nick Ryder, Melanie Subbiah, Jared Kaplan, Prafulla
  Dhariwal, Arvind Neelakantan, Pranav Shyam, Girish Sastry, Amanda Askell,
  et~al.
\newblock Language models are few-shot learners.
\newblock \emph{arXiv preprint 2005.14165}, 2020.

\bibitem[Budhathoki and Vreeken(2016)]{Vreeken}
Kailash Budhathoki and Jilles Vreeken.
\newblock Causal inference by compression.
\newblock In \emph{IEEE 16th International Conference on Data Mining}, 2016.

\bibitem[Buesing et~al.(2018)Buesing, Weber, Zwols, Racaniere, Guez, Lespiau,
  and Heess]{1811.06272}
Lars Buesing, Theophane Weber, Yori Zwols, Sebastien Racaniere, Arthur Guez,
  Jean-Baptiste Lespiau, and Nicolas Heess.
\newblock Woulda, coulda, shoulda: Counterfactually-guided policy search.
\newblock \emph{arXiv preprint 1811.06272}, 2018.

\bibitem[Burges and Sch{\"o}lkopf(1997)]{BurSch97}
C.~J.~C. Burges and B.~Sch{\"o}lkopf.
\newblock Improving the accuracy and speed of support vector learning machines.
\newblock In M.~Mozer, M.~Jordan, and T.~Petsche, editors, \emph{Advances in
  Neural Information Processing Systems}, volume~9, pages 375--381, Cambridge,
  MA, USA, 1997. MIT Press.

\bibitem[Burgess et~al.(2019)Burgess, Matthey, Watters, Kabra, Higgins,
  Botvinick, and Lerchner]{burgess2019monet}
Christopher~P Burgess, Loic Matthey, Nicholas Watters, Rishabh Kabra, Irina
  Higgins, Matt Botvinick, and Alexander Lerchner.
\newblock Monet: Unsupervised scene decomposition and representation.
\newblock \emph{arXiv preprint 1901.11390}, 2019.

\bibitem[Caruana(1997)]{caruana1997multitask}
Rich Caruana.
\newblock Multitask learning.
\newblock \emph{Machine learning}, 28\penalty0 (1):\penalty0 41--75, 1997.

\bibitem[Chalupka et~al.(2015)Chalupka, Perona, and Eberhardt]{1512.07942}
Krzysztof Chalupka, Pietro Perona, and Frederick Eberhardt.
\newblock Multi-level cause-effect systems.
\newblock \emph{arXiv preprint 1512.07942}, 2015.

\bibitem[Chalupka et~al.(2018)Chalupka, Perona, and Eberhardt]{1804.02747}
Krzysztof Chalupka, Pietro Perona, and Frederick Eberhardt.
\newblock Fast conditional independence test for vector variables with large
  sample sizes.
\newblock \emph{arXiv preprint 1804.02747}, 2018.

\bibitem[Chang et~al.(2017)Chang, Ullman, Torralba, and
  Tenenbaum]{chang2016compositional}
Michael~B Chang, Tomer Ullman, Antonio Torralba, and Joshua~B Tenenbaum.
\newblock A compositional object-based approach to learning physical dynamics.
\newblock In \emph{5th International Conference on Learning Representations
  (ICLR)}, 2017.

\bibitem[Chapelle et~al.(2006)Chapelle, Sch{\"o}lkopf, and Zien]{ChaSchZie06}
O.~Chapelle, B.~Sch{\"o}lkopf, and A.~Zien, editors.
\newblock \emph{Semi-Supervised Learning}.
\newblock MIT Press, Cambridge, MA, USA, 2006.
\newblock URL \url{http://www.kyb.tuebingen.mpg.de/ssl-book/}.

\bibitem[Chen et~al.(2020{\natexlab{a}})Chen, Radford, Child, Wu, Jun,
  Dhariwal, Luan, and Sutskever]{chen2020generative}
Mark Chen, Alec Radford, Rewon Child, Jeff Wu, Heewoo Jun, Prafulla Dhariwal,
  David Luan, and Ilya Sutskever.
\newblock Generative pretraining from pixels.
\newblock In \emph{Proceedings of the 37th International Conference on Machine
  Learning}, 2020{\natexlab{a}}.

\bibitem[Chen et~al.(2020{\natexlab{b}})Chen, Kornblith, Norouzi, and
  Hinton]{chen2020simple}
Ting Chen, Simon Kornblith, Mohammad Norouzi, and Geoffrey Hinton.
\newblock A simple framework for contrastive learning of visual
  representations.
\newblock \emph{arXiv preprint 2002.05709}, 2020{\natexlab{b}}.

\bibitem[Chiappa et~al.(2017)Chiappa, Racaniere, Wierstra, and
  Mohamed]{chiappa2017recurrent}
Silvia Chiappa, S{\'e}bastien Racaniere, Daan Wierstra, and Shakir Mohamed.
\newblock Recurrent environment simulators.
\newblock In \emph{5th International Conference on Learning Representations
  (ICLR)}, 2017.

\bibitem[Cubuk et~al.(2019)Cubuk, Zoph, Mane, Vasudevan, and
  Le]{cubuk2019autoaugment}
Ekin~D Cubuk, Barret Zoph, Dandelion Mane, Vijay Vasudevan, and Quoc~V Le.
\newblock Autoaugment: Learning augmentation strategies from data.
\newblock In \emph{Proceedings of the IEEE conference on computer vision and
  pattern recognition}, pages 113--123, 2019.

\bibitem[Daniu\v{s}is et~al.(2010)Daniu\v{s}is, Janzing, Mooij, Zscheischler,
  Steudel, Zhang, and Sch{\"o}lkopf]{Daniusisetal10}
P.~Daniu\v{s}is, D.~Janzing, J.~M. Mooij, J.~Zscheischler, B.~Steudel,
  K.~Zhang, and B.~Sch{\"o}lkopf.
\newblock Inferring deterministic causal relations.
\newblock In \emph{Proceedings of the 26th Annual Conference on {U}ncertainty
  in {A}rtificial {I}ntelligence ({UAI})}, pages 143--150, 2010.

\bibitem[Dasgupta et~al.(2019)Dasgupta, Wang, Chiappa, Mitrovic, Ortega,
  Raposo, Hughes, Battaglia, Botvinick, and Kurth-Nelson]{dasgupta2019causal}
Ishita Dasgupta, Jane Wang, Silvia Chiappa, Jovana Mitrovic, Pedro Ortega,
  David Raposo, Edward Hughes, Peter Battaglia, Matthew Botvinick, and Zeb
  Kurth-Nelson.
\newblock Causal reasoning from meta-reinforcement learning.
\newblock \emph{arXiv preprint 1901.08162}, 2019.

\bibitem[Dawid(1979)]{Dawid79}
A.~P. Dawid.
\newblock Conditional independence in statistical theory.
\newblock \emph{Journal of the Royal Statistical Society B}, 41\penalty0
  (1):\penalty0 1--31, 1979.

\bibitem[Dehaene(2020)]{dehaene2020we}
Stanislas Dehaene.
\newblock \emph{How We Learn: Why Brains Learn Better Than Any Machine... for
  Now}.
\newblock Penguin, 2020.

\bibitem[Deng et~al.(2009)Deng, Dong, Socher, Li, Li, and
  Fei-Fei]{deng2009imagenet}
Jia Deng, Wei Dong, Richard Socher, Li-Jia Li, Kai Li, and Li~Fei-Fei.
\newblock Imagenet: A large-scale hierarchical image database.
\newblock In \emph{2009 IEEE conference on computer vision and pattern
  recognition}, pages 248--255. Ieee, 2009.

\bibitem[Devlin et~al.(2018)Devlin, Chang, Lee, and Toutanova]{devlin2018bert}
Jacob Devlin, Ming-Wei Chang, Kenton Lee, and Kristina Toutanova.
\newblock Bert: Pre-training of deep bidirectional transformers for language
  understanding.
\newblock \emph{arXiv preprint 1810.04805}, 2018.

\bibitem[Devroye et~al.(1996)Devroye, Gy{\"o}rfi, and Lugosi]{DevGyoLug96}
L.~Devroye, L.~Gy{\"o}rfi, and G.~Lugosi.
\newblock \emph{A Probabilistic Theory of Pattern Recognition}, volume~31 of
  \emph{Applications of Mathematics}.
\newblock Springer, New York, NY, 1996.

\bibitem[Dhariwal et~al.(2020)Dhariwal, Jun, Payne, Kim, Radford, and
  Sutskever]{dhariwal2020jukebox}
Prafulla Dhariwal, Heewoo Jun, Christine Payne, Jong~Wook Kim, Alec Radford,
  and Ilya Sutskever.
\newblock Jukebox: A generative model for music.
\newblock \emph{arXiv preprint 2005.00341}, 2020.

\bibitem[Dittadi et~al.(2021)Dittadi, Tr{\"a}uble, Locatello, W{\"u}thrich,
  Agrawal, Winther, Bauer, and Sch{\"o}lkopf]{dittadi2021on}
Andrea Dittadi, Frederik Tr{\"a}uble, Francesco Locatello, Manuel W{\"u}thrich,
  Vaibhav Agrawal, Ole Winther, Stefan Bauer, and Bernhard Sch{\"o}lkopf.
\newblock On the transfer of disentangled representations in realistic
  settings.
\newblock In \emph{International Conference on Learning Representations}, 2021.

\bibitem[Diuk et~al.(2008)Diuk, Cohen, and Littman]{diuk2008object}
Carlos Diuk, Andre Cohen, and Michael~L Littman.
\newblock An object-oriented representation for efficient reinforcement
  learning.
\newblock In \emph{Proceedings of the 25th international conference on Machine
  learning}, pages 240--247, 2008.

\bibitem[Djolonga et~al.(2020)Djolonga, Yung, Tschannen, Romijnders, Beyer,
  Kolesnikov, Puigcerver, Minderer, D'Amour, Moldovan,
  et~al.]{djolonga2020robustness}
Josip Djolonga, Jessica Yung, Michael Tschannen, Rob Romijnders, Lucas Beyer,
  Alexander Kolesnikov, Joan Puigcerver, Matthias Minderer, Alexander D'Amour,
  Dan Moldovan, et~al.
\newblock On robustness and transferability of convolutional neural networks.
\newblock \emph{arXiv preprint 2007.08558}, 2020.

\bibitem[Doran et~al.(2014)Doran, Muandet, Zhang, and
  Sch{\"o}lkopf]{DorMuaZhaSch14}
G.~Doran, K.~Muandet, K.~Zhang, and B.~Sch{\"o}lkopf.
\newblock A permutation-based kernel conditional independence test.
\newblock In N.~L. Zhang and J.~Tian, editors, \emph{Proceedings of the 30th
  Conference on Uncertainty in Artificial Intelligence}, pages 132--141,
  Corvallis, OR, 2014. AUAI Press.
\newblock URL \url{http://auai.org/uai2014/proceedings/individuals/194.pdf}.

\bibitem[Eastwood and Williams(2018)]{eastwood2018framework}
Cian Eastwood and Christopher~KI Williams.
\newblock A framework for the quantitative evaluation of disentangled
  representations.
\newblock In \emph{International Conference on Learning Representations}, 2018.

\bibitem[Eaton and Murphy(2007)]{eaton2007exact}
Daniel Eaton and Kevin Murphy.
\newblock Exact {Bayesian} structure learning from uncertain interventions.
\newblock In \emph{Artificial Intelligence and Statistics}, pages 107--114,
  2007.

\bibitem[Engstrom et~al.(2017)Engstrom, Tran, Tsipras, Schmidt, and
  Madry]{engstrom2017exploring}
Logan Engstrom, Brandon Tran, Dimitris Tsipras, Ludwig Schmidt, and Aleksander
  Madry.
\newblock Exploring the landscape of spatial robustness.
\newblock \emph{arXiv preprint 1712.02779}, 2017.

\bibitem[Epstude and Roese(2008)]{epstude2008functional}
Kai Epstude and Neal~J Roese.
\newblock The functional theory of counterfactual thinking.
\newblock \emph{Personality and social psychology review}, 12\penalty0
  (2):\penalty0 168--192, 2008.

\bibitem[Esteva et~al.(2019)Esteva, Robicquet, Ramsundar, Kuleshov, DePristo,
  Chou, Cui, Corrado, Thrun, and Dean]{esteva2019guide}
Andre Esteva, Alexandre Robicquet, Bharath Ramsundar, Volodymyr Kuleshov, Mark
  DePristo, Katherine Chou, Claire Cui, Greg Corrado, Sebastian Thrun, and Jeff
  Dean.
\newblock A guide to deep learning in healthcare.
\newblock \emph{{N}ature {M}edicine}, 25\penalty0 (1):\penalty0 24--29, 2019.

\bibitem[Farag{\'o} and Lugosi(2006)]{farago2006strong}
Andr{\'a}s Farag{\'o} and G{\'a}bor Lugosi.
\newblock Strong universal consistency of neural network classifiers.
\newblock \emph{IEEE Transactions on Information Theory}, 39\penalty0
  (4):\penalty0 1146--1151, 2006.

\bibitem[Finn et~al.(2017)Finn, Abbeel, and Levine]{finn2017model}
Chelsea Finn, Pieter Abbeel, and Sergey Levine.
\newblock Model-agnostic meta-learning for fast adaptation of deep networks.
\newblock \emph{arXiv preprint 1703.03400}, 2017.

\bibitem[Foerster et~al.(2018)Foerster, Farquhar, Afouras, Nardelli, and
  Whiteson]{foerster2018counterfactual}
Jakob~N Foerster, Gregory Farquhar, Triantafyllos Afouras, Nantas Nardelli, and
  Shimon Whiteson.
\newblock Counterfactual multi-agent policy gradients.
\newblock In \emph{Thirty-second AAAI conference on artificial intelligence},
  2018.

\bibitem[F{\"o}ldi{\'a}k(1991)]{foldiak1991learning}
Peter F{\"o}ldi{\'a}k.
\newblock Learning invariance from transformation sequences.
\newblock \emph{Neural Computation}, 3\penalty0 (2):\penalty0 194--200, 1991.

\bibitem[Foreman-Mackey et~al.(2015)Foreman-Mackey, Montet, Hogg, Morton, Wang,
  and Sch{\"o}lkopf]{Foreman-Mackeyetal15}
D.~Foreman-Mackey, B.~T. Montet, D.~W. Hogg, T.~D. Morton, D.~Wang, and
  B.~Sch{\"o}lkopf.
\newblock A systematic search for transiting planets in the {K2} data.
\newblock \emph{The Astrophysical Journal}, 806\penalty0 (2), 2015.
\newblock URL \url{http://stacks.iop.org/0004-637X/806/i=2/a=215}.

\bibitem[Frisch et~al.(1948)Frisch, Haavelmo, Koopmans, and
  Tinbergen]{Frisch1948}
R.~Frisch, T.~Haavelmo, T.C. Koopmans, and J.~Tinbergen.
\newblock \emph{Autonomy of economic relations}.
\newblock Universitets Social{\o}konomiske Institutt, Oslo, Norway, 1948.

\bibitem[Fujimoto et~al.(2019)Fujimoto, Meger, and Precup]{fujimoto2019off}
Scott Fujimoto, David Meger, and Doina Precup.
\newblock Off-policy deep reinforcement learning without exploration.
\newblock In \emph{International Conference on Machine Learning}, pages
  2052--2062, 2019.

\bibitem[Fukumizu et~al.(2008)Fukumizu, Gretton, Sun, and
  Sch\"olkopf]{Fukumizu2008}
K.~Fukumizu, A.~Gretton, X.~Sun, and B.~Sch\"olkopf.
\newblock Kernel measures of conditional dependence.
\newblock In \emph{{A}dvances in {N}eural {I}nformation {P}rocessing {S}ystems
  20}, pages 489--496, 2008.

\bibitem[Geiger and Pearl(1990)]{GeiPea90}
D.~Geiger and J.~Pearl.
\newblock Logical and algorithmic properties of independence and their
  application to {Bayesian} networks.
\newblock \emph{Annals of Mathematics and Artificial Intelligence}, 2:\penalty0
  165–178, 1990.

\bibitem[Geirhos et~al.(2018)Geirhos, Rubisch, Michaelis, Bethge, Wichmann, and
  Brendel]{geirhos2018imagenet}
Robert Geirhos, Patricia Rubisch, Claudio Michaelis, Matthias Bethge, Felix~A
  Wichmann, and Wieland Brendel.
\newblock Imagenet-trained cnns are biased towards texture; increasing shape
  bias improves accuracy and robustness.
\newblock \emph{arXiv preprint 1811.12231}, 2018.

\bibitem[Gondal et~al.(2019)Gondal, W{\"u}thrich, Miladinovi{\'c}, Locatello,
  Breidt, Volchkov, Akpo, Bachem, Sch{\"o}lkopf, and Bauer]{gondal2019transfer}
Muhammad~Waleed Gondal, Manuel W{\"u}thrich, Djordje Miladinovi{\'c}, Francesco
  Locatello, Martin Breidt, Valentin Volchkov, Joel Akpo, Olivier Bachem,
  Bernhard Sch{\"o}lkopf, and Stefan Bauer.
\newblock On the transfer of inductive bias from simulation to the real world:
  a new disentanglement dataset.
\newblock In \emph{Advances in Neural Information Processing Systems}, pages
  15740--15751, 2019.

\bibitem[Gong et~al.(2016)Gong, Zhang, Liu, Tao, Glymour, and
  Sch{\"o}lkopf]{GonZhaLiuTaoSch16}
M.~Gong, K.~Zhang, T.~Liu, D.~Tao, C.~Glymour, and B.~Sch{\"o}lkopf.
\newblock Domain adaptation with conditional transferable components.
\newblock In \emph{Proceedings of the 33nd International Conference on Machine
  Learning}, pages 2839--2848, 2016.

\bibitem[Gong et~al.(2017)Gong, Zhang, Sch{\"o}lkopf, Glymour, and
  Tao]{Gongetal17}
M.~Gong, K.~Zhang, B.~Sch{\"o}lkopf, C.~Glymour, and D.~Tao.
\newblock Causal discovery from temporally aggregated time series.
\newblock In \emph{Proceedings of the Thirty-Third Conference on Uncertainty in
  Artificial Intelligence (UAI)}, page ID 269, 2017.

\bibitem[Goodfellow et~al.(2014)Goodfellow, Shlens, and
  Szegedy]{goodfellow2014explaining}
Ian~J Goodfellow, Jonathon Shlens, and Christian Szegedy.
\newblock Explaining and harnessing adversarial examples.
\newblock \emph{arXiv preprint 1412.6572}, 2014.

\bibitem[Gopnik et~al.(2004)Gopnik, Glymour, Sobel, Schulz, Kushnir, and
  Danks]{gopnik2004theory}
Alison Gopnik, Clark Glymour, David~M Sobel, Laura~E Schulz, Tamar Kushnir, and
  David Danks.
\newblock A theory of causal learning in children: causal maps and {Bayes}
  nets.
\newblock \emph{Psychological review}, 111\penalty0 (1):\penalty0 3, 2004.

\bibitem[Gottesman et~al.(2018)Gottesman, Johansson, Meier, Dent, Lee,
  Srinivasan, Zhang, Ding, Wihl, Peng, Yao, Lage, Mosch, wei H.~Lehman,
  Komorowski, Komorowski, Faisal, Celi, Sontag, and Doshi-Velez]{1805.12298}
Omer Gottesman, Fredrik Johansson, Joshua Meier, Jack Dent, Donghun Lee,
  Srivatsan Srinivasan, Linying Zhang, Yi~Ding, David Wihl, Xuefeng Peng, Jiayu
  Yao, Isaac Lage, Christopher Mosch, Li~wei H.~Lehman, Matthieu Komorowski,
  Matthieu Komorowski, Aldo Faisal, Leo~Anthony Celi, David Sontag, and Finale
  Doshi-Velez.
\newblock Evaluating reinforcement learning algorithms in observational health
  settings.
\newblock \emph{arXiv preprint 1805.12298}, 2018.

\bibitem[Goudet et~al.(2017)Goudet, Kalainathan, Caillou, Guyon, Lopez-Paz, and
  Sebag]{1711.08936}
Olivier Goudet, Diviyan Kalainathan, Philippe Caillou, Isabelle Guyon, David
  Lopez-Paz, and Michèle Sebag.
\newblock Causal generative neural networks.
\newblock \emph{arXiv preprint 1711.08936}, 2017.

\bibitem[Goyal et~al.(2020)Goyal, Lamb, Gampa, Beaudoin, Levine, Blundell,
  Bengio, and Mozer]{goyal2020object}
Anirudh Goyal, Alex Lamb, Phanideep Gampa, Philippe Beaudoin, Sergey Levine,
  Charles Blundell, Yoshua Bengio, and Michael Mozer.
\newblock Object files and schemata: Factorizing declarative and procedural
  knowledge in dynamical systems.
\newblock \emph{arXiv preprint 2006.16225}, 2020.

\bibitem[Goyal et~al.(2021)Goyal, Lamb, Hoffmann, Sodhani, Levine, Bengio, and
  Schölkopf]{RIMs}
Anirudh Goyal, Alex Lamb, Jordan Hoffmann, Shagun Sodhani, Sergey Levine,
  Yoshua Bengio, and Bernhard Schölkopf.
\newblock Recurrent independent mechanisms.
\newblock In \emph{International Conference on Learning Representations}, 2021.

\bibitem[Graves et~al.(2013)Graves, Mohamed, and Hinton]{graves2013speech}
Alex Graves, Abdel-rahman Mohamed, and Geoffrey Hinton.
\newblock Speech recognition with deep recurrent neural networks.
\newblock In \emph{2013 IEEE international conference on acoustics, speech and
  signal processing}, pages 6645--6649. IEEE, 2013.

\bibitem[Greff et~al.(2019)Greff, Kaufman, Kabra, Watters, Burgess, Zoran,
  Matthey, Botvinick, and Lerchner]{greff2019multi}
Klaus Greff, Rapha{\"e}l~Lopez Kaufman, Rishabh Kabra, Nick Watters,
  Christopher Burgess, Daniel Zoran, Loic Matthey, Matthew Botvinick, and
  Alexander Lerchner.
\newblock Multi-object representation learning with iterative variational
  inference.
\newblock In \emph{International Conference on Machine Learning}, pages
  2424--2433, 2019.

\bibitem[Greff et~al.(2020)Greff, van Steenkiste, and
  Schmidhuber]{greff2020binding}
Klaus Greff, Sjoerd van Steenkiste, and J{\"u}rgen Schmidhuber.
\newblock On the binding problem in artificial neural networks.
\newblock \emph{arXiv preprint 2012.05208}, 2020.

\bibitem[Gregor et~al.(2019)Gregor, Rezende, Besse, Wu, Merzic, and van~den
  Oord]{gregor2019shaping}
Karol Gregor, Danilo~Jimenez Rezende, Frederic Besse, Yan Wu, Hamza Merzic, and
  Aaron van~den Oord.
\newblock Shaping belief states with generative environment models for rl.
\newblock In \emph{Advances in Neural Information Processing Systems}, pages
  13475--13487, 2019.

\bibitem[Gresele et~al.(2019)Gresele, Rubenstein, Mehrjou, Locatello, and
  Sch{\"o}lkopf]{gresele2019incomplete}
Luigi Gresele, Paul~K Rubenstein, Arash Mehrjou, Francesco Locatello, and
  Bernhard Sch{\"o}lkopf.
\newblock The incomplete rosetta stone problem: Identifiability results for
  multi-view nonlinear ica.
\newblock \emph{arXiv preprint 1905.06642}, 2019.

\bibitem[Gretton et~al.(2005{\natexlab{a}})Gretton, Bousquet, Smola, and
  Sch\"{o}lkopf]{Gretton2005}
A.~Gretton, O.~Bousquet, A.~Smola, and B.~Sch\"{o}lkopf.
\newblock Measuring statistical dependence with {H}ilbert-{S}chmidt norms.
\newblock In \emph{Algorithmic Learning Theory}, pages 63--78. Springer-Verlag,
  2005{\natexlab{a}}.

\bibitem[Gretton et~al.(2005{\natexlab{b}})Gretton, Herbrich, Smola, Bousquet,
  and Sch\"olkopf]{Gretton2005JMLR}
A.~Gretton, R.~Herbrich, A.~Smola, O.~Bousquet, and B.~Sch\"olkopf.
\newblock Kernel methods for measuring independence.
\newblock \emph{Journal of Machine Learning Research}, 6:\penalty0 2075--2129,
  2005{\natexlab{b}}.

\bibitem[Grill et~al.(2020)Grill, Strub, Altch{\'e}, Tallec, Richemond,
  Buchatskaya, Doersch, Pires, Guo, Azar, et~al.]{grill2020bootstrap}
Jean-Bastien Grill, Florian Strub, Florent Altch{\'e}, Corentin Tallec,
  Pierre~H Richemond, Elena Buchatskaya, Carl Doersch, Bernardo~Avila Pires,
  Zhaohan~Daniel Guo, Mohammad~Gheshlaghi Azar, et~al.
\newblock Bootstrap your own latent: A new approach to self-supervised
  learning.
\newblock \emph{arXiv preprint 2006.07733}, 2020.

\bibitem[Grzeszczuk et~al.(1998)Grzeszczuk, Terzopoulos, and
  Hinton]{grzeszczuk1998neuroanimator}
Radek Grzeszczuk, Demetri Terzopoulos, and Geoffrey Hinton.
\newblock Neuroanimator: Fast neural network emulation and control of
  physics-based models.
\newblock In \emph{Proceedings of the 25th annual conference on Computer
  graphics and interactive techniques}, pages 9--20, 1998.

\bibitem[Gu et~al.(2019)Gu, Yang, Ngiam, Le, and Shlens]{gu2019using}
Keren Gu, Brandon Yang, Jiquan Ngiam, Quoc Le, and Jonathan Shlens.
\newblock Using videos to evaluate image model robustness.
\newblock \emph{arXiv preprint 1904.10076}, 2019.

\bibitem[Gu and Rigazio(2014)]{DBR}
Shixiang Gu and Luca Rigazio.
\newblock Towards deep neural network architectures robust to adversarial
  examples, 2014.
\newblock arXiv:1412.5068.

\bibitem[Guo et~al.(2018)Guo, Cheng, Li, Hahn, and Liu]{1809.09337}
Ruocheng Guo, Lu~Cheng, Jundong Li, P.~Richard Hahn, and Huan Liu.
\newblock A survey of learning causality with data: Problems and methods.
\newblock \emph{arXiv preprint 1809.09337}, 2018.

\bibitem[Guyon et~al.(2010)Guyon, Janzing, and Sch{\"o}lkopf]{GuyonJS2010}
I.~Guyon, D.~Janzing, and B.~Sch{\"o}lkopf.
\newblock Causality: Objectives and assessment.
\newblock In I.~Guyon, D.~Janzing, and B.~Sch{\"o}lkopf, editors, \emph{JMLR
  Workshop and Conference Proceedings: Volume 6}, pages 1--42, Cambridge, MA,
  USA, 2010. MIT Press.

\bibitem[Ha and Schmidhuber(2018)]{ha2018world}
David Ha and J{\"u}rgen Schmidhuber.
\newblock World models.
\newblock \emph{arXiv preprint 1803.10122}, 2018.

\bibitem[Haavelmo(1944)]{Haavelmo1944}
T.~Haavelmo.
\newblock The probability approach in econometrics.
\newblock \emph{Econometrica}, 12:\penalty0 S1--S115 (supplement), 1944.

\bibitem[H{\"a}lv{\"a} and Hyv{\"a}rinen(2020)]{halva2020hidden}
Hermanni H{\"a}lv{\"a} and Aapo Hyv{\"a}rinen.
\newblock Hidden markov nonlinear ica: Unsupervised learning from nonstationary
  time series.
\newblock \emph{arXiv preprint 2006.12107}, 2020.

\bibitem[He et~al.(2016)He, Zhang, Ren, and Sun]{he2016deep}
Kaiming He, Xiangyu Zhang, Shaoqing Ren, and Jian Sun.
\newblock Deep residual learning for image recognition.
\newblock In \emph{Proceedings of the IEEE conference on computer vision and
  pattern recognition}, pages 770--778, 2016.

\bibitem[He et~al.(2020)He, Fan, Wu, Xie, and Girshick]{he2020momentum}
Kaiming He, Haoqi Fan, Yuxin Wu, Saining Xie, and Ross Girshick.
\newblock Momentum contrast for unsupervised visual representation learning.
\newblock In \emph{Proceedings of the IEEE/CVF Conference on Computer Vision
  and Pattern Recognition}, pages 9729--9738, 2020.

\bibitem[He et~al.(2019)He, Li, Feng, Ho, Ravanbakhsh, Chen, and
  P{\'o}czos]{he2019learning}
Siyu He, Yin Li, Yu~Feng, Shirley Ho, Siamak Ravanbakhsh, Wei Chen, and
  Barnab{\'a}s P{\'o}czos.
\newblock Learning to predict the cosmological structure formation.
\newblock \emph{Proceedings of the National Academy of Sciences}, 116\penalty0
  (28):\penalty0 13825--13832, 2019.

\bibitem[Heinze-Deml and Meinshausen(2017)]{1710.11469}
Christina Heinze-Deml and Nicolai Meinshausen.
\newblock Conditional variance penalties and domain shift robustness.
\newblock \emph{arXiv preprint 1710.11469}, 2017.

\bibitem[Heinze-Deml et~al.(2017)Heinze-Deml, Peters, and
  Meinshausen]{1706.08576}
Christina Heinze-Deml, Jonas Peters, and Nicolai Meinshausen.
\newblock Invariant causal prediction for nonlinear models.
\newblock \emph{arXiv preprint 1706.08576}, 2017.

\bibitem[Hendrycks and Dietterich(2019)]{hendrycks2019benchmarking}
Dan Hendrycks and Thomas Dietterich.
\newblock Benchmarking neural network robustness to common corruptions and
  perturbations.
\newblock \emph{arXiv preprint 1903.12261}, 2019.

\bibitem[Henrich(2016)]{Henrich}
Joseph Henrich.
\newblock \emph{The Secret of our Success}.
\newblock Princeton University Press, 2016.

\bibitem[Henry et~al.(2015)Henry, Hager, Pronovost, and
  Saria]{henry2015targeted}
Katharine~E Henry, David~N Hager, Peter~J Pronovost, and Suchi Saria.
\newblock A targeted real-time early warning score (trewscore) for septic
  shock.
\newblock \emph{Science translational medicine}, 7\penalty0 (299):\penalty0
  299ra122--299ra122, 2015.

\bibitem[Higgins et~al.(2016)Higgins, Matthey, Pal, Burgess, Glorot, Botvinick,
  Mohamed, and Lerchner]{higgins2016beta}
Irina Higgins, Loic Matthey, Arka Pal, Christopher Burgess, Xavier Glorot,
  Matthew Botvinick, Shakir Mohamed, and Alexander Lerchner.
\newblock beta-vae: Learning basic visual concepts with a constrained
  variational framework.
\newblock In \emph{International Conference on Learning Representations}, 2016.

\bibitem[Hjelm and Buchwalter(2019)]{bachman2019learning}
R~Devon Hjelm and William Buchwalter.
\newblock Learning representations by maximizing mutual information across
  views.
\newblock In \emph{Advances in Neural Information Processing Systems}, pages
  15535--15545, 2019.

\bibitem[Hoover(2008)]{Hoover06}
K.~D. Hoover.
\newblock Causality in economics and econometrics.
\newblock In S.~N. Durlauf and L.~E. Blume, editors, \emph{The New Palgrave
  Dictionary of Economics}. Palgrave Macmillan, Basingstoke, UK, 2nd edition,
  2008.

\bibitem[Howard and Ruder(2018)]{howard2018universal}
Jeremy Howard and Sebastian Ruder.
\newblock Universal language model fine-tuning for text classification.
\newblock \emph{arXiv preprint 1801.06146}, 2018.

\bibitem[Hoyer et~al.(2009)Hoyer, Janzing, Mooij, Peters, and
  Sch\"olkopf]{Hoyer2008}
P.~O. Hoyer, D.~Janzing, J.~M. Mooij, J.~Peters, and B.~Sch\"olkopf.
\newblock Nonlinear causal discovery with additive noise models.
\newblock In \emph{{A}dvances in {N}eural {I}nformation {P}rocessing {S}ystems
  21 ({NIPS})}, pages 689--696, 2009.

\bibitem[Huang et~al.(2017)Huang, Zhang, Zhang, Sanchez-Romero, Glymour, and
  Sch{\"o}lkopf]{HuaZhaZhaSanGlySch17}
B.~Huang, K.~Zhang, J.~Zhang, R.~Sanchez-Romero, C.~Glymour, and
  B.~Sch{\"o}lkopf.
\newblock Behind distribution shift: Mining driving forces of changes and
  causal arrows.
\newblock In \emph{{IEEE} 17th International Conference on Data Mining (ICDM
  2017)}, pages 913--918, 2017.

\bibitem[Huang et~al.(2020)Huang, Zhang, Zhang, Ramsey, Sanchez-Romero,
  Glymour, and Schölkopf]{JMLR:v21:19-232}
Biwei Huang, Kun Zhang, Jiji Zhang, Joseph Ramsey, Ruben Sanchez-Romero, Clark
  Glymour, and Bernhard Schölkopf.
\newblock Causal discovery from heterogeneous/nonstationary data.
\newblock \emph{Journal of Machine Learning Research}, 21\penalty0
  (89):\penalty0 1--53, 2020.
\newblock URL \url{http://jmlr.org/papers/v21/19-232.html}.

\bibitem[Hyv{\"a}rinen and Pajunen(1999)]{hyvarinen1999nonlinear}
Aapo Hyv{\"a}rinen and Petteri Pajunen.
\newblock Nonlinear independent component analysis: Existence and uniqueness
  results.
\newblock \emph{Neural networks}, 12\penalty0 (3):\penalty0 429--439, 1999.

\bibitem[Hyvarinen and Morioka(2017)]{hyvarinen2017nonlinear}
AJ~Hyvarinen and Hiroshi Morioka.
\newblock Nonlinear ica of temporally dependent stationary sources.
\newblock In \emph{Proceedings of Machine Learning Research}, 2017.

\bibitem[Imbens and Rubin(2015)]{imbens2015causal}
Guido~W Imbens and Donald~B Rubin.
\newblock \emph{Causal inference in statistics, social, and biomedical
  sciences}.
\newblock Cambridge University Press, 2015.

\bibitem[Janzing(2019)]{Janzing_NIPS2019}
D.~Janzing.
\newblock Causal regularization.
\newblock In \emph{Advances in Neural Information Processing Systems 33}, 2019.

\bibitem[Janzing and Sch{\"o}lkopf(2010)]{JanSch10}
D.~Janzing and B.~Sch{\"o}lkopf.
\newblock Causal inference using the algorithmic {M}arkov condition.
\newblock \emph{IEEE Transactions on Information Theory}, 56\penalty0
  (10):\penalty0 5168--5194, 2010.

\bibitem[Janzing and Sch{{\"o}}lkopf(2015)]{JanSch15}
D.~Janzing and B.~Sch{{\"o}}lkopf.
\newblock Semi-supervised interpolation in an anticausal learning scenario.
\newblock \emph{Journal of Machine Learning Research}, 16:\penalty0 1923--1948,
  2015.

\bibitem[Janzing and Sch{\"o}lkopf(2018)]{JanSch18b}
D.~Janzing and B.~Sch{\"o}lkopf.
\newblock Detecting non-causal artifacts in multivariate linear regression
  models.
\newblock In \emph{Proceedings of the 35th International Conference on Machine
  Learning (ICML)}, pages 2250--2258, 2018.

\bibitem[Janzing et~al.(2009)Janzing, Peters, Mooij, and
  Sch\"{o}lkopf]{Janzing2009uai}
D.~Janzing, J.~Peters, J.~M. Mooij, and B.~Sch\"{o}lkopf.
\newblock Identifying confounders using additive noise models.
\newblock In \emph{Proceedings of the 25th Annual Conference on {U}ncertainty
  in {A}rtificial {I}ntelligence ({UAI})}, pages 249--257, 2009.

\bibitem[Janzing et~al.(2010)Janzing, Hoyer, and Sch{\"o}lkopf]{JanHoySch10}
D.~Janzing, P.~Hoyer, and B.~Sch{\"o}lkopf.
\newblock Telling cause from effect based on high-dimensional observations.
\newblock In J.~F{\"u}rnkranz and T.~Joachims, editors, \emph{Proceedings of
  the 27th International Conference on Machine Learning}, pages 479--486, 2010.

\bibitem[Janzing et~al.(2012)Janzing, Mooij, Zhang, Lemeire, Zscheischler,
  Daniu\v{s}is, Steudel, and Sch\"olkopf]{Janzingetal12}
D.~Janzing, J.~M. Mooij, K.~Zhang, J.~Lemeire, J.~Zscheischler,
  P.~Daniu\v{s}is, B.~Steudel, and B.~Sch\"olkopf.
\newblock Information-geometric approach to inferring causal directions.
\newblock \emph{Artificial Intelligence}, 182--183:\penalty0 1--31, 2012.

\bibitem[Janzing et~al.(2016)Janzing, Chaves, and Sch{\"o}lkopf]{JanChaSch16}
D.~Janzing, R.~Chaves, and B.~Sch{\"o}lkopf.
\newblock Algorithmic independence of initial condition and dynamical law in
  thermodynamics and causal inference.
\newblock \emph{New Journal of Physics}, 18\penalty0 (9), 2016.
\newblock URL \url{http://stacks.iop.org/1367-2630/18/i=9/a=093052}.

\bibitem[Kaelbling et~al.(1996)Kaelbling, Littman, and
  Moore]{kaelbling1996reinforcement}
Leslie~Pack Kaelbling, Michael~L Littman, and Andrew~W Moore.
\newblock Reinforcement learning: A survey.
\newblock \emph{Journal of artificial intelligence research}, 4:\penalty0
  237--285, 1996.

\bibitem[Kahneman(2011)]{kahneman2011thinking}
Daniel Kahneman.
\newblock \emph{Thinking, fast and slow}.
\newblock Farrar, Straus and Giroux, New York, 2011.
\newblock ISBN 9780374275631 0374275637.
\newblock URL
  \url{https://www.amazon.de/Thinking-Fast-Slow-Daniel-Kahneman/dp/0374275637/ref=wl_it_dp_o_pdT1_nS_nC?ie=UTF8&colid=151193SNGKJT9&coliid=I3OCESLZCVDFL7}.

\bibitem[Karahan et~al.(2016)Karahan, Yildirum, Kirtac, Rende, Butun, and
  Ekenel]{karahan2016image}
Samil Karahan, Merve~Kilinc Yildirum, Kadir Kirtac, Ferhat~Sukru Rende,
  Gultekin Butun, and Hazim~Kemal Ekenel.
\newblock How image degradations affect deep cnn-based face recognition?
\newblock In \emph{2016 International Conference of the Biometrics Special
  Interest Group (BIOSIG)}, pages 1--5. IEEE, 2016.

\bibitem[Karimi et~al.(2020)Karimi, von Kügelgen, Schölkopf, and
  Valera]{karimi2020algorithmic}
Amir-Hossein Karimi, Julius von Kügelgen, Bernhard Schölkopf, and Isabel
  Valera.
\newblock Algorithmic recourse under imperfect causal knowledge: a
  probabilistic approach.
\newblock \emph{arXiv 2006.06831}, 2020.
\newblock Published at NeurIPS.

\bibitem[Ke et~al.(2020)Ke, Bilaniuk, Goyal, Bauer, Larochelle, Schölkopf,
  Mozer, Pal, and Bengio]{ke2019learning}
Nan~Rosemary Ke, Olexa Bilaniuk, Anirudh Goyal, Stefan Bauer, Hugo Larochelle,
  Bernhard Schölkopf, Michael Mozer, Chris Pal, and Yoshua Bengio.
\newblock Learning neural causal models from unknown interventions.
\newblock \emph{arXiv preprint 1910.01075v2}, 2020.

\bibitem[Khajehnejad et~al.(2019)Khajehnejad, Tabibian, Schölkopf, Singla, and
  Gomez-Rodriguez]{1905.09239}
Moein Khajehnejad, Behzad Tabibian, Bernhard Schölkopf, Adish Singla, and
  Manuel Gomez-Rodriguez.
\newblock Optimal decision making under strategic behavior.
\newblock \emph{arXiv preprint 1905.09239}, 2019.

\bibitem[Kilbertus et~al.(2017)Kilbertus, Rojas~Carulla, Parascandolo, Hardt,
  Janzing, and Sch\"{o}lkopf]{Kilbertusetal17}
N.~Kilbertus, M.~Rojas~Carulla, G.~Parascandolo, M.~Hardt, D.~Janzing, and
  B.~Sch\"{o}lkopf.
\newblock Avoiding discrimination through causal reasoning.
\newblock In \emph{Advances in Neural Information Processing Systems 30}, pages
  656--666, 2017.

\bibitem[Kilbertus et~al.(2018)Kilbertus, Parascandolo, and
  Schölkopf]{KilParSch19arxiv}
Niki Kilbertus, Giambattista Parascandolo, and Bernhard Schölkopf.
\newblock Generalization in anti-causal learning.
\newblock \emph{arXiv preprint 1812.00524}, 2018.

\bibitem[Kim and Mnih(2018)]{kim2018disentangling}
Hyunjik Kim and Andriy Mnih.
\newblock Disentangling by factorising.
\newblock In \emph{International Conference on Machine Learning}, 2018.

\bibitem[Kipf et~al.(2018)Kipf, Fetaya, Wang, Welling, and
  Zemel]{kipf2018neural}
Thomas Kipf, Ethan Fetaya, Kuan-Chieh Wang, Max Welling, and Richard Zemel.
\newblock Neural relational inference for interacting systems.
\newblock In \emph{International Conference on Machine Learning}, pages
  2688--2697, 2018.

\bibitem[Kolesnikov et~al.(2019)Kolesnikov, Beyer, Zhai, Puigcerver, Yung,
  Gelly, and Houlsby]{kolesnikov2019big}
Alexander Kolesnikov, Lucas Beyer, Xiaohua Zhai, Joan Puigcerver, Jessica Yung,
  Sylvain Gelly, and Neil Houlsby.
\newblock Big transfer (bit): General visual representation learning.
\newblock \emph{arXiv preprint 1912.11370}, 2019.

\bibitem[Kosiorek et~al.(2018)Kosiorek, Kim, Teh, and
  Posner]{kosiorek2018sequential}
Adam Kosiorek, Hyunjik Kim, Yee~Whye Teh, and Ingmar Posner.
\newblock Sequential attend, infer, repeat: Generative modelling of moving
  objects.
\newblock \emph{Advances in Neural Information Processing Systems},
  31:\penalty0 8606--8616, 2018.

\bibitem[Kpotufe et~al.(2014)Kpotufe, Sgouritsa, Janzing, and
  Sch{\"{o}}lkopf]{Kpotufe14}
S.~Kpotufe, E.~Sgouritsa, D.~Janzing, and B.~Sch{\"{o}}lkopf.
\newblock Consistency of causal inference under the additive noise model.
\newblock In \emph{Proceedings of the 31th International Conference on Machine
  Learning}, pages 478--486, 2014.

\bibitem[Krizhevsky et~al.(2012)Krizhevsky, Sutskever, and
  Hinton]{krizhevsky2012imagenet}
Alex Krizhevsky, Ilya Sutskever, and Geoffrey~E Hinton.
\newblock Imagenet classification with deep convolutional neural networks.
\newblock In \emph{Advances in neural information processing systems}, pages
  1097--1105, 2012.

\bibitem[Kulkarni et~al.(2019)Kulkarni, Gupta, Ionescu, Borgeaud, Reynolds,
  Zisserman, and Mnih]{kulkarni2019unsupervised}
Tejas~D Kulkarni, Ankush Gupta, Catalin Ionescu, Sebastian Borgeaud, Malcolm
  Reynolds, Andrew Zisserman, and Volodymyr Mnih.
\newblock Unsupervised learning of object keypoints for perception and control.
\newblock In \emph{Advances in Neural Information Processing Systems}, pages
  10723--10733, 2019.

\bibitem[Kusner et~al.(2017)Kusner, Loftus, Russell, and Silva]{NIPS2017_6995}
Matt~J Kusner, Joshua Loftus, Chris Russell, and Ricardo Silva.
\newblock Counterfactual fairness.
\newblock In \emph{Advances in Neural Information Processing Systems 30}, pages
  4066--4076. Curran Associates, Inc., 2017.

\bibitem[Ladick{\`y} et~al.(2015)Ladick{\`y}, Jeong, Solenthaler, Pollefeys,
  and Gross]{ladicky2015data}
L'ubor Ladick{\`y}, SoHyeon Jeong, Barbara Solenthaler, Marc Pollefeys, and
  Markus Gross.
\newblock Data-driven fluid simulations using regression forests.
\newblock \emph{ACM Transactions on Graphics (TOG)}, 34\penalty0 (6):\penalty0
  1--9, 2015.

\bibitem[Lake et~al.(2017)Lake, Ullman, Tenenbaum, and
  Gershman]{lake2017building}
Brenden~M Lake, Tomer~D Ullman, Joshua~B Tenenbaum, and Samuel~J Gershman.
\newblock Building machines that learn and think like people.
\newblock \emph{Behavioral and brain sciences}, 40, 2017.

\bibitem[Landman et~al.(1995)Landman, Vandewater, Stewart, and
  Malley]{landman1995missed}
Janet Landman, Elizabeth~A Vandewater, Abigail~J Stewart, and Janet~E Malley.
\newblock Missed opportunities: Psychological ramifications of counterfactual
  thought in midlife women.
\newblock \emph{Journal of Adult Development}, 2\penalty0 (2):\penalty0 87--97,
  1995.

\bibitem[Lange et~al.(2012)Lange, Gabel, and Riedmiller]{Lange2012}
Sascha Lange, Thomas Gabel, and Martin Riedmiller.
\newblock Batch reinforcement learning.
\newblock In Marco Wiering and Martijn van Otterlo, editors,
  \emph{Reinforcement Learning: State-of-the-Art}, pages 45--73. Springer,
  Berlin, Heidelberg, 2012.

\bibitem[Lauritzen(1996)]{Lauritzen1996}
S.~L. Lauritzen.
\newblock \emph{Graphical Models}.
\newblock Oxford University Press, New York, NY, 1996.

\bibitem[LeCun et~al.(2015)LeCun, Bengio, and Hinton]{LeCBenHin15}
Yann LeCun, Yoshua Bengio, and Geoffrey Hinton.
\newblock Deep learning.
\newblock \emph{{N}ature}, 521\penalty0 (7553):\penalty0 436--444, 2015.

\bibitem[Leeb et~al.(2020)Leeb, Annadani, Bauer, and Schölkopf]{Leeb-SAE}
Felix Leeb, Yashas Annadani, Stefan Bauer, and Bernhard Schölkopf.
\newblock Structural autoencoders improve representations for generation and
  transfer.
\newblock \emph{arXiv preprint 2006.07796}, 2020.

\bibitem[Levine et~al.(2020)Levine, Kumar, Tucker, and Fu]{levine2020offline}
Sergey Levine, Aviral Kumar, George Tucker, and Justin Fu.
\newblock Offline reinforcement learning: Tutorial, review, and perspectives on
  open problems.
\newblock \emph{arXiv preprint 2005.01643}, 2020.

\bibitem[Lewis(1974)]{lewis1974causation}
David Lewis.
\newblock Causation.
\newblock \emph{The journal of philosophy}, 70\penalty0 (17):\penalty0
  556--567, 1974.

\bibitem[Li et~al.(2018{\natexlab{a}})Li, Gong, Tian, Liu, and Tao]{1807.08479}
Ya~Li, Mingming Gong, Xinmei Tian, Tongliang Liu, and Dacheng Tao.
\newblock Domain generalization via conditional invariant representation.
\newblock \emph{arXiv preprint 1807.08479}, 2018{\natexlab{a}}.

\bibitem[Li et~al.(2018{\natexlab{b}})Li, Tian, Gong, Liu, Liu, Zhang, and
  Tao]{Li_2018_ECCV}
Ya~Li, Xinmei Tian, Mingming Gong, Yajing Liu, Tongliang Liu, Kun Zhang, and
  Dacheng Tao.
\newblock Deep domain generalization via conditional invariant adversarial
  networks.
\newblock In \emph{The European Conference on Computer Vision (ECCV)},
  2018{\natexlab{b}}.

\bibitem[Lim et~al.(2019)Lim, Kim, Kim, Kim, and Kim]{lim2019fast}
Sungbin Lim, Ildoo Kim, Taesup Kim, Chiheon Kim, and Sungwoong Kim.
\newblock Fast autoaugment.
\newblock In \emph{Advances in Neural Information Processing Systems}, pages
  6665--6675, 2019.

\bibitem[Lin et~al.(2019)Lin, Wu, Peri, Sun, Singh, Deng, Jiang, and
  Ahn]{lin2019space}
Zhixuan Lin, Yi-Fu Wu, Skand~Vishwanath Peri, Weihao Sun, Gautam Singh, Fei
  Deng, Jindong Jiang, and Sungjin Ahn.
\newblock Space: Unsupervised object-oriented scene representation via spatial
  attention and decomposition.
\newblock In \emph{International Conference on Learning Representations}, 2019.

\bibitem[Lipton et~al.(2018)Lipton, Wang, and Smola]{1802.03916}
Zachary~C. Lipton, Yu-Xiang Wang, and Alex Smola.
\newblock Detecting and correcting for label shift with black box predictors.
\newblock \emph{arXiv preprint 1802.03916}, 2018.

\bibitem[Locatello et~al.(2019{\natexlab{a}})Locatello, Abbati, Rainforth,
  Bauer, Sch{\"o}lkopf, and Bachem]{locatello2019fairness}
Francesco Locatello, Gabriele Abbati, Tom Rainforth, Stefan Bauer, Bernhard
  Sch{\"o}lkopf, and Olivier Bachem.
\newblock On the fairness of disentangled representations.
\newblock In \emph{Advances in Neural Information Processing Systems}, pages
  14544--14557, 2019{\natexlab{a}}.

\bibitem[Locatello et~al.(2019{\natexlab{b}})Locatello, Bauer, Lucic, Rätsch,
  Gelly, Schölkopf, and Bachem]{1811.12359}
Francesco Locatello, Stefan Bauer, Mario Lucic, Gunnar Rätsch, Sylvain Gelly,
  Bernhard Schölkopf, and Olivier Bachem.
\newblock Challenging common assumptions in the unsupervised learning of
  disentangled representations.
\newblock \emph{Proceedings of the 36th International Conference on Machine
  Learning}, 2019{\natexlab{b}}.

\bibitem[Locatello et~al.(2020{\natexlab{a}})Locatello, Poole, R{\"a}tsch,
  Sch{\"o}lkopf, Bachem, and Tschannen]{locatello2020weakly}
Francesco Locatello, Ben Poole, Gunnar R{\"a}tsch, Bernhard Sch{\"o}lkopf,
  Olivier Bachem, and Michael Tschannen.
\newblock Weakly-supervised disentanglement without compromises.
\newblock In \emph{Proceedings of the 37th International Conference on Machine
  Learning ({ICML})}, 2020{\natexlab{a}}.

\bibitem[Locatello et~al.(2020{\natexlab{b}})Locatello, Weissenborn,
  Unterthiner, Mahendran, Heigold, Uszkoreit, Dosovitskiy, and
  Kipf]{locatello2020object}
Francesco Locatello, Dirk Weissenborn, Thomas Unterthiner, Aravindh Mahendran,
  Georg Heigold, Jakob Uszkoreit, Alexey Dosovitskiy, and Thomas Kipf.
\newblock Object-centric learning with slot attention.
\newblock In \emph{Advances in Neural Information Processing Systems},
  2020{\natexlab{b}}.

\bibitem[Lopez-Paz et~al.(2015)Lopez-Paz, Muandet, Sch{\"o}lkopf, and
  Tolstikhin]{LopMuaSchTol15}
D.~Lopez-Paz, K.~Muandet, B.~Sch{\"o}lkopf, and I.~Tolstikhin.
\newblock Towards a learning theory of cause-effect inference.
\newblock In \emph{Proceedings of the 32nd International Conference on Machine
  Learning}, pages 1452--1461, 2015.

\bibitem[Lopez-Paz et~al.(2017)Lopez-Paz, Nishihara, Chintala, Sch{\"o}lkopf,
  and Bottou]{LopNisChiSchBot17}
D.~Lopez-Paz, R.~Nishihara, S.~Chintala, B.~Sch{\"o}lkopf, and L.~Bottou.
\newblock Discovering causal signals in images.
\newblock In \emph{IEEE Conference on Computer Vision and Pattern Recognition
  (CVPR)}, pages 58--66, 2017.

\bibitem[Lorenz(1973)]{Lorenz73}
K.~Lorenz.
\newblock \emph{Die {R}{\"u}ckseite des Spiegels}.
\newblock R.~Piper \& Co.~Verlag, 1973.

\bibitem[Lu et~al.(2018)Lu, Schölkopf, and Hernández-Lobato]{1812.10576}
Chaochao Lu, Bernhard Schölkopf, and José~Miguel Hernández-Lobato.
\newblock Deconfounding reinforcement learning in observational settings.
\newblock \emph{arXiv preprint 1812.10576}, 2018.

\bibitem[Lu et~al.(2020)Lu, Huang, Wang, Hernández-Lobato, Zhang, and
  Schölkopf]{2012.09092}
Chaochao Lu, Biwei Huang, Ke~Wang, José~Miguel Hernández-Lobato, Kun Zhang,
  and Bernhard Schölkopf.
\newblock Sample-efficient reinforcement learning via counterfactual-based data
  augmentation.
\newblock \emph{arXiv preprint 2012.09092}, 2020.

\bibitem[Lundervold and Lundervold(2019)]{lundervold2019overview}
Alexander~Selvikv{\aa}g Lundervold and Arvid Lundervold.
\newblock An overview of deep learning in medical imaging focusing on {MRI}.
\newblock \emph{Zeitschrift f{\"u}r Medizinische Physik}, 29\penalty0
  (2):\penalty0 102--127, 2019.

\bibitem[Magliacane et~al.(2018)Magliacane, van Ommen, Claassen, Bongers,
  Versteeg, and Mooij]{1707.06422}
Sara Magliacane, Thijs van Ommen, Tom Claassen, Stephan Bongers, Philip
  Versteeg, and Joris~M. Mooij.
\newblock Domain adaptation by using causal inference to predict invariant
  conditional distributions.
\newblock In \emph{Proc. NeurIPS}, 2018.

\bibitem[Matthews(2000)]{matthews2000storks}
Robert Matthews.
\newblock Storks deliver babies (p= 0.008).
\newblock \emph{Teaching Statistics}, 22\penalty0 (2):\penalty0 36--38, 2000.

\bibitem[Meinshausen(2018)]{meinshausen2018causality}
Nicolai Meinshausen.
\newblock Causality from a distributional robustness point of view.
\newblock In \emph{2018 IEEE Data Science Workshop (DSW)}, pages 6--10. IEEE,
  2018.

\bibitem[Michaelis et~al.(2019)Michaelis, Mitzkus, Geirhos, Rusak, Bringmann,
  Ecker, Bethge, and Brendel]{michaelis2019benchmarking}
Claudio Michaelis, Benjamin Mitzkus, Robert Geirhos, Evgenia Rusak, Oliver
  Bringmann, Alexander~S Ecker, Matthias Bethge, and Wieland Brendel.
\newblock Benchmarking robustness in object detection: Autonomous driving when
  winter is coming.
\newblock \emph{arXiv preprint 1907.07484}, 2019.

\bibitem[Mnih et~al.(2015)Mnih, Kavukcuoglu, Silver, Rusu, Veness, Bellemare,
  Graves, Riedmiller, Fidjeland, Ostrovski, Petersen, Beattie, Sadik,
  Antonoglou, King, Kumaran, Wierstra, Legg, and Hassabis]{mnih2015human}
Volodymyr Mnih, Koray Kavukcuoglu, David Silver, Andrei~A. Rusu, Joel Veness,
  Marc~G. Bellemare, Alex Graves, Martin Riedmiller, Andreas~K. Fidjeland,
  Georg Ostrovski, Stig Petersen, Charles Beattie, Amir Sadik, Ioannis
  Antonoglou, Helen King, Dharshan Kumaran, Daan Wierstra, Shane Legg, and
  Demis Hassabis.
\newblock Human-level control through deep reinforcement learning.
\newblock \emph{{N}ature}, 518\penalty0 (7540):\penalty0 529--533, 2015.

\bibitem[Montet et~al.(2015)Montet, Morton, Foreman-Mackey, Johnson, Hogg,
  Bowler, Latham, Bieryla, and Mann]{Montet_2015}
B.~T. Montet, T.~D. Morton, D.~Foreman-Mackey, J.~A. Johnson, D.~W. Hogg, B.~P.
  Bowler, D.~W. Latham, A.~Bieryla, and A.~W. Mann.
\newblock Stellar and planetary properties of {K2} campaign 1 candidates and
  validation of 17 planets, including a planet receiving earth-like insolation.
\newblock \emph{The Astrophysical Journal}, 809\penalty0 (1):\penalty0 25,
  2015.

\bibitem[Mooij et~al.(2013)Mooij, Janzing, and Sch{\"o}lkopf]{MooijJ2013}
J.~Mooij, D.~Janzing, and B.~Sch{\"o}lkopf.
\newblock From ordinary differential equations to structural causal models: the
  deterministic case.
\newblock In A.~Nicholson and P.~Smyth, editors, \emph{Proceedings of the
  Twenty-Ninth Conference Annual Conference on Uncertainty in Artificial
  Intelligence}, pages 440--448, Corvallis, OR, 2013. AUAI Press.
\newblock URL
  \url{http://www.is.tuebingen.mpg.de/fileadmin/user_upload/files/publications/2013/MooijJS2013-uai.pdf}.

\bibitem[Mooij et~al.(2009)Mooij, Janzing, Peters, and
  Sch\"{o}lkopf]{Mooij2009}
J.~M. Mooij, D.~Janzing, J.~Peters, and B.~Sch\"{o}lkopf.
\newblock Regression by dependence minimization and its application to causal
  inference.
\newblock In \emph{Proceedings of the 26th International Conference on Machine
  Learning ({ICML})}, pages 745--752, 2009.

\bibitem[Mooij et~al.(2011)Mooij, Janzing, Heskes, and Sch{\"o}lkopf]{Mooij11}
J.~M. Mooij, D.~Janzing, T.~Heskes, and B.~Sch{\"o}lkopf.
\newblock On causal discovery with cyclic additive noise models.
\newblock In \emph{{A}dvances in {N}eural {I}nformation {P}rocessing {S}ystems
  24 ({NIPS})}, 2011.

\bibitem[Mooij et~al.(2016)Mooij, Peters, Janzing, Zscheischler, and
  Sch\"olkopf]{Mooijetal16}
J.~M. Mooij, J.~Peters, D.~Janzing, J.~Zscheischler, and B.~Sch\"olkopf.
\newblock Distinguishing cause from effect using observational data: methods
  and benchmarks.
\newblock \emph{Journal of Machine Learning Research}, 17\penalty0
  (32):\penalty0 1--102, 2016.

\bibitem[Mrowca et~al.(2018)Mrowca, Zhuang, Wang, Haber, Fei-Fei, Tenenbaum,
  and Yamins]{mrowca2018flexible}
Damian Mrowca, Chengxu Zhuang, Elias Wang, Nick Haber, Li~Fei-Fei, Josh
  Tenenbaum, and Daniel L~K Yamins.
\newblock Flexible neural representation for physics prediction.
\newblock In \emph{Advances in Neural Information Processing Systems}, pages
  8799--8810, 2018.

\bibitem[Oh et~al.(2015)Oh, Guo, Lee, Lewis, and Singh]{oh2015action}
Junhyuk Oh, Xiaoxiao Guo, Honglak Lee, Richard~L Lewis, and Satinder Singh.
\newblock Action-conditional video prediction using deep networks in atari
  games.
\newblock In \emph{Advances in neural information processing systems}, pages
  2863--2871, 2015.

\bibitem[Oord et~al.(2018)Oord, Li, and Vinyals]{oord2018representation}
Aaron van~den Oord, Yazhe Li, and Oriol Vinyals.
\newblock Representation learning with contrastive predictive coding.
\newblock \emph{arXiv preprint 1807.03748}, 2018.

\bibitem[Parascandolo et~al.(2017)Parascandolo, Rojas-Carulla, Kilbertus, and
  Sch{\"o}lkopf]{ParRojKilSch17}
G.~Parascandolo, M.~Rojas-Carulla, N.~Kilbertus, and B.~Sch{\"o}lkopf.
\newblock Learning independent causal mechanisms.
\newblock In \emph{Workshop: Learning Disentangled Representations: from
  Perception to Control at the 31st Conference on Neural Information Processing
  Systems (NIPS)}, 2017.

\bibitem[Parascandolo et~al.(2018)Parascandolo, Kilbertus, Rojas-Carulla, and
  Sch{\"o}lkopf]{ParKilRojSch18}
G.~Parascandolo, N.~Kilbertus, M.~Rojas-Carulla, and B.~Sch{\"o}lkopf.
\newblock Learning independent causal mechanisms.
\newblock In \emph{Proceedings of the 35th International Conference on Machine
  Learning, PMLR 80:4036-4044}, 2018.

\bibitem[Parascandolo et~al.(2021)Parascandolo, Neitz, ORVIETO, Gresele, and
  Sch{\"o}lkopf]{parascandolo2021learning}
Giambattista Parascandolo, Alexander Neitz, ANTONIO ORVIETO, Luigi Gresele, and
  Bernhard Sch{\"o}lkopf.
\newblock Learning explanations that are hard to vary.
\newblock In \emph{International Conference on Learning Representations}, 2021.

\bibitem[Pearl(2009{\natexlab{a}})]{Pearl2009}
J.~Pearl.
\newblock \emph{Causality: Models, Reasoning, and Inference}.
\newblock Cambridge University Press, New York, NY, 2nd edition,
  2009{\natexlab{a}}.

\bibitem[Pearl(2009{\natexlab{b}})]{Pearl2009forbes}
J.~Pearl.
\newblock Giving computers free will.
\newblock \emph{Forbes}, 2009{\natexlab{b}}.

\bibitem[Pearl and Bareinboim(2015)]{1503.01603}
Judea Pearl and Elias Bareinboim.
\newblock External validity: From do-calculus to transportability across
  populations.
\newblock \emph{arXiv preprint 1503.01603}, 2015.

\bibitem[Peters et~al.(2011)Peters, Mooij, Janzing, and
  Sch\"{o}lkopf]{Peters2011b}
J.~Peters, J.~M. Mooij, D.~Janzing, and B.~Sch\"{o}lkopf.
\newblock Identifiability of causal graphs using functional models.
\newblock In \emph{Proceedings of the 27th Annual Conference on {U}ncertainty
  in {A}rtificial {I}ntelligence ({UAI})}, pages 589--598, 2011.

\bibitem[Peters et~al.(2014)Peters, Mooij, Janzing, and
  Sch{\"o}lkopf]{PetMooJanSch14}
J.~Peters, J.~M. Mooij, D.~Janzing, and B.~Sch{\"o}lkopf.
\newblock Causal discovery with continuous additive noise models.
\newblock \emph{Journal of Machine Learning Research}, 15:\penalty0 2009--2053,
  2014.
\newblock URL \url{http://jmlr.org/papers/v15/peters14a.html}.

\bibitem[Peters et~al.(2017)Peters, Janzing, and Sch{\"o}lkopf]{PetJanSch17}
J.~Peters, D.~Janzing, and B.~Sch{\"o}lkopf.
\newblock \emph{Elements of Causal Inference - Foundations and Learning
  Algorithms}.
\newblock MIT Press, Cambridge, MA, USA, 2017.

\bibitem[Peters et~al.(2016)Peters, B{\"u}hlmann, and
  Meinshausen]{peters2016causal}
Jonas Peters, Peter B{\"u}hlmann, and Nicolai Meinshausen.
\newblock Causal inference by using invariant prediction: identification and
  confidence intervals.
\newblock \emph{Journal of the Royal Statistical Society: Series B (Statistical
  Methodology)}, 78\penalty0 (5):\penalty0 947--1012, 2016.

\bibitem[Peters et~al.(2020)Peters, Bauer, and Pfister]{peters2020causal}
Jonas Peters, Stefan Bauer, and Niklas Pfister.
\newblock Causal models for dynamical systems.
\newblock \emph{arXiv preprint 2001.06208}, 2020.

\bibitem[Pfister et~al.(2018)Pfister, B{\"u}hlmann, Sch{\"o}lkopf, and
  Peters]{PfiBuhSchPet18}
N.~Pfister, P.~B{\"u}hlmann, B.~Sch{\"o}lkopf, and J.~Peters.
\newblock Kernel-based tests for joint independence.
\newblock \emph{Journal of the Royal Statistical Society: Series B (Statistical
  Methodology)}, 80\penalty0 (1):\penalty0 5--31, 2018.

\bibitem[Pfister et~al.(2019{\natexlab{a}})Pfister, Bauer, and
  Peters]{pfister2019learning}
Niklas Pfister, Stefan Bauer, and Jonas Peters.
\newblock Learning stable and predictive structures in kinetic systems.
\newblock \emph{Proceedings of the National Academy of Sciences}, 116\penalty0
  (51):\penalty0 25405--25411, 2019{\natexlab{a}}.

\bibitem[Pfister et~al.(2019{\natexlab{b}})Pfister, B{\"u}hlmann, and
  Peters]{pfister2019invariant}
Niklas Pfister, Peter B{\"u}hlmann, and Jonas Peters.
\newblock Invariant causal prediction for sequential data.
\newblock \emph{Journal of the American Statistical Association}, 114\penalty0
  (527):\penalty0 1264--1276, 2019{\natexlab{b}}.

\bibitem[Priol et~al.(2020)Priol, Harikandeh, Bengio, and
  Lacoste-Julien]{priol2020analysis}
R{\'e}mi~Le Priol, Reza~Babanezhad Harikandeh, Yoshua Bengio, and Simon
  Lacoste-Julien.
\newblock An analysis of the adaptation speed of causal models.
\newblock \emph{arXiv preprint 2005.09136}, 2020.

\bibitem[Rabanser et~al.(2018)Rabanser, Günnemann, and Lipton]{1810.11953}
Stephan Rabanser, Stephan Günnemann, and Zachary~C. Lipton.
\newblock Failing loudly: An empirical study of methods for detecting dataset
  shift.
\newblock \emph{arXiv preprint 1810.11953}, 2018.

\bibitem[Radford et~al.(2018)Radford, Narasimhan, Salimans, and
  Sutskever]{radford2018improving}
Alec Radford, Karthik Narasimhan, Tim Salimans, and Ilya Sutskever.
\newblock Improving language understanding by generative pre-training, 2018.

\bibitem[Rahaman et~al.(2021)Rahaman, Goyal, Gondal, Wuthrich, Bauer, Sharma,
  Bengio, and Sch{\"o}lkopf]{rahaman2021spatially}
Nasim Rahaman, Anirudh Goyal, Muhammad~Waleed Gondal, Manuel Wuthrich, Stefan
  Bauer, Yash Sharma, Yoshua Bengio, and Bernhard Sch{\"o}lkopf.
\newblock Spatially structured recurrent modules.
\newblock In \emph{International Conference on Learning Representations}, 2021.

\bibitem[Reichenbach(1956)]{Reichenbach1956}
H.~Reichenbach.
\newblock \emph{The Direction of Time}.
\newblock University of California Press, Berkeley, CA, 1956.

\bibitem[Reichert and Slate(1999)]{reichert1999reflective}
Laine~K Reichert and John~R Slate.
\newblock Reflective learning: The use of “if only...” statements to
  improve performance.
\newblock \emph{Social Psychology of Education}, 3\penalty0 (4):\penalty0
  261--275, 1999.

\bibitem[Rezende et~al.(2020)Rezende, Danihelka, Papamakarios, Ke, Jiang,
  Weber, Gregor, Merzic, Viola, Wang, et~al.]{rezende2020causally}
Danilo~J Rezende, Ivo Danihelka, George Papamakarios, Nan~Rosemary Ke, Ray
  Jiang, Theophane Weber, Karol Gregor, Hamza Merzic, Fabio Viola, Jane Wang,
  et~al.
\newblock Causally correct partial models for reinforcement learning.
\newblock \emph{arXiv preprint 2002.02836}, 2020.

\bibitem[Richens et~al.(2020)Richens, Lee, and Johri]{Richens2020vs}
Jonathan~G Richens, Ciar{\'a}n~M Lee, and Saurabh Johri.
\newblock {Improving the accuracy of medical diagnosis with causal machine
  learning}.
\newblock \emph{{Nature Communications}}, 11\penalty0 (1):\penalty0 3923, 2020.

\bibitem[Ridgeway and Mozer(2018)]{ridgeway2018learning}
Karl Ridgeway and Michael~C Mozer.
\newblock Learning deep disentangled embeddings with the f-statistic loss.
\newblock In \emph{Advances in Neural Information Processing Systems}, pages
  185--194, 2018.

\bibitem[Roese(1994)]{roese1994functional}
Neal~J Roese.
\newblock The functional basis of counterfactual thinking.
\newblock \emph{Journal of personality and Social Psychology}, 66\penalty0
  (5):\penalty0 805, 1994.

\bibitem[Rojas-Carulla et~al.(2018)Rojas-Carulla, Sch{\"o}lkopf, Turner, and
  Peters]{RojSchTurPet18}
M.~Rojas-Carulla, B.~Sch{\"o}lkopf, R.~Turner, and J.~Peters.
\newblock Invariant models for causal transfer learning.
\newblock \emph{Journal of Machine Learning Research}, 19\penalty0
  (36):\penalty0 1--34, 2018.

\bibitem[Rolinek et~al.(2019)Rolinek, Zietlow, and
  Martius]{rolinek2019variational}
Michal Rolinek, Dominik Zietlow, and Georg Martius.
\newblock Variational autoencoders pursue {PCA} directions (by accident).
\newblock In \emph{Proceedings of the IEEE Conference on Computer Vision and
  Pattern Recognition}, pages 12406--12415, 2019.

\bibitem[Roy et~al.(2018)Roy, Ghosh, Bhattacharya, and Pal]{roy2018effects}
Prasun Roy, Subhankar Ghosh, Saumik Bhattacharya, and Umapada Pal.
\newblock Effects of degradations on deep neural network architectures.
\newblock \emph{arXiv preprint 1807.10108}, 2018.

\bibitem[Rubenstein et~al.(2017)Rubenstein, Weichwald, Bongers, Mooij, Janzing,
  Grosse-Wentrup, and Sch{\"o}lkopf]{Rubensteinetal17}
P.~K. Rubenstein, S.~Weichwald, S.~Bongers, J.~M. Mooij, D.~Janzing,
  M.~Grosse-Wentrup, and B.~Sch{\"o}lkopf.
\newblock Causal consistency of structural equation models.
\newblock In \emph{Proceedings of the Thirty-Third Conference on Uncertainty in
  Artificial Intelligence}, pages 808--817, 2017.

\bibitem[Rubenstein et~al.(2018)Rubenstein, Bongers, Sch{\"o}lkopf, and
  Mooij]{RubBonMooSch18}
P.~K. Rubenstein, S.~Bongers, B.~Sch{\"o}lkopf, and J.~M. Mooij.
\newblock From deterministic {ODEs} to dynamic structural causal models.
\newblock In \emph{Proceedings of the 34th Conference on Uncertainty in
  Artificial Intelligence (UAI)}, 2018.

\bibitem[Ruder(2017)]{ruder2017overview}
Sebastian Ruder.
\newblock An overview of multi-task learning in deep neural networks.
\newblock \emph{arXiv preprint 1706.05098}, 2017.

\bibitem[Russell and Norvig(2002)]{russell2002artificial}
Stuart Russell and Peter Norvig.
\newblock \emph{Artificial intelligence: a modern approach}.
\newblock Prentice Hall, 2002.

\bibitem[Sanchez-Gonzalez et~al.(2020)Sanchez-Gonzalez, Godwin, Pfaff, Ying,
  Leskovec, and Battaglia]{sanchez2020learning}
Alvaro Sanchez-Gonzalez, Jonathan Godwin, Tobias Pfaff, Rex Ying, Jure
  Leskovec, and Peter~W Battaglia.
\newblock Learning to simulate complex physics with graph networks.
\newblock \emph{arXiv preprint 2002.09405}, 2020.

\bibitem[Santoro et~al.(2017)Santoro, Raposo, Barrett, Malinowski, Pascanu,
  Battaglia, and Lillicrap]{santoro2017simple}
Adam Santoro, David Raposo, David~G Barrett, Mateusz Malinowski, Razvan
  Pascanu, Peter Battaglia, and Timothy Lillicrap.
\newblock A simple neural network module for relational reasoning.
\newblock In \emph{Advances in neural information processing systems}, pages
  4967--4976, 2017.

\bibitem[Schmidhuber(1987)]{schmidhuber1987evolutionary}
J{\"u}rgen Schmidhuber.
\newblock \emph{Evolutionary principles in self-referential learning, or on
  learning how to learn: the meta-meta-... hook}.
\newblock PhD thesis, Technische Universit{\"a}t M{\"u}nchen, 1987.

\bibitem[Schmidhuber(1991)]{schmidhuber1991curious}
J{\"u}rgen Schmidhuber.
\newblock Curious model-building control systems.
\newblock In \emph{Proc. international joint conference on neural networks},
  pages 1458--1463, 1991.

\bibitem[Sch{\"o}lkopf(2015)]{schoelkopf15}
B.~Sch{\"o}lkopf.
\newblock Artificial intelligence: Learning to see and act.
\newblock \emph{{N}ature}, 518\penalty0 (7540):\penalty0 486--487, 2015.

\bibitem[Sch{\"o}lkopf(2017)]{Schoelkopf2017icml}
B.~Sch{\"o}lkopf.
\newblock Causal learning, 2017.
\newblock Invited Talk, 34th International Conference on Machine Learning
  ({ICML}), \url{https://vimeo.com/238274659}.

\bibitem[Sch{\"o}lkopf and Smola(2002)]{SchSmo02}
B.~Sch{\"o}lkopf and A.~J. Smola.
\newblock \emph{Learning with Kernels}.
\newblock MIT Press, Cambridge, MA, 2002.

\bibitem[Sch{\"o}lkopf et~al.(2012)Sch{\"o}lkopf, Janzing, Peters, Sgouritsa,
  Zhang, and Mooij]{Schoelkopf2012}
B.~Sch{\"o}lkopf, D.~Janzing, J.~Peters, E.~Sgouritsa, K.~Zhang, and J.~M.
  Mooij.
\newblock On causal and anticausal learning.
\newblock In \emph{Proceedings of the 29th International Conference on Machine
  Learning ({ICML})}, pages 1255--1262, 2012.

\bibitem[Sch{\"o}lkopf et~al.(2016{\natexlab{a}})Sch{\"o}lkopf, Hogg, Wang,
  Foreman-Mackey, Janzing, Simon-Gabriel, and Peters]{Scholkopfetal16}
B.~Sch{\"o}lkopf, D.~Hogg, D.~Wang, D.~Foreman-Mackey, D.~Janzing, C.-J.
  Simon-Gabriel, and J.~Peters.
\newblock Modeling confounding by half-sibling regression.
\newblock \emph{Proceedings of the National Academy of Science (PNAS)},
  113\penalty0 (27):\penalty0 7391--7398, 2016{\natexlab{a}}.

\bibitem[Sch{\"o}lkopf et~al.(2016{\natexlab{b}})Sch{\"o}lkopf, Janzing, and
  Lopez-Paz]{SchJanLop16}
B.~Sch{\"o}lkopf, D.~Janzing, and D.~Lopez-Paz.
\newblock Causal and statistical learning.
\newblock In \emph{Oberwolfach Reports}, volume 13(3), pages 1896--1899,
  2016{\natexlab{b}}.
\newblock \doi{10.14760/OWR-2016-33}.
\newblock URL \url{https://publications.mfo.de/handle/mfo/3537}.

\bibitem[Sch\"olkopf(2019)]{1911.10500}
Bernhard Sch\"olkopf.
\newblock Causality for machine learning.
\newblock \emph{arXiv preprint 1911.10500}, 2019.

\bibitem[Schott et~al.(2019)Schott, Rauber, Bethge, and
  Brendel]{schott2018towards}
Lukas Schott, Jonas Rauber, Matthias Bethge, and Wieland Brendel.
\newblock Towards the first adversarially robust neural network model on
  {MNIST}.
\newblock In \emph{International Conference on Learning Representations}, 2019.
\newblock URL \url{https://openreview.net/forum?id=S1EHOsC9tX}.

\bibitem[Schrittwieser et~al.(2019)Schrittwieser, Antonoglou, Hubert, Simonyan,
  Sifre, Schmitt, Guez, Lockhart, Hassabis, Graepel,
  et~al.]{schrittwieser2019mastering}
Julian Schrittwieser, Ioannis Antonoglou, Thomas Hubert, Karen Simonyan,
  Laurent Sifre, Simon Schmitt, Arthur Guez, Edward Lockhart, Demis Hassabis,
  Thore Graepel, et~al.
\newblock Mastering atari, go, chess and shogi by planning with a learned
  model.
\newblock \emph{arXiv preprint 1911.08265}, 2019.

\bibitem[Schulam and Saria(2017)]{schulam2017reliable}
Peter Schulam and Suchi Saria.
\newblock Reliable decision support using counterfactual models.
\newblock In \emph{Advances in Neural Information Processing Systems}, pages
  1697--1708, 2017.

\bibitem[Shah and Peters(2018)]{1804.07203}
Rajen~D. Shah and Jonas Peters.
\newblock The hardness of conditional independence testing and the generalised
  covariance measure.
\newblock \emph{arXiv preprint 1804.07203}, 2018.

\bibitem[Shajarisales et~al.(2015)Shajarisales, Janzing, Sch{\"o}lkopf, and
  Besserve]{Shajarisales15}
N.~Shajarisales, D.~Janzing, B.~Sch{\"o}lkopf, and M.~Besserve.
\newblock Telling cause from effect in deterministic linear dynamical systems.
\newblock In \emph{Proceedings of the 32nd International Conference on Machine
  Learning ({ICML})}, pages 285--294, 2015.

\bibitem[Shankar et~al.(2019)Shankar, Dave, Roelofs, Ramanan, Recht, and
  Schmidt]{shankarimage}
Vaishaal Shankar, Achal Dave, Rebecca Roelofs, Deva Ramanan, Benjamin Recht,
  and Ludwig Schmidt.
\newblock Do image classifiers generalize across time?
\newblock \emph{arXiv preprint 1906.02168}, 2019.

\bibitem[Shetty et~al.(2019)Shetty, Schiele, and Fritz]{shetty2019not}
Rakshith Shetty, Bernt Schiele, and Mario Fritz.
\newblock Not using the car to see the sidewalk--quantifying and controlling
  the effects of context in classification and segmentation.
\newblock In \emph{Proceedings of the IEEE Conference on Computer Vision and
  Pattern Recognition}, pages 8218--8226, 2019.

\bibitem[Shimizu et~al.(2006)Shimizu, Hoyer, Hyv\"{a}rinen, and
  Kerminen]{Shimizu2006}
S.~Shimizu, P.~O. Hoyer, A.~Hyv\"{a}rinen, and A.~J. Kerminen.
\newblock A linear non-{G}aussian acyclic model for causal discovery.
\newblock \emph{Journal of Machine Learning Research}, 7:\penalty0 2003--2030,
  2006.

\bibitem[Shu et~al.(2019)Shu, Chen, Kumar, Ermon, and Poole]{shu2019weakly}
Rui Shu, Yining Chen, Abhishek Kumar, Stefano Ermon, and Ben Poole.
\newblock Weakly supervised disentanglement with guarantees.
\newblock \emph{arXiv preprint 1910.09772}, 2019.

\bibitem[Silver et~al.(2016)Silver, Huang, Maddison, Guez, Sifre, Van
  Den~Driessche, Schrittwieser, Antonoglou, Panneershelvam, Lanctot,
  et~al.]{silver2016mastering}
David Silver, Aja Huang, Chris~J Maddison, Arthur Guez, Laurent Sifre, George
  Van Den~Driessche, Julian Schrittwieser, Ioannis Antonoglou, Veda
  Panneershelvam, Marc Lanctot, et~al.
\newblock Mastering the game of go with deep neural networks and tree search.
\newblock \emph{{N}ature}, 529\penalty0 (7587):\penalty0 484--489, 2016.

\bibitem[Silver et~al.(2017)Silver, Hasselt, Hessel, Schaul, Guez, Harley,
  Dulac-Arnold, Reichert, Rabinowitz, Barreto, et~al.]{silver2017predictron}
David Silver, Hado Hasselt, Matteo Hessel, Tom Schaul, Arthur Guez, Tim Harley,
  Gabriel Dulac-Arnold, David Reichert, Neil Rabinowitz, Andre Barreto, et~al.
\newblock The predictron: End-to-end learning and planning.
\newblock In \emph{International Conference on Machine Learning}, pages
  3191--3199. PMLR, 2017.

\bibitem[Simard et~al.(1992)Simard, Victorri, LeCun, and Denker]{tangent_prop}
Patrice Simard, Bernard Victorri, Yann LeCun, and John Denker.
\newblock Tangent prop - a formalism for specifying selected invariances in an
  adaptive network.
\newblock In J.~Moody, S.~Hanson, and R.~P. Lippmann, editors, \emph{Advances
  in Neural Information Processing Systems}, volume~4, pages 895--903.
  Morgan-Kaufmann, 1992.
\newblock URL
  \url{https://proceedings.neurips.cc/paper/1991/file/65658fde58ab3c2b6e5132a39fae7cb9-Paper.pdf}.

\bibitem[Simard et~al.(2003)Simard, Steinkraus, Platt, et~al.]{simard2003best}
Patrice~Y Simard, David Steinkraus, John~C Platt, et~al.
\newblock Best practices for convolutional neural networks applied to visual
  document analysis.
\newblock In \emph{{Proceedings of the Seventh International Conference on
  Document Analysis and Recognition (ICDAR 2003)}}, volume~3, 2003.

\bibitem[Simon(1953)]{Simon53}
H.~A. Simon.
\newblock Causal ordering and identifiability.
\newblock In W.~C. Hood and T.~C. Koopmans, editors, \emph{Studies in
  Econometric Methods}, pages 49--74. John Wiley \& Sons, New York, NY, 1953.
\newblock Cowles Commission for Research in Economics, Monograph No. 14.

\bibitem[Spelke(1990)]{spelke1990principles}
Elizabeth~S Spelke.
\newblock Principles of object perception.
\newblock \emph{Cognitive science}, 14\penalty0 (1):\penalty0 29--56, 1990.

\bibitem[Spirtes et~al.(2000)Spirtes, Glymour, and Scheines]{Spirtes2000}
P.~Spirtes, C.~Glymour, and R.~Scheines.
\newblock \emph{Causation, Prediction, and Search}.
\newblock MIT Press, Cambridge, MA, 2nd edition, 2000.

\bibitem[Spohn(1978)]{Spohn78}
W.~Spohn.
\newblock \emph{Grundlagen der Entscheidungstheorie}.
\newblock Scriptor-Verlag, 1978.

\bibitem[Steinwart and Christmann(2008)]{SteChr08}
I.~Steinwart and A.~Christmann.
\newblock \emph{Support Vector Machines}.
\newblock Springer, New York, NY, 2008.

\bibitem[Steudel et~al.(2010)Steudel, Janzing, and Sch\"olkopf]{Steudel2010a}
B.~Steudel, D.~Janzing, and B.~Sch\"olkopf.
\newblock Causal {M}arkov condition for submodular information measures.
\newblock In \emph{Proceedings of the 23rd Annual Conference on Learning Theory
  ({COLT})}, pages 464--476, 2010.

\bibitem[Su et~al.(2020)Su, Adams, and Beling]{su2020counterfactual}
Jianyu Su, Stephen Adams, and Peter~A Beling.
\newblock Counterfactual multi-agent reinforcement learning with graph
  convolution communication.
\newblock \emph{arXiv preprint 2004.00470}, 2020.

\bibitem[Subbaswamy and Saria(2018)]{subbaswamy2018counterfactual}
Adarsh Subbaswamy and Suchi Saria.
\newblock Counterfactual normalization: Proactively addressing dataset shift
  and improving reliability using causal mechanisms.
\newblock \emph{arXiv preprint 1808.03253}, 2018.

\bibitem[Subbaswamy et~al.(2018)Subbaswamy, Schulam, and Saria]{1812.04597}
Adarsh Subbaswamy, Peter Schulam, and Suchi Saria.
\newblock Preventing failures due to dataset shift: Learning predictive models
  that transport.
\newblock \emph{arXiv preprint 1812.04597}, 2018.

\bibitem[Sun et~al.(2017)Sun, Shrivastava, Singh, and Gupta]{sun2017revisiting}
Chen Sun, Abhinav Shrivastava, Saurabh Singh, and Abhinav Gupta.
\newblock Revisiting unreasonable effectiveness of data in deep learning era.
\newblock In \emph{Proceedings of the IEEE international conference on computer
  vision}, pages 843--852, 2017.

\bibitem[Sun et~al.(2019)Sun, Karlsson, Wu, Tenenbaum, and
  Murphy]{sun2019stochastic}
Chen Sun, Per Karlsson, Jiajun Wu, Joshua~B Tenenbaum, and Kevin Murphy.
\newblock Stochastic prediction of multi-agent interactions from partial
  observations.
\newblock \emph{arXiv preprint 1902.09641}, 2019.

\bibitem[Sun et~al.(2006)Sun, Janzing, and Sch\"{o}lkopf]{Sun2006}
X.~Sun, D.~Janzing, and B.~Sch\"{o}lkopf.
\newblock Causal inference by choosing graphs with most plausible {M}arkov
  kernels.
\newblock In \emph{Proceedings of the 9th International Symposium on Artificial
  Intelligence and Mathematics}, 2006.

\bibitem[Suter et~al.(2019)Suter, Miladinovic, Sch{\"o}lkopf, and
  Bauer]{Suter.1811.00007}
Raphael Suter, Djordje Miladinovic, Bernhard Sch{\"o}lkopf, and Stefan Bauer.
\newblock Robustly disentangled causal mechanisms: Validating deep
  representations for interventional robustness.
\newblock In \emph{International Conference on Machine Learning}, pages
  6056--6065. PMLR, 2019.

\bibitem[Sutton et~al.(1998)Sutton, Barto, et~al.]{sutton1998introduction}
Richard~S Sutton, Andrew~G Barto, et~al.
\newblock \emph{Introduction to reinforcement learning}, volume 135.
\newblock MIT press Cambridge, 1998.

\bibitem[Szegedy et~al.(2013)Szegedy, Zaremba, Sutskever, Bruna, Erhan,
  Goodfellow, and Fergus]{1312.6199}
Christian Szegedy, Wojciech Zaremba, Ilya Sutskever, Joan Bruna, Dumitru Erhan,
  Ian Goodfellow, and Rob Fergus.
\newblock Intriguing properties of neural networks.
\newblock \emph{arXiv preprint 1312.6199}, 2013.

\bibitem[T{\'e}gl{\'a}s et~al.(2011)T{\'e}gl{\'a}s, Vul, Girotto, Gonzalez,
  Tenenbaum, and Bonatti]{teglas2011pure}
Ern{\H{o}} T{\'e}gl{\'a}s, Edward Vul, Vittorio Girotto, Michel Gonzalez,
  Joshua~B Tenenbaum, and Luca~L Bonatti.
\newblock Pure reasoning in 12-month-old infants as probabilistic inference.
\newblock \emph{{Science}}, 332\penalty0 (6033):\penalty0 1054--1059, 2011.

\bibitem[Tian and Pearl(2001)]{Tian2001}
J.~Tian and J.~Pearl.
\newblock Causal discovery from changes.
\newblock In \emph{Proceedings of the 17th Annual Conference on Uncertainty in
  Artificial Intelligence ({UAI})}, pages 512--522, 2001.

\bibitem[Tr{\"a}uble et~al.(2020)Tr{\"a}uble, Creager, Kilbertus, Goyal,
  Locatello, Sch{\"o}lkopf, and Bauer]{trauble2020independence}
Frederik Tr{\"a}uble, Elliot Creager, Niki Kilbertus, Anirudh Goyal, Francesco
  Locatello, Bernhard Sch{\"o}lkopf, and Stefan Bauer.
\newblock Is independence all you need? on the generalization of
  representations learned from correlated data.
\newblock \emph{arXiv preprint 2006.07886}, 2020.

\bibitem[Tschannen et~al.(2020)Tschannen, Djolonga, Ritter, Mahendran, Houlsby,
  Gelly, and Lucic]{tschannen2020self}
Michael Tschannen, Josip Djolonga, Marvin Ritter, Aravindh Mahendran, Neil
  Houlsby, Sylvain Gelly, and Mario Lucic.
\newblock Self-supervised learning of video-induced visual invariances.
\newblock In \emph{Proceedings of the IEEE/CVF Conference on Computer Vision
  and Pattern Recognition}, pages 13806--13815, 2020.

\bibitem[Tsiaras et~al.(2019)Tsiaras, Waldmann, Tinetti, Tennyson, and
  Yurchenko]{Tsiaras}
Angelos Tsiaras, Ingo Waldmann, G.~Tinetti, Jonathan Tennyson, and Sergei
  Yurchenko.
\newblock Water vapour in the atmosphere of the habitable-zone eight-earth-mass
  planet {K2}-18b.
\newblock \emph{{Nature Astronomy}}, 2019.
\newblock \doi{10.1038/s41550-019-0878-9}.

\bibitem[Van~Steenkiste et~al.(2018)Van~Steenkiste, Chang, Greff, and
  Schmidhuber]{van2018relational}
Sjoerd Van~Steenkiste, Michael Chang, Klaus Greff, and J{\"u}rgen Schmidhuber.
\newblock Relational neural expectation maximization: Unsupervised discovery of
  objects and their interactions.
\newblock In \emph{6th International Conference on Learning Representations
  (ICLR)}, 2018.

\bibitem[van Steenkiste et~al.(2019)van Steenkiste, Locatello, Schmidhuber, and
  Bachem]{van2019disentangled}
Sjoerd van Steenkiste, Francesco Locatello, J{\"u}rgen Schmidhuber, and Olivier
  Bachem.
\newblock Are disentangled representations helpful for abstract visual
  reasoning?
\newblock In \emph{Advances in Neural Information Processing Systems}, pages
  14178--14191, 2019.

\bibitem[Vapnik(1998)]{Vapnik98}
V.~N. Vapnik.
\newblock \emph{Statistical Learning Theory}.
\newblock Wiley, New York, NY, 1998.

\bibitem[Vinyals et~al.(2019)Vinyals, Babuschkin, Czarnecki, Mathieu, Dudzik,
  Chung, Choi, Powell, Ewalds, Georgiev, et~al.]{vinyals2019grandmaster}
Oriol Vinyals, Igor Babuschkin, Wojciech~M Czarnecki, Micha{\"e}l Mathieu,
  Andrew Dudzik, Junyoung Chung, David~H Choi, Richard Powell, Timo Ewalds,
  Petko Georgiev, et~al.
\newblock Grandmaster level in {StarCraft II} using multi-agent reinforcement
  learning.
\newblock \emph{{N}ature}, 575\penalty0 (7782):\penalty0 350--354, 2019.

\bibitem[von Kügelgen et~al.(2020{\natexlab{a}})von Kügelgen, Mey, Loog, and
  Sch{\"o}lkopf]{KugMeyLooSch19}
J.~von Kügelgen, A.~Mey, M.~Loog, and B.~Sch{\"o}lkopf.
\newblock Semi-supervised learning, causality and the conditional cluster
  assumption.
\newblock \emph{Conference on Uncertainty in Artificial Intelligence (UAI)},
  2020{\natexlab{a}}.

\bibitem[von Kügelgen et~al.(2020{\natexlab{b}})von Kügelgen, Bhatt, Karimi,
  Valera, Weller, and Schölkopf]{vonkugelgen2020fairness}
Julius von Kügelgen, Umang Bhatt, Amir-Hossein Karimi, Isabel Valera, Adrian
  Weller, and Bernhard Schölkopf.
\newblock On the fairness of causal algorithmic recourse.
\newblock \emph{arXiv 2010.06529}, 2020{\natexlab{b}}.

\bibitem[von Kügelgen et~al.(2020{\natexlab{c}})von Kügelgen, Gresele, and
  Schölkopf]{vonkugelgen2020simpsons}
Julius von Kügelgen, Luigi Gresele, and Bernhard Schölkopf.
\newblock Simpson's paradox in {Covid-19} case fatality rates: a mediation
  analysis of age-related causal effects.
\newblock \emph{arXiv 2005.07180}, 2020{\natexlab{c}}.

\bibitem[von Kügelgen et~al.(2020{\natexlab{d}})von Kügelgen, Ustyuzhaninov,
  Gehler, Bethge, and Schölkopf]{Julius_ECON}
Julius von Kügelgen, Ivan Ustyuzhaninov, Peter Gehler, Matthias Bethge, and
  Bernhard Schölkopf.
\newblock Towards causal generative scene models via competition of experts.
\newblock \emph{arXiv 2004.12906}, 2020{\natexlab{d}}.

\bibitem[Wang et~al.(2019)Wang, He, Lipton, and Xing]{1903.06256}
Haohan Wang, Zexue He, Zachary~C. Lipton, and Eric~P. Xing.
\newblock Learning robust representations by projecting superficial statistics
  out.
\newblock \emph{arXiv preprint 1903.06256}, 2019.

\bibitem[Watters et~al.(2017)Watters, Zoran, Weber, Battaglia, Pascanu, and
  Tacchetti]{watters2017visual}
Nicholas Watters, Daniel Zoran, Theophane Weber, Peter Battaglia, Razvan
  Pascanu, and Andrea Tacchetti.
\newblock Visual interaction networks: Learning a physics simulator from video.
\newblock In \emph{Advances in neural information processing systems}, pages
  4539--4547, 2017.

\bibitem[Watters et~al.(2019)Watters, Matthey, Bosnjak, Burgess, and
  Lerchner]{watters2019cobra}
Nicholas Watters, Loic Matthey, Matko Bosnjak, Christopher~P Burgess, and
  Alexander Lerchner.
\newblock Cobra: Data-efficient model-based rl through unsupervised object
  discovery and curiosity-driven exploration.
\newblock \emph{arXiv preprint 1905.09275}, 2019.

\bibitem[Weichwald et~al.(2014)Weichwald, Sch{\"o}lkopf, Ball, and
  Grosse-Wentrup]{WeiSchBalGro14}
S.~Weichwald, B.~Sch{\"o}lkopf, T.~Ball, and M.~Grosse-Wentrup.
\newblock Causal and anti-causal learning in pattern recognition for
  neuroimaging.
\newblock In \emph{4th International Workshop on Pattern Recognition in
  Neuroimaging (PRNI)}. IEEE, 2014.

\bibitem[Weichwald(2019)]{weichwald2019pragmatism}
Sebastian Weichwald.
\newblock \emph{Pragmatism and Variable Transformations in Causal Modelling}.
\newblock PhD thesis, ETH Zurich, 2019.

\bibitem[Wiering and Van~Otterlo(2012)]{wiering2012reinforcement}
Marco Wiering and Martijn Van~Otterlo.
\newblock \emph{Reinforcement learning}, volume~12.
\newblock Springer, 2012.

\bibitem[Wiewel et~al.(2019)Wiewel, Becher, and Thuerey]{wiewel2019latent}
Steffen Wiewel, Moritz Becher, and Nils Thuerey.
\newblock Latent space physics: Towards learning the temporal evolution of
  fluid flow.
\newblock In \emph{Computer Graphics Forum}, volume~38, pages 71--82. Wiley
  Online Library, 2019.

\bibitem[Wiskott and Sejnowski(2002)]{Wiskott2002}
Laurenz Wiskott and Terrence~J Sejnowski.
\newblock {Slow feature analysis: unsupervised learning of invariances.}
\newblock \emph{Neural computation}, 14\penalty0 (4):\penalty0 715--70, April
  2002.
\newblock ISSN 0899-7667.

\bibitem[Xie et~al.(2016)Xie, Patil, Moldovan, Levine, and
  Abbeel]{xie2016model}
Chris Xie, Sachin Patil, Teodor Moldovan, Sergey Levine, and Pieter Abbeel.
\newblock Model-based reinforcement learning with parametrized physical models
  and optimism-driven exploration.
\newblock In \emph{2016 IEEE international conference on robotics and
  automation (ICRA)}, pages 504--511. IEEE, 2016.

\bibitem[Yi et~al.(2019)Yi, Gan, Li, Kohli, Wu, Torralba, and
  Tenenbaum]{yi2019clevrer}
Kexin Yi, Chuang Gan, Yunzhu Li, Pushmeet Kohli, Jiajun Wu, Antonio Torralba,
  and Joshua~B Tenenbaum.
\newblock {CLEVRER}: Collision events for video representation and reasoning.
\newblock \emph{arXiv preprint 1910.01442}, 2019.

\bibitem[Yoon et~al.(2018)Yoon, Jordon, and van~der Schaar]{yoon2018ganite}
Jinsung Yoon, James Jordon, and Mihaela van~der Schaar.
\newblock Ganite: Estimation of individualized treatment effects using
  generative adversarial nets.
\newblock In \emph{International Conference on Learning Representations}, 2018.

\bibitem[Zambaldi et~al.(2018)Zambaldi, Raposo, Santoro, Bapst, Li, Babuschkin,
  Tuyls, Reichert, Lillicrap, Lockhart, et~al.]{zambaldi2018deep}
Vinicius Zambaldi, David Raposo, Adam Santoro, Victor Bapst, Yujia Li, Igor
  Babuschkin, Karl Tuyls, David Reichert, Timothy Lillicrap, Edward Lockhart,
  et~al.
\newblock Deep reinforcement learning with relational inductive biases.
\newblock In \emph{International Conference on Learning Representations}, 2018.

\bibitem[Zhang and Bareinboim(2019)]{Bareinboim_NIPS2019}
J.~Zhang and E.~Bareinboim.
\newblock Near-optimal reinforcement learning in dynamic treatment regimes.
\newblock In \emph{Advances in Neural Information Processing Systems 33}, pages
  13401--13411, 2019.

\bibitem[Zhang and Bareinboim(2018)]{ZhaBar18}
Junzhe Zhang and Elias Bareinboim.
\newblock Fairness in decision-making - the causal explanation formula.
\newblock In \emph{Proceedings of the Thirty-Second {AAAI} Conference on
  Artificial Intelligence, New Orleans, Louisiana, USA}, pages 2037--2045,
  2018.

\bibitem[Zhang and Hyv\"{a}rinen(2009)]{Zhang2009}
K.~Zhang and A.~Hyv\"{a}rinen.
\newblock On the identifiability of the post-nonlinear causal model.
\newblock In \emph{Proceedings of the 25th Annual Conference on {U}ncertainty
  in {A}rtificial {I}ntelligence ({UAI})}, pages 647--655, 2009.

\bibitem[Zhang et~al.(2011)Zhang, Peters, Janzing, and
  Sch{\"o}lkopf]{Zhang2011uai}
K.~Zhang, J.~Peters, D.~Janzing, and B.~Sch{\"o}lkopf.
\newblock Kernel-based conditional independence test and application in causal
  discovery.
\newblock In \emph{Proceedings of the 27th Annual Conference on {U}ncertainty
  in {A}rtificial {I}ntelligence ({UAI})}, pages 804--813, 2011.

\bibitem[Zhang et~al.(2013)Zhang, Sch{\"o}lkopf, Muandet, and
  Wang]{zhang_domain_2013}
K.~Zhang, B.~Sch{\"o}lkopf, K.~Muandet, and Z.~Wang.
\newblock Domain adaptation under target and conditional shift.
\newblock In \emph{Proceedings of the 30th International Conference on Machine
  Learning ({ICML})}, pages 819--827, 2013.

\bibitem[Zhang et~al.(2015)Zhang, Gong, and
  Sch{\"o}lkopf]{zhang_multi-source_2015}
K.~Zhang, M.~Gong, and B.~Sch{\"o}lkopf.
\newblock Multi-source domain adaptation: A causal view.
\newblock In \emph{Proceedings of the 29th AAAI Conference on Artificial
  Intelligence}, pages 3150--3157, 2015.

\bibitem[Zhang et~al.(2017)Zhang, Huang, Zhang, Glymour, and
  Sch{\"o}lkopf]{Zhangetal17}
K.~Zhang, B.~Huang, J.~Zhang, C.~Glymour, and B.~Sch{\"o}lkopf.
\newblock Causal discovery from nonstationary/heterogeneous data: Skeleton
  estimation and orientation determination.
\newblock In \emph{Proceedings of the Twenty-Sixth International Joint
  Conference on Artificial Intelligence (IJCAI 2017)}, pages 1347--1353, 2017.

\bibitem[Zhang(2019)]{zhang2019making}
Richard Zhang.
\newblock Making convolutional networks shift-invariant again.
\newblock \emph{arXiv preprint 1904.11486}, 2019.

\end{thebibliography}
}

\end{document}